\documentclass{article}

\usepackage{iclr2027_conference,times}

\usepackage{amsmath,amsfonts,bm}

\def\eqref#1{equation~\ref{#1}}

\def\1{\bm{1}}

\DeclareMathAlphabet{\mathsfit}{\encodingdefault}{\sfdefault}{m}{sl}
\SetMathAlphabet{\mathsfit}{bold}{\encodingdefault}{\sfdefault}{bx}{n}

\usepackage{booktabs}
\usepackage{multirow}
\usepackage{colortbl}
\usepackage{tcolorbox}
\tcbuselibrary{breakable,skins}
\usepackage{algorithm}
\usepackage{algpseudocode}
\usepackage{amssymb}
\usepackage[section]{placeins}
\usepackage{needspace}
\usepackage{graphicx}
\usepackage{subcaption}
\usepackage{wrapfig}
\usepackage{hyperref}
\usepackage{url}
\hypersetup{
  pdftitle={Memorizing Who, Knowing How: Benchmarking Personal and Procedural Memory for Persistent Assistants},
  pdfauthor={Maosheng Qin, Qi Zhang, Mingxuan Xia, Jinglei Chen, Jiexiang Wang, Sheng Guo, Cheng Peng, Haobo Wang}
}

\newcommand{\cmark}{\ensuremath{\checkmark}}
\newcommand{\pmark}{\ensuremath{\circ}}
\newcommand{\xmark}{\ensuremath{\times}}

\title{Memorizing Who, Knowing How:\\
Benchmarking Personal and Procedural\\
Memory for Persistent Assistants}

\author{%
\textbf{Maosheng Qin}$^{1}$ \quad \textbf{Qi Zhang}$^{1}$ \quad
\textbf{Mingxuan Xia}$^{1}$ \quad \textbf{Jinglei Chen}$^{2}$\footnotemark[1]\\
\textbf{Jiexiang Wang}$^{2}$ \quad \textbf{Sheng Guo}$^{2}$ \quad
\textbf{Cheng Peng}$^{1}$ \quad \textbf{Haobo Wang}$^{1}$\thanks{\raggedright Corresponding authors: Haobo Wang (\texttt{wanghaobo@zju.edu.cn}) and Jinglei Chen (\texttt{chenjinglei.cjl@mybank.cn}).}\\[0.3em]
\normalfont\small $^{1}$Zhejiang University \quad $^{2}$Ant Group\\
\texttt{\{qinmaosheng,wanghaobo\}@zju.edu.cn} \quad
\texttt{chenjinglei.cjl@mybank.cn}%
}

\iclrfinalcopy

\begin{document}
\maketitle
\lhead{Preprint}
\begin{abstract}
Memory systems for persistent personal agents must move beyond isolated memories, maintaining both \emph{personal memory} about whom they serve and \emph{procedural memory} about how to act. Crucially, in harness-based personal assistance, \textbf{procedures or skills are inherently user-conditioned}: personal context determines which procedures apply, how they are instantiated, and when they should change. Existing benchmarks typically emphasize either personal or procedural memory, leaving their interaction underexplored. We introduce \textsc{PraxisMemBench}, a harness-based benchmark for persistent personal assistance across daily-life and work trajectories, where personal and procedural memory evolve together and later tasks depend on both. Experiments show that maintaining two memory types separately is insufficient, with completion dropping from single-type to interdependent dependencies (\textbf{84.6\%}$\rightarrow$\textbf{77.9\%}). More broadly, across existing methods, relation-related failures increase from early to late sessions as interaction history grows (\textbf{59.8\%}$\rightarrow$\textbf{72.9\%}), revealing a \textbf{continual-maintenance} challenge: writing correct memories is not enough if their evolving content and organization cannot be preserved for later retrieval. To address this, we propose \textsc{PraxisMem}, which uses privileged joint-state supervision to internalize the ability to maintain memory content and organization into the memory manager, with location-aware GRPO and dynamic residual-routed OPSD. \textsc{PraxisMem} reaches \textbf{40.78\%} overall Session Success Rate, compared with \textbf{20.60\%} for Claude Code's native memory.
\ificlrfinal
Our code is publicly available at \url{https://github.com/QMMMS/PraxisMemBench}.
\fi
\end{abstract}

\section{Introduction}
\label{sec:introduction}

Agent harnesses such as OpenClaw and Claude Code enable language models to operate applications, invoke tools, and execute multi-step tasks across practical domains \citep{zhu2026semaclaw,workbuddy2026benchmark,ji2026clawarena}. Beyond task-level capability, a broader goal is \emph{persistent personal assistance}, in which the same agent supports one user across a continuing stream of everyday life and work. In this longer-term setting, memory is a key mechanism for such persistence, allowing accumulated interaction history to shape later understanding and action \citep{packer2023memgpt,chhikara2025mem0,xu2025amem}. \emph{Personal memory} maintains evolving user-specific states and preferences, while \emph{procedural memory} preserves reusable ways of acting across applications. Crucially, \textbf{procedures are inherently user-conditioned}: personal context shapes which procedures apply, how they are instantiated, and when they should change.

Existing memory benchmarks capture two complementary but incomplete sides of persistent personal assistance. \textbf{Long-term memory benchmarks} ask whether models can recover information from dialogue or histories. Many emphasize \emph{personal memory} like facts, preferences, and their evolution \citep{maharana2024locomo,wu2025longmemeval,chen2026vitabench}. However, recall alone does not show whether such personal context can govern consequential actions in external applications. \textbf{Executable-agent benchmarks} instead evaluate \emph{procedural memory} through application workflows, reusable skills and executable action sequences \citep{he2026memoryarena,li2026skillsbench,ji2026clawarena}. Yet without persistent user context, they cannot determine whether a procedure is appropriate for a particular user or how it should adapt as that user changes. Together, these lines leave how personal and procedural memory interact in persistent personal assistance underexplored.

\begin{wrapfigure}{r}{0.60\textwidth}
  \centering
  \includegraphics[width=\linewidth]{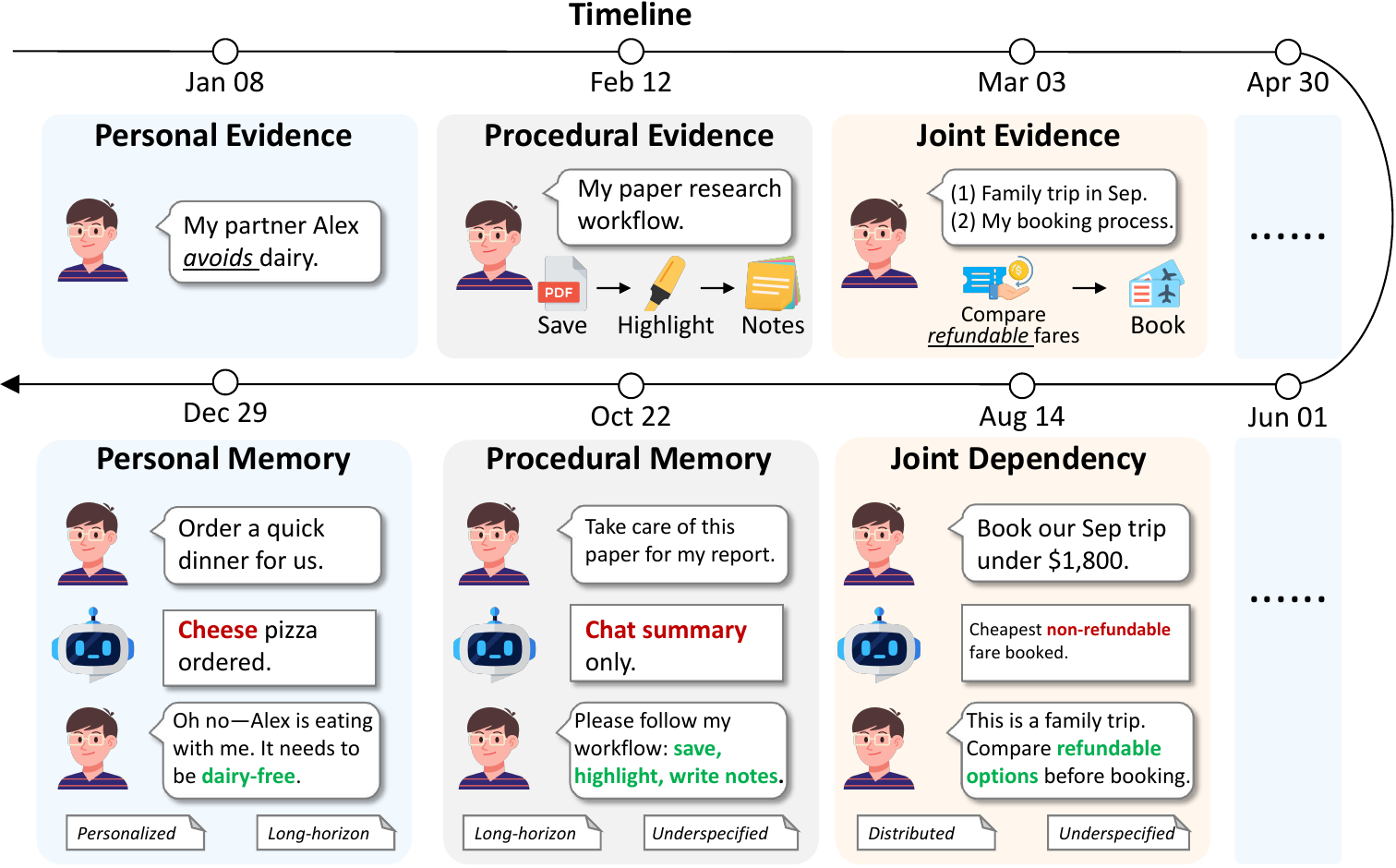}
  \caption{Representative personal, procedural, and joint memory failures in persistent personal assistance.}
  \label{fig:daily-life-cases}
\end{wrapfigure}

As illustrated in Figure~\ref{fig:daily-life-cases}, the central challenge in persistent personal assistance is that \textbf{procedures are user-conditioned}: personal memory determines which procedures apply, how they should be instantiated, and when they should change. This coupling is often triggered by \emph{underspecified} requests, since users rarely restate the relevant personal constraints or previously established procedures \citep{chen2026vitabench,zhu2026semaclaw}. Moreover, the required evidence is \emph{distributed} across sessions and application domains \citep{dou2026clbenchlife,chen2026vitabench}, while \emph{persistence} adds a maintenance challenge: changing preferences, prior task outcomes, and memory-maintenance decisions can alter which procedure remains appropriate in later sessions \citep{wu2025longmemeval,he2026memoryarena}. Consequently, personal and procedural memory cannot be maintained as two independent context slots; they must be continually revised and reorganized so that their evolving dependencies remain recoverable amid vast accumulated memory \citep{xu2025amem,zhang2026deltamem}.

\begin{table}[t]
  \centering
  \caption{Comparison of representative long-term memory (unshaded) and
  general agent or skill (shaded) benchmarks.
  \cmark, \pmark, and \xmark denote full, partial, and absent coverage.
  \emph{Mixed} denotes trajectories spanning multiple application domains; \emph{State dep.}
  denotes cross-session dependence on agent-created state or artifacts.}
  \label{tab:benchmark-comparison}
  \setlength{\tabcolsep}{3.2pt}
  \resizebox{0.95\linewidth}{!}{%
  \begin{tabular}{@{}lllrrcccc@{}}
    \toprule
    Benchmark & Setting & Horizon & Eval. & Domains & Mixed & Personal & Procedural & State dep. \\
    \midrule
    \rowcolor{gray!12}
    SkillsBench & Exec. agent & 1 task & 84 tasks & 11 & \xmark & \xmark & \cmark & \xmark \\
    \rowcolor{gray!12}
    ClawArena & Harness & 24--30 rounds & 12 scenarios & $\geq$7 & \xmark & \cmark & \xmark & \xmark \\
    \rowcolor{gray!12}
    WorkBuddy Bench & Harness & 1 task & 260 tasks & 4 & \xmark & \xmark & \xmark & \xmark \\
    \midrule
    LongMemEval & QA & 50--500 sessions & 500 Q & 1 & \xmark & \cmark & \xmark & \xmark \\
    Memoria-Bench & QA & Hundreds of turns & 58,422 Q & 7 & \xmark & \xmark & \cmark & \xmark \\
    VitaBench 2.0 & Exec. agent & 10--20 tasks & 56 users & 3 & \cmark & \cmark & \xmark & \xmark \\
    MemoryArena & Exec. agent & 2--16 sessions & 766 tasks & 4 & \xmark & \pmark & \pmark & \cmark \\
    \textbf{PraxisMemBench} & Harness & 25 sessions & 1,000 tasks & 40 & \cmark & \cmark & \cmark & \cmark \\
    \bottomrule
  \end{tabular}%
  }
\end{table}

To address this gap, we introduce \textsc{PraxisMemBench}, a benchmark that couples personal and procedural memory in persistent personal assistance. Across its executable tasks, \textbf{84.2\% require both forms of memory}, reflecting how user context and application procedures jointly determine successful action. Each persona follows a distinct \emph{trajectory spanning life and work domains}, where preferences, procedures, their associations, and application state evolve across sessions. Later requests often omit the relevant historical conditions, requiring the memory method to \textbf{recover the appropriate personal and procedural context} for downstream tool action. As these dependencies persist and change over time, earlier memory-management decisions can directly affect later tasks.

Experiments expose a sharper difficulty than retrieving either memory type alone. A dual-store baseline's completion declines from personal-only and procedural-only tasks to interdependent tasks (\textbf{84.6\%} $\rightarrow$ \textbf{77.9\%}). Moreover, persistence changes the nature of the failures: the share of relation-related failures rises from early to late sessions (\textbf{59.8\%} $\rightarrow$ \textbf{72.9\%}). These results point to a \textbf{continual-maintenance} problem: Correctly writing memory at one session does not guarantee long-term maintenance of personal--procedural associations. Yet scalar outcome supervision provides little indication of which part of the joint personal--procedural memory state is incorrect.

This observation motivates \textsc{PraxisMem}. During training, we use the canonical joint memory state as \emph{privileged supervision} and derive dynamic residuals against the manager's realized state to expose \emph{what} remains incorrect, \emph{where} it should be organized, and \emph{when} the defect persists. \textsc{PraxisMem} combines location-aware GRPO, which internalizes correct memory content and placement, with dynamic residual-routed OPSD, which supplies teacher guidance only for unresolved criteria and the chunks where their defects persist. On overall Session SR, \textsc{PraxisMem} reaches \textbf{40.78\%}, compared with 20.60\% for Claude Code's native memory.

\begin{figure}[t]
    \centering
    \includegraphics[width=0.9\textwidth]{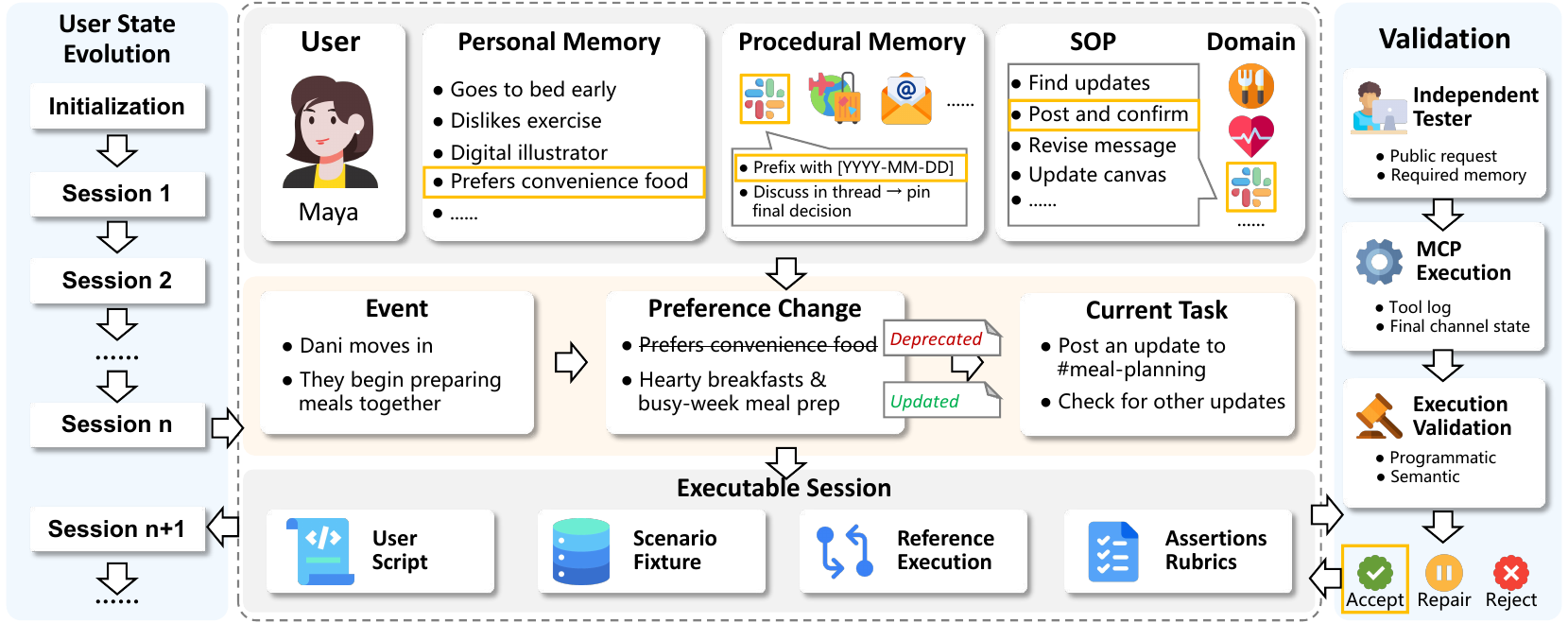}
    \caption{PraxisMemBench trajectory construction. Each session is instantiated from the evolving user, personal and procedural memory, and application context, then independently validated before updating the canonical state for subsequent sessions.}
    \label{fig:benchmark-construction}
\end{figure}

\section{PraxisMemBench}
\label{sec:benchmark}

\subsection{Problem Formulation}
\label{sec:benchmark-formulation}

PraxisMemBench follows a persistent agent serving the same user across life and work sessions. At session $n$, the user issues a natural request $q_n$ in application domain $d_n$. The task-agent policy $\pi$ interacts with persistent application state $S_n$, while the memory method $f$ maintains memory state $M_n$, comprising personal memory,
procedural memory, and their evolving association structure: 
\begin{equation}
M_n =
\bigl(
M_n^{\mathrm{pers}},
M_n^{\mathrm{proc}},
\mathcal I_n
\bigr),
\qquad
\bigl(H_n,M_{n+1},S_{n+1}\bigr)
=
\Phi_{\pi,f}^{d_n}
\bigl(q_n,M_n,S_n\bigr),
\label{eq:trajectory}
\end{equation}
where $H_n$ contains the dialogue, tool interactions, observations, and generated artifacts. Crucially, $\mathcal I_n$ denotes the \textbf{evolving association structure} through which personal context conditions relevant procedural memory. Thus, persistent interaction continually updates not only the two memory contents, but also the \emph{relations} that determine how they should be jointly used in later sessions.

\subsection{Benchmark Construction}
\label{sec:benchmark-construction}

\paragraph{Continuous trajectory annotation.}
Because PraxisMemBench targets \textbf{user-conditioned procedures} that evolve with personal context, benchmark construction must preserve how the \emph{joint memory state changes over time}. Therefore, we construct each persona trajectory \emph{sequentially}, rather than as independent tasks around a fixed profile. Before session $n$, the canonical annotation state $\Omega_n^\star$ contains the canonical memory state $M_n^\star$ and persistent application state $S_n^\star$:
\begin{equation}
\Omega_n^\star =
\bigl(M_n^\star,S_n^\star\bigr),
\qquad
(X_n,\Omega_{n+1}^\star)
=
\operatorname{Annotate}(\Omega_n^\star;e_n,d_n,\sigma_n),
\label{eq:construction}
\end{equation}
where $M_n^\star=(M_n^{\star,\mathrm{pers}}, M_n^{\star,\mathrm{proc}},\mathcal I_n^\star)$ is the canonical memory state, $e_n$ is a user event, $d_n$ an application domain, and $\sigma_n$ an application SOP. Each accepted session instantiates an executable task $X_n$ and advances the canonical state used to construct subsequent sessions. For example, in Figure~\ref{fig:benchmark-construction}, $e_n$ is the change that Dani moves in, $d_n$ is the selected communication application, and $\sigma_n$ specifies the reusable workflow for
posting and confirming an update.

\paragraph{Executable session instantiation.}
In order to instantiate these joint dependencies, the annotator selects relevant personal memories, applicable procedural memories, an application, and an application-specific SOP from $\Omega_n^\star$. Following Eq.~\ref{eq:construction}, these ingredients are instantiated through a current event or preference change, a natural user request, and execution in the selected application. By construction, the task requires \textbf{personal memory} (e.g., Maya's updated meal preference) and \textbf{procedural memory} (e.g., a previously established Slack workflow) to jointly determine a measurable outcome. Their \textbf{association} $\mathcal I_n^\star$ is instantiated when personal context constrains the candidate space, selects among procedural outcomes, or changes how a procedure should be applied. When a memory is first established, the user states the relevant personal information or procedure naturally but explicitly. Crucially, in later dependent sessions, these details are \textbf{omitted} from the public request, so the evaluated memory method must select the \textbf{related personal and procedural memories} relevant to the current request and user state. The resulting session $X_n$ contains the application fixture, reference execution, and evaluation checks. As detailed in Appendix~\ref{app:construction-pipeline}, only independently validated sessions advance the canonical state in Eq.~\ref{eq:construction}; their outcomes and new memories may then shape subsequent sessions.

\subsection{Evaluation}
\label{sec:benchmark-evaluation}

\begin{figure}[t]
    \centering
    \includegraphics[width=0.9\textwidth]{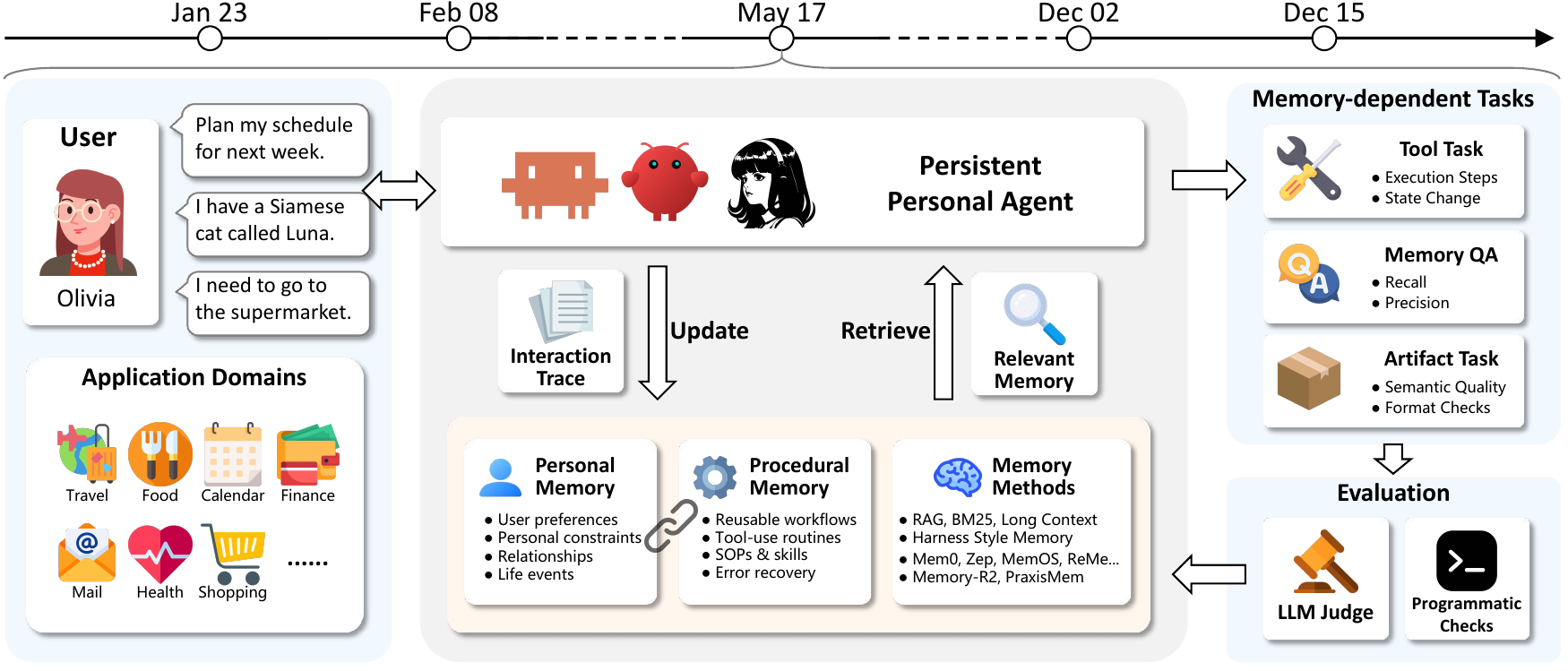}
    \caption{PraxisMemBench evaluation lifecycle. Across sessions, the memory method continuously updates and retrieves personal and procedural memory from user interactions. The maintained memory persists across the trajectory and shapes later memory-dependent tasks, whose outcomes are evaluated by LLM judges and programmatic checks.}
    \label{fig:benchmark-runtime}
\end{figure}

\paragraph{Persistent agent lifecycle.}
To realize the persistent transition in Eq.~\ref{eq:trajectory}, PraxisMemBench evaluates each memory method $f$ within the agent--harness--environment loop shown in Figure~\ref{fig:benchmark-runtime}. In session $n$, the user issues $q_n$; memory is accessed according to the evaluation setting, and the task agent interacts with the applications to produce trace $H_n$ and state $S_{n+1}$. After each session, the resulting interaction updates $M_{n+1}$ before the next request arrives. Both $M_{n+1}$ and $S_{n+1}$ are propagated to the next session without canonical-state correction. Thus, persistent evaluation requires \emph{continuously} maintaining personal and procedural memory together with their \textbf{evolving associations}, so that user-conditioned procedures remain current, organized, and retrievable over the trajectory.

\paragraph{Diagnostic evaluation design.} To capture both memory-conditioned action and diagnostic memory quality, PraxisMemBench evaluates three session families. \emph{Tool} and \emph{Artifact} tasks report personal-memory, procedural-memory, and execution-level scores (Tool Step SR or Artifact Assert), while \emph{Memory QA} isolates retrieval quality through Precision, Recall, and F1. All three families additionally report Session SR. Semantic requirements are judged by an LLM, whereas tool and artifact outcomes are checked programmatically. Detailed definitions are provided in Appendix~\ref{app:evaluation-details}.

\paragraph{Benchmark coverage.}
PraxisMemBench provides separate \emph{training} and \emph{test} splits, each with \textbf{20} users and \textbf{500} sessions spanning \textbf{20} MCP-accessible application domains (\textbf{40} in total), including travel, shopping, health, food, and finance. The test split is evaluated under two settings, \emph{Scheduled} and \emph{Autonomous}, covering \textbf{2,185} memory dependencies. Scheduled fixes memory intervention points for controlled and reproducible evaluation, whereas Autonomous uses more underspecified requests and lets the assistant decide when to invoke memory operations; full setting details are provided in the appendix. Figure~\ref{fig:benchmark-statistics} shows that \textbf{303} of 360 executable tasks require joint memory, while many dependencies span months of time. This combination places strong pressure on \textbf{continual relation maintenance}: a correct update at one session does not guarantee that its personal--procedural relation remains intact over time. As we show later, longer dependencies increasingly produce fragmented or competing relations.

\begin{figure*}[t]
    \centering
    \begin{minipage}[t]{0.3\textwidth}
        \vspace{0pt}
        \centering
        \includegraphics[width=0.96\linewidth]{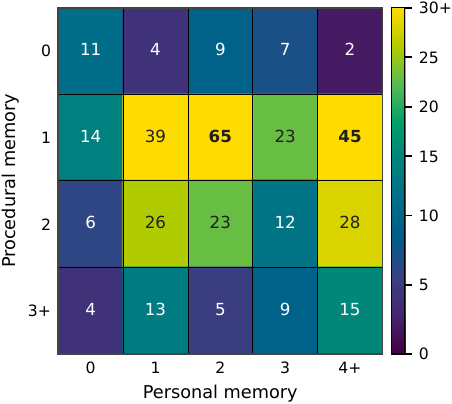}
    \end{minipage}
    \hfill
    \begin{minipage}[t]{0.3\textwidth}
        \vspace{0pt}
        \centering
        \includegraphics[width=0.96\linewidth]{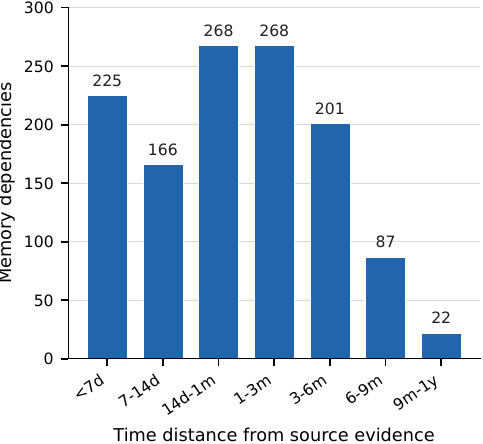}
    \end{minipage}
    \hfill
    \begin{minipage}[t]{0.26\textwidth}
        \vspace{0pt}
        \centering
        \includegraphics[width=0.96\linewidth]{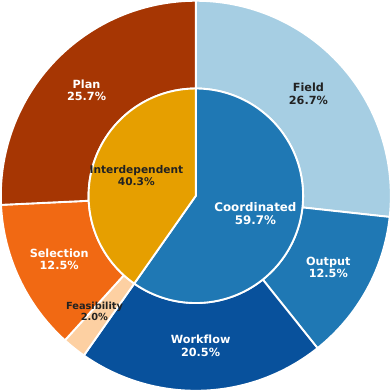}
    \end{minipage}
    \caption{Memory-dependency structure in the test split. From left to right:
    (a) Distribution of personal and procedural memory points required by
    executable tasks;
    (b) Temporal distance from source evidence for dependencies in later
    sessions; (c)
    Joint user-conditioning taxonomy.}
    \label{fig:benchmark-statistics}
\end{figure*}

\section{PraxisMem: Learning a Persistent Memory Manager}
\label{sec:method}

Notably, outcome-driven memory managers typically learn from final task rewards. However, such scalar supervision provides little indication of which part of the joint personal--procedural memory state is incorrect. Addressing this \textbf{ambiguity} requires the memory manager to understand, during each update, how user context and procedural knowledge should jointly shape the desired memory state. Fortunately, during training, the canonical structured memory---including user and event states, reusable skills, and their associations---provides exactly this information. The key is therefore to \textbf{internalize this supervision} into the manager without requiring it at deployment. 

Hence, we propose \textsc{PraxisMem}, which converts the canonical user--procedure state into \emph{privileged information}~\citep{zhao2026selfdistilled} by computing a dynamic residual that reveals \emph{what} is missing or incorrect, \emph{where} it belongs, and \emph{when} the defect persists along the update trajectory. Specifically, \textbf{location-aware GRPO} learns \emph{what} information should be maintained and \emph{where} it should be organized, while \textbf{dynamic residual-routed OPSD} identifies \emph{when} unresolved defects persist along the update trajectory and provides targeted guidance only at those chunks. Together, the two objectives train the manager to maintain personal memory, procedural memory, and their evolving associations within a unified memory state. We focus on Update below because its state changes persist and shape later memory use.

\subsection{Privileged Joint-State Supervision}
\label{sec:method-markdown}

During training, each session provides a canonical joint state $M_n^\star=(M_n^{\star,\mathrm{pers}},M_n^{\star,\mathrm{proc}}, \mathcal I_n^\star)$, specifying the desired personal memory, procedural memory, and their associations. We realize this state with a simple structured Markdown store: \texttt{USER.md} and \texttt{events/*.md} maintain personal and event context, domain-specific \texttt{SKILL.md} files maintain reusable procedures, and \texttt{MEMORY.md} indexes their associations. We treat the canonical states and their discrepancy from the manager's realized states as \textbf{privileged supervision}: they specify the desired joint personal--procedural organization and expose the remaining gap that the manager must correct. Full memory organization and update mechanics are provided in Appendix~\ref{app:memrsd-state}.

\subsection{Location-Aware GRPO}
\label{sec:method-grpo}

As motivated above, single-step semantic correctness does not guarantee \emph{long-term maintenance of the association structure} $\mathcal I_n$: a memory update can be correct yet poorly placed for preserving its \textbf{relation} to the relevant user context or procedure after subsequent updates. We therefore extend the soft semantic state matching of DeltaMem~\citep{zhang2026deltamem} with location-aware supervision, using the canonical joint state to internalize in the manager both \emph{what} to write and \emph{where} it belongs.

Let
$\widehat{\Delta}_i=\{\widehat u_a\}$ and
$\Delta_n^\star=\{u_b^\star\}$ denote the predicted and canonical memory changes from $M_n^\star$. Specifically, type-compatible units above a semantic threshold form candidate matches. We select a one-to-one assignment that prioritizes correct location, then match coverage and semantic similarity:
\begin{equation}
\gamma_i^\star
=
\operatorname*{lexargmax}_{\gamma\subseteq\mathcal E_i}
\left(
\sum_{(a,b)\in\gamma}\ell_{ab},
\,|\gamma|,
\,\sum_{(a,b)\in\gamma}s_{ab}
\right),
\label{eq:location-aware-matching}
\end{equation}
where $\ell_{ab}$ indicates location correctness and $s_{ab}$ semantic similarity. Matched units contribute according to both content coverage and location, while unmatched predictions penalize redundant memory. Their location-weighted soft precision and recall define $F_{\mathrm{state}}^{(i)}$.
The Update reward is: 
\begin{equation}
R_i^{\mathrm{upd}}
=
0.8F_{\mathrm{state}}^{(i)}
+
0.1R_{\mathrm{format}}^{(i)}
+
0.1R_{\mathrm{eff}}^{(i)},
\label{eq:update-reward}
\end{equation}
where the auxiliary terms encourage valid memory operations and discourage unnecessary actions or state growth. With this reward, we use standard GRPO~\citep{shao2024deepseekmath} to compute group-relative advantages and optimize the complete manager rollout. However, when every rollout misses the same required memory criterion, group-relative learning provides no successful behavior for that criterion, motivating the residual-routed self-distillation introduced next.

\subsection{Dynamic Residual-Routed Self-Distillation}
\label{sec:method-opsd}

\begin{figure}[t]
\centering
\includegraphics[width=0.9\textwidth]{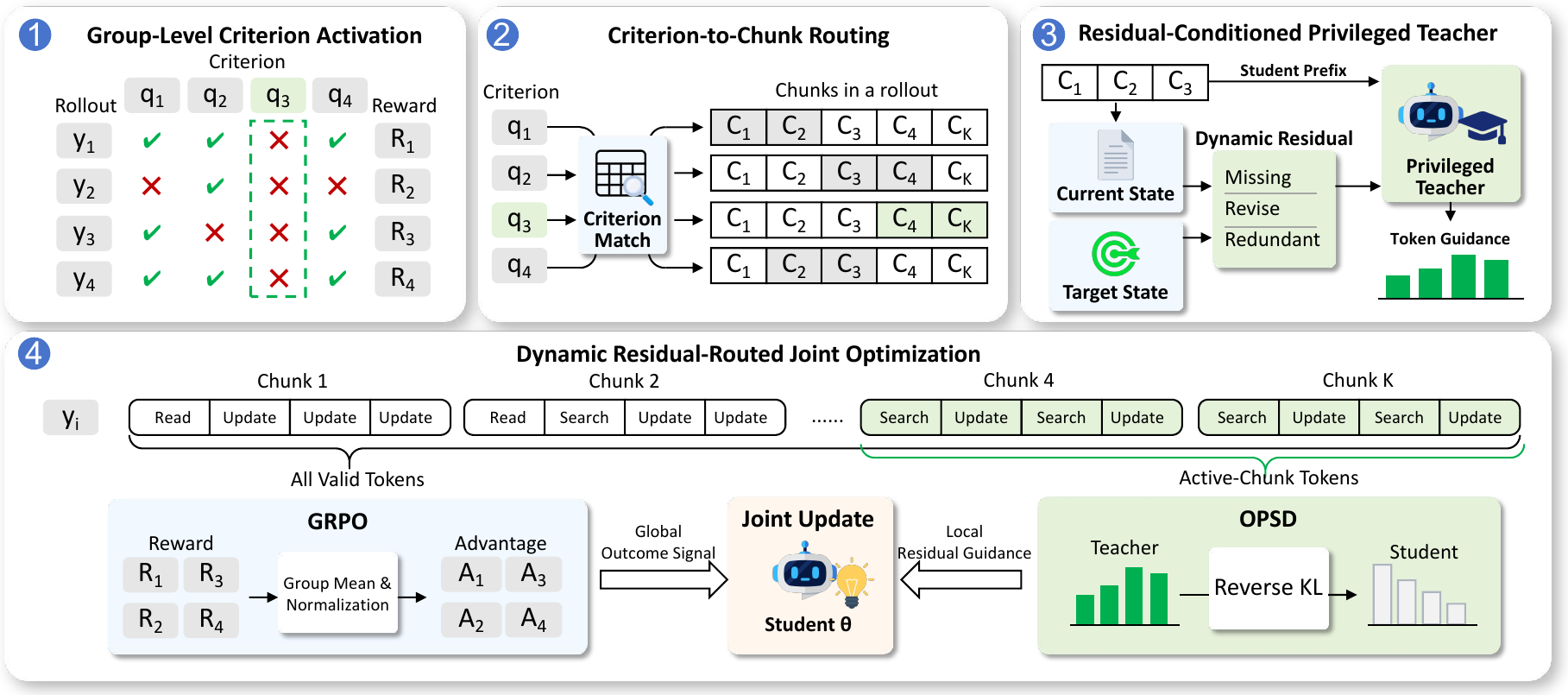}
\caption{Overview of \textsc{PraxisMem}. (1) Criteria missed by all rollouts are activated. (2) They are routed to chunks where the corresponding memory defects persist. (3) A privileged teacher receives the dynamic residual between current and target memory states to provide token-level guidance. (4) GRPO trains the full rollout, while OPSD supervises only routed chunks.}
\label{fig:dynamic-residual-opsd}
\end{figure}

Even with location-aware rewards, a joint memory failure remains ambiguous when every rollout misses the same criterion.  \textsc{PraxisMem} therefore 
turns the privileged state discrepancy into targeted on-policy self-distillation (OPSD)~\citep{zhao2026selfdistilled}. Criterion routing identifies \emph{which} content or relation remains unresolved, while dynamic residual routing identifies \emph{when} that defect persists along the trajectory and supplies guidance only to those chunks. Figure~\ref{fig:dynamic-residual-opsd} illustrates this process.

\paragraph{Selective criterion--chunk routing.}
Let $c_{ij}$ denote the final location-aware coverage of required memory criterion $\kappa_j$ in rollout $i$. We activate OPSD only when no rollout satisfies that criterion:
\begin{equation}
u_j
=
\mathbb{I}\!\left[
\max_i c_{ij}<\tau_{\mathrm{cov}}
\right],
\qquad
\mathcal U=\{j:u_j=1\}.
\label{eq:criterion-activation}
\end{equation}
For each update chunk $k$, let $M_{i,k}^{-}$ denote the realized memory state of rollout $i$ before update, and $M_{n,k}^\star$ its canonical target. We then compare the rollout's realized state with the canonical state:
\begin{equation}
z_{i,k}
=
\operatorname{SemDiff}
\bigl(
M_{i,k}^{-},
M_{n,k}^\star;
M_{n,k-1}^\star
\bigr),
\qquad
g_{i,k}
=
\mathbb{I}\!\left[
\mathcal J(z_{i,k})\cap\mathcal U\neq\varnothing
\right].
\label{eq:dynamic-residual}
\end{equation}
Here, $\mathcal J(z_{i,k})$ extracts unresolved required criteria from the current residual. An earlier omission, conflict, or misplacement keeps the gate active until it is repaired.

\paragraph{Dynamic residual-conditioned privileged guidance.}
The residual $z_{i,k}$ is recomputed from the rollout's realized memory state at each chunk, \textbf{dynamically tracking discrepancies} from the canonical state, including missing, conflicting, misplaced, and redundant content. It is exposed only to the self-teacher during training, providing chunk-specific guidance on what memory should be added, revised, moved, or removed as the rollout evolves. Finally, we compute the reverse-KL OPSD loss between the residual-conditioned teacher and the unprivileged student, as detailed in Appendix~\ref{app:memrsd-opsd}.

\paragraph{Joint training.}
Let $\mathcal V$ contain all valid manager-generated tokens and
$\mathcal A=\{(i,k,t)\in\mathcal V:g_{i,k}=1\}$ the tokens selected by residual routing. We optimize
\begin{equation}
\mathcal L_{\mathrm{joint}}
=
\mathcal L_{\mathrm{GRPO}}(\mathcal V)
+
\lambda_{\mathrm{OPSD}}
\mathcal L_{\mathrm{OPSD}}(\mathcal A)
+
\beta\mathcal L_{\mathrm{refKL}}(\mathcal V).
\label{eq:joint-memory-objective}
\end{equation}
Thus, location-aware GRPO shapes the content and organization of the joint memory state, while residual-routed OPSD supplies guidance for unresolved personal, procedural, or association defects.

\begin{table}[t]
    \centering
    \caption{Overall results on the PraxisMemBench test set. Tool tasks report
    personal-memory (Pers.), procedural-memory (Proc.), step success rate
    (Step SR), and session success rate (Session SR). Memory QA reports recall,
    precision, F1, and SR. Artifact tasks report Pers., Proc., artifact-assertion
    score (Assert.), and SR.}
    \label{tab:overall-results}
    \providecommand{\tbdscore}{\makebox[2.05em][c]{\textcolor{gray}{--}}}
    \resizebox{\linewidth}{!}{%
    \begin{tabular}{@{}lcccccccccccc@{}}
        \toprule
        Method & \multicolumn{4}{c}{Tool}
        & \multicolumn{4}{c}{Memory QA}
        & \multicolumn{4}{c}{Artifact} \\
        \cmidrule(lr){2-5}\cmidrule(lr){6-9}\cmidrule(l){10-13}
        & Pers. & Proc. & Step SR & SR
        & Recall & Precision & F1 & SR
        & Pers. & Proc. & Assert. & SR \\
        \midrule
        \rowcolor{gray!12}
        \multicolumn{13}{@{}l}{\textit{Diagnostic}} \\
        Oracle
        & 90.42 & 93.98 & 93.59 & 83.90
        & 98.45 & 99.29 & 98.82 & 99.29
        & 85.18 & 90.32 & 93.16 & 77.42 \\
        Long Context
        & 39.68 & 44.91 & 72.71 & 21.46
        & 43.37 & 57.06 & 46.88 & 30.71
        & 52.69 & 51.80 & 80.56 & 30.26 \\
        DualMem
        & 27.73 & 51.30 & 79.23 & 15.90
        & 16.21 & 35.61 & 20.19 & 0.71
        & 39.77 & 47.41 & 86.89 & 13.73 \\
        \midrule
        \rowcolor{gray!12}
        \multicolumn{13}{@{}l}{\textit{Basic baselines}} \\
        BM25
        & 22.57 & 30.93 & 68.08 & 15.12
        & 37.73 & 57.07 & 42.84 & 22.14
        & 27.88 & 25.42 & 77.71 & 6.45 \\
        Embedding RAG
        & 26.83 & 36.30 & 73.47 & 16.10
        & 34.86 & 50.51 & 38.83 & 20.71
        & 34.41 & 31.63 & 81.35 & 11.61 \\
        \midrule
        \rowcolor{gray!12}
        \multicolumn{13}{@{}l}{\textit{Harness-native memory}} \\
        OpenClaw
        & 20.58 & 24.91 & 62.48 & 15.12
        & 12.28 & 25.06 & 15.50 & 1.43
        & 21.03 & 27.67 & 64.36 & 8.39 \\
        Hermes
        & 37.22 & 54.49 & 74.84 & 26.34
        & 21.87 & 42.98 & 25.93 & 10.71
        & 45.80 & 55.67 & 77.66 & 23.23 \\
        Claude Code
        & 37.83 & 46.53 & 71.92 & 22.44
        & 35.17 & 54.75 & 40.00 & 15.71
        & 41.87 & 47.68 & 75.18 & 22.58 \\
        \midrule
        \rowcolor{gray!12}
        \multicolumn{13}{@{}l}{\textit{Advanced memory systems}} \\
        Mem0
        & 20.78 & 47.89 & 77.78 & 12.42
        & 19.17 & 44.64 & 24.05 & 2.14
        & 29.34 & 41.54 & 80.29 & 13.01 \\
        Zep
        & 24.11 & 39.20 & 73.69 & 18.41
        & 13.92 & 29.64 & 17.61 & 1.43
        & 33.37 & 33.90 & 81.23 & 15.48 \\
        A-Mem
        & 25.03 & 42.70 & 76.45 & 17.46
        & 24.23 & 40.73 & 28.02 & 7.14
        & 32.37 & 34.17 & 80.56 & 11.03 \\
        ReMe
        & 24.39 & 29.97 & 67.39 & 13.33
        & 10.35 & 22.80 & 13.05 & 0.00
        & 31.43 & 22.79 & 75.60 & 6.90 \\
        MemOS
        & 29.47 & 56.34 & 80.30 & 19.12
        & 18.98 & 41.48 & 23.09 & 3.57
        & 40.01 & 50.94 & 84.77 & 18.18 \\
        \midrule
        \rowcolor{gray!12}
        \multicolumn{13}{@{}l}{\textit{Learned memory managers}} \\
        Memory-R2
        & 34.93 & 51.76 & 79.14 & 22.93
        & 22.61 & 45.07 & 26.79 & 5.71
        & 38.85 & 42.90 & 84.00 & 17.42 \\
        \textbf{PraxisMem (Ours)}
        & \textbf{63.46} & \textbf{61.02} & \textbf{80.91} & \textbf{37.88}
        & \textbf{69.64} & \textbf{82.30} & \textbf{74.27} & \textbf{55.00}
        & \textbf{69.40} & \textbf{59.32} & \textbf{89.66} & \textbf{31.33} \\
        \bottomrule
    \end{tabular}
    }
\end{table}

\section{Experiments}
\label{sec:experiments}

\subsection{Experimental Setup}
\label{sec:experimental-setup}

\paragraph{Models and protocol.}
We use Qwen3.7-Plus as the task agent and GLM-5 as both the user simulator and LLM judge, with deterministic programmatic verifiers for tool and artifact outcomes. Qwen3.5-9B serves as the base model for LLM-based memory methods, and Qwen3-Embedding-0.6B for embedding-based components. We evaluate the \emph{Scheduled} and \emph{Autonomous} settings separately using the metrics in Section~\ref{sec:benchmark-evaluation}, with results aggregated over complete trajectories.

\begin{figure*}[t]
    \centering
    \begin{minipage}[t]{0.325\textwidth}
        \vspace{0pt}
        \centering
        \includegraphics[width=0.96\linewidth]{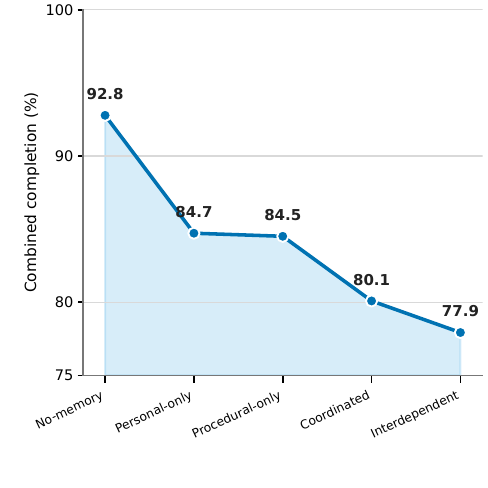}
    \end{minipage}\hfill
    \begin{minipage}[t]{0.325\textwidth}
        \vspace{0pt}
        \centering
        \includegraphics[width=0.96\linewidth]{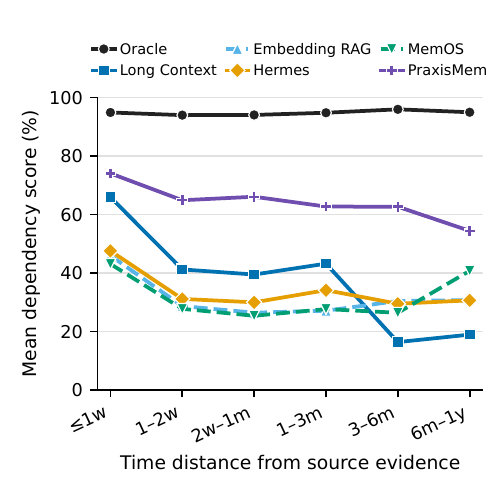}
    \end{minipage}
    \hfill
    \begin{minipage}[t]{0.325\textwidth}
        \vspace{0pt}
        \centering
        \includegraphics[width=0.96\linewidth]{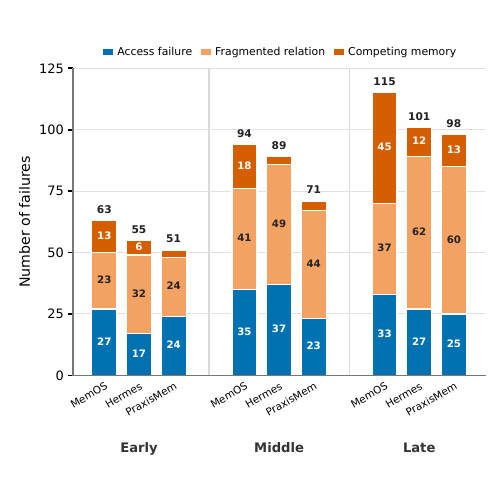}
    \end{minipage}
    \caption{Memory use across dependency types and persistent interaction.
    From left to right: (a) DualMem Combined Completion by dependency type; (b) Long-range
    degradation by evidence age; (c)
    Counts of memory failures for
    MemOS, Hermes, and \textsc{PraxisMem} across trajectory stages.}
    \label{fig:dependency-persistence-findings}
\end{figure*}

\paragraph{Memory systems.}
We compare five groups of methods. Diagnostic controls include Oracle Memory, Long Context, and \emph{DualMem}, which maintains personal and procedural memory in separate stores and combines their retrieved contents only at task time. Retrieval baselines include BM25 and embedding RAG. Harness-style baselines include OpenClaw, Hermes, and Claude Code. Advanced memory systems include Mem0, Zep, A-Mem, ReMe, MemOS, Memory-R2, and \textsc{PraxisMem} \citep{chhikara2025mem0,rasmussen2025zep,xu2025amem,cao2026reme,li2025memos,yan2026memoryr2}. We preserve each system's original memory representation and update--retrieval behavior; versions and method-specific configurations are provided in Appendix~\ref{app:configurations}.

\subsection{Overall Benchmark Results}
\label{sec:overall-results}

\paragraph{Overall performance.}
Table~\ref{tab:overall-results} reveals four consistent patterns. (1) PraxisMemBench remains challenging even for strong memory-augmented agents: component-level scores are consistently higher than session success, showing that recalling individual memories or satisfying isolated execution requirements does not guarantee complete task success. (2) Among harness-native systems, Hermes gives the strongest overall Session SR, while Claude Code performs best on Memory QA. (3) Advanced memory systems remain uneven across task families: For Session SR, MemOS performs
best on Tool and Artifact tasks, whereas A-Mem leads on Memory QA, suggesting
that stronger performance on one form of memory use does not reliably transfer
to others. (4) \textsc{PraxisMem} performs strongly across all three task families and achieves the best overall results.

\subsection{Analysis and Findings}
\label{sec:findings}

\paragraph{Finding 1: Joint dependencies take coordinated and interdependent forms.}
We distinguish two forms of joint dependency, both reflecting how procedures in persistent personal assistance are conditioned on user context. In \textbf{Coordinated} tasks, personal and procedural memory must be used together to determine the task outcome. This includes \emph{Field Coordination}, where both memories jointly determine the same field; \emph{Output Coordination}, where they jointly shape different parts of an output; and \emph{Workflow Coordination}, where they jointly govern different actions or steps. In \textbf{Interdependent} tasks, this conditioning is stronger: personal context changes how procedural knowledge can be applied. \emph{Action Feasibility} determines which actions remain valid, \emph{Action Selection} determines which candidate action or object should be chosen, and \emph{Plan Formation} changes how the overall procedure or plan should be composed. Among audited joint tasks, 59.7\% are Coordinated and 40.3\% are Interdependent. The right panel of Figure~\ref{fig:benchmark-statistics} summarizes this distribution.

\paragraph{Finding 2: Separate maintenance struggles with stronger interdependence.}
To test whether maintaining personal and procedural memory independently is sufficient, we implement \emph{DualMem}, which uses separate stores for the two memory types and combines their retrieved contents only before task execution. However, in general, its performance remains limited, motivating a closer analysis by dependency type. We combine Tool Step SR and Artifact Assertion into a single diagnostic metric, \emph{Combined Completion}. As shown in the left panel of Figure~\ref{fig:dependency-persistence-findings}, DualMem performs progressively worse from personal-only (84.7\%) and procedural-only (84.5\%) tasks to Coordinated tasks (80.1\%), and drops further on Interdependent tasks (77.9\%). This suggests that Coordinated tasks already require compatible personal and procedural evidence to be maintained together, while Interdependent tasks are harder because the \textbf{relevant procedural memory depends on the current personal context}. DualMem preserves the two memory contents separately but leaves this association implicit, making individually plausible but jointly \emph{incompatible} memories more likely.

\begin{figure*}[t]
    \centering
    \begin{minipage}[t]{0.62\textwidth}
        \centering
        \captionof{table}{Controlled ablation on eight fixed test
        trajectories. Combined and Interdependent Completion use Tool Step SR
        and Artifact Assertion.}
        \label{tab:memrsd-ablation}
        \vspace{0.35em}
        \small
        \setlength{\tabcolsep}{4.2pt}
        \renewcommand{\arraystretch}{1.12}
        \resizebox{\linewidth}{!}{%
        \begin{tabular}{@{}lccc@{}}
            \toprule
            Variant & \shortstack{Combined\\Completion} &
            \shortstack{Interdependent\\Completion} &
            Avg SR \\
            \midrule
            Content-only GRPO & 66.2 & 54.9 & 23.5 \\
            $+$ Location-aware reward & 73.6 & 66.5 & 29.0 \\
            $+$ Criterion-routed OPSD & 76.9 & 72.1 & 32.7 \\
            \textbf{\textsc{PraxisMem} (criterion$\times$chunk)} & \textbf{83.4} & \textbf{82.1} & \textbf{43.1} \\
            \bottomrule
        \end{tabular}
        }
    \end{minipage}\hfill
    \begin{minipage}[t]{0.34\textwidth}
        \vspace{0pt}
        \centering
        \includegraphics[width=\linewidth]{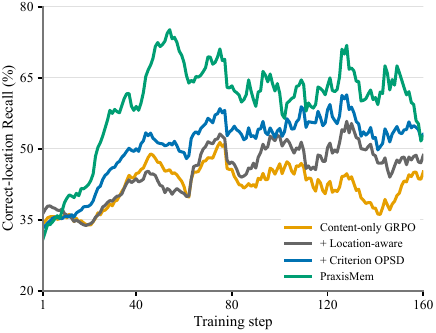}
        \captionsetup{justification=raggedright,singlelinecheck=false}
        \captionof{figure}{Correct-location Recall across training variants.}
        \label{fig:memrsd-training}
    \end{minipage}
\end{figure*}

\paragraph{Finding 3: Persistence turns memory coupling into a continual-maintenance problem.}
The middle panel of Figure~\ref{fig:dependency-persistence-findings} examines evidence age using \emph{Mean Dependency Score}, the average score over all personal and procedural dependency points within each time-distance bucket. Compared with dependencies sourced within one week, the point-weighted mean over later buckets drops by 30.0 points for Long Context, 17.9 for Embedding RAG, and 15.3 for MemOS. This indicates that \textbf{persistent interaction} increasingly stresses memory use as relevant evidence recedes in time.

To understand the source of degradation, we audit the inference-time evidence produced by MemOS, Hermes, and \textsc{PraxisMem}. We classify failures as \emph{Access failure}, where relevant personal or procedural evidence is missing; \emph{Fragmented relation}, where both are available but their task-relevant relation is not coherently surfaced; and \emph{Competing memory}, where a concrete but inapplicable candidate is retrieved instead. The right panel of Figure~\ref{fig:dependency-persistence-findings} shows a clear shift in failure composition: Access failures fall from 40.2\% in early sessions to 27.1\% in late sessions, while relation-related failures rise from 59.8\% to 72.9\%, increasingly dominated by fragmented relations. These failures point to a \textbf{continual-maintenance problem}: persistent memory systems must not only retain \emph{personal} and \emph{procedural} content, but continually revise and reorganize the \textbf{evolving associations} between them as the trajectory grows. Appendix~\ref{app:failure-cases} provides representative cases.

\paragraph{Finding 4: \textsc{PraxisMem} improves persistent memory use by learning both memory content and organization.}
We next examine whether \textsc{PraxisMem} improves not only memory content, but also the organization that maintains personal--procedural associations over time. As shown in Figure~\ref{fig:memrsd-training}, content-only GRPO learns useful memory updates but remains weak at placing them consistently, with a final-20-step Correct-location Recall of 41.1\%. Adding the location-aware reward raises this score to 47.2\%, while criterion-routed OPSD further improves it to 53.3\%. Full \textsc{PraxisMem}, which further localizes unresolved defects to the chunks where they persist, reaches \textbf{61.3\%}, showing a progression from content correctness toward better maintained memory organization.

These training-time improvements also transfer to persistent task performance. On the fixed eight-trajectory ablation subset in Table~\ref{tab:memrsd-ablation}, Combined Completion increases from 66.2\% under content-only GRPO to \textbf{83.4\%} with full \textsc{PraxisMem}, while Interdependent Completion rises more substantially from 54.9\% to \textbf{82.1\%}; Avg SR similarly improves from 23.5\% to \textbf{43.1\%}. Together, these results show that persistent memory management benefits from learning not only \emph{what} personal and procedural content to preserve, but also \emph{where and how} to organize it so that the evolving association structure $\mathcal I$ remains recoverable as interaction history grows.

\section{Conclusion}
\label{sec:conclusion}

We introduced \textsc{PraxisMemBench}, a harness-based benchmark for evaluating persistent personal agents across daily-life and work trajectories. \textsc{PraxisMemBench} focuses on the joint maintenance of personal and procedural memory, motivated by the observation that \textbf{procedures or skills in personal assistance are inherently user-conditioned}. Our experiments show that existing methods increasingly struggle as interaction history grows and their relations must be preserved across repeated updates. Failure analysis further reveals a \textbf{continual-maintenance} challenge, where memory errors shift from simple access failures toward fragmented or competing personal--procedural relations. To address this challenge, we introduced \textsc{PraxisMem}, which learns both memory content and organization through location-aware GRPO and dynamic residual OPSD. We hope \textsc{PraxisMemBench} provides a useful testbed for developing memory systems that can support reliable, personalized, and persistent agent behavior over long-term interaction.

\subsubsection*{AI Use Statement}
Generative AI tools were used in several parts of this work. GPT-5.5 was used
during benchmark construction to expand persona seeds into structured
profiles and to synthesize personal change events, as described in
Appendix~\ref{app:construction-pipeline}. Coding agents, including Codex, were
used to assist with software implementation, debugging, experiment tooling,
and the preparation and refinement of benchmark instances. Codex was also
used to improve the readability and presentation of the manuscript. All
AI-assisted benchmark content, manuscript text, factual claims, citations,
code, experimental outputs, and interpretations were reviewed and verified
by the authors or the independent validation procedures described in the
paper. The authors made all final decisions regarding the research design,
experimental protocol, analysis, and conclusions, and take full
responsibility for the contents of this work.

\subsubsection*{Ethics Statement}
PraxisMemBench is constructed from fictional personas and synthetic interaction
trajectories and does not contain real user histories or personally
identifiable information. All tool interactions are executed in isolated
benchmark applications; no real accounts, payments, communications, or other
external actions are involved. Synthetic personas and LLM-generated events
may inherit cultural, demographic, or behavioral biases and should not be
interpreted as representative of real users. Although PraxisMemBench is intended
for evaluating memory systems, persistent personal assistants may create
privacy and security risks by retaining sensitive information, relying on
outdated user states, or applying remembered procedures in inappropriate
contexts. Real-world deployments should therefore provide appropriate access
controls, user-visible retention and deletion mechanisms, consent-aware data
handling, and safeguards for consequential actions. The authors have read
and adhered to the ICLR Code of Ethics.

\subsubsection*{Reproducibility Statement}
We document the PraxisMemBench benchmark formulation and construction overview in
Section~\ref{sec:benchmark}, with detailed domain and SOP grounding, sequential
trajectory construction, persona and event synthesis, task and gold
instantiation, and independent validation in
Appendix~\ref{app:construction}. The chronological evaluation protocol,
task-family metrics, programmatic verifiers, semantic judging procedure, and
judge-agreement audit are described in
Appendix~\ref{app:evaluation-details}, while all construction and evaluation
prompts are provided in Appendix~\ref{app:prompts}. Model and harness versions,
context and memory budgets, inference settings, hardware, and training
configurations are reported in Appendix~\ref{app:configurations}, with further
details of \textsc{PraxisMem} in Appendix~\ref{app:memrsd}. We will release the
fixed benchmark splits, application fixtures and SOPs, gold annotations and
verifiers, construction and evaluation code, baseline configurations, prompts,
and \textsc{PraxisMem} training code and checkpoint to support replication and
future extensions.

\bibliography{references}

@article{ge2024personahub,
  author       = {Xin Chan and
                  Xiaoyang Wang and
                  Dian Yu and
                  Haitao Mi and
                  Dong Yu},
  title        = {Scaling Synthetic Data Creation with 1,000,000,000 Personas},
  journal      = {CoRR},
  volume       = {abs/2406.20094},
  year         = {2024},
  url          = {https://doi.org/10.48550/arXiv.2406.20094},
  doi          = {10.48550/ARXIV.2406.20094},
  eprinttype   = {arXiv},
  eprint       = {2406.20094},
  bibsource    = {dblp computer science bibliography, https://dblp.org}
}

@article{shao2024deepseekmath,
  author       = {Zhihong Shao and
                  Peiyi Wang and
                  Qihao Zhu and
                  Runxin Xu and
                  Junxiao Song and
                  Mingchuan Zhang and
                  Y. K. Li and
                  Y. Wu and
                  Daya Guo},
  title        = {DeepSeekMath: Pushing the Limits of Mathematical Reasoning in Open
                  Language Models},
  journal      = {CoRR},
  volume       = {abs/2402.03300},
  year         = {2024},
  url          = {https://doi.org/10.48550/arXiv.2402.03300},
  doi          = {10.48550/ARXIV.2402.03300},
  eprinttype   = {arXiv},
  eprint       = {2402.03300},
  bibsource    = {dblp computer science bibliography, https://dblp.org}
}

@article{zhao2026selfdistilled,
  author       = {Siyan Zhao and
                  Zhihui Xie and
                  Mengchen Liu and
                  Jing Huang and
                  Guan Pang and
                  Feiyu Chen and
                  Aditya Grover},
  title        = {Self-Distilled Reasoner: On-Policy Self-Distillation for Large Language
                  Models},
  journal      = {CoRR},
  volume       = {abs/2601.18734},
  year         = {2026},
  url          = {https://doi.org/10.48550/arXiv.2601.18734},
  doi          = {10.48550/ARXIV.2601.18734},
  eprinttype   = {arXiv},
  eprint       = {2601.18734},
  bibsource    = {dblp computer science bibliography, https://dblp.org}
}

@inproceedings{maharana2024locomo,
  author       = {Adyasha Maharana and
                  Dong{-}Ho Lee and
                  Sergey Tulyakov and
                  Mohit Bansal and
                  Francesco Barbieri and
                  Yuwei Fang},
  editor       = {Lun{-}Wei Ku and
                  Andre Martins and
                  Vivek Srikumar},
  title        = {Evaluating Very Long-Term Conversational Memory of {LLM} Agents},
  booktitle    = {Proceedings of the 62nd Annual Meeting of the Association for Computational
                  Linguistics (Volume 1: Long Papers), {ACL} 2024, Bangkok, Thailand,
                  August 11-16, 2024},
  pages        = {13851--13870},
  publisher    = {Association for Computational Linguistics},
  year         = {2024},
  url          = {https://doi.org/10.18653/v1/2024.acl-long.747},
  doi          = {10.18653/V1/2024.ACL-LONG.747},
  bibsource    = {dblp computer science bibliography, https://dblp.org}
}

@inproceedings{wu2025longmemeval,
  author       = {Di Wu and
                  Hongwei Wang and
                  Wenhao Yu and
                  Yuwei Zhang and
                  Kai{-}Wei Chang and
                  Dong Yu},
  title        = {LongMemEval: Benchmarking Chat Assistants on Long-Term Interactive
                  Memory},
  booktitle    = {The Thirteenth International Conference on Learning Representations,
                  {ICLR} 2025, Singapore, April 24-28, 2025},
  publisher    = {OpenReview.net},
  year         = {2025},
  url          = {https://openreview.net/forum?id=pZiyCaVuti},
  bibsource    = {dblp computer science bibliography, https://dblp.org}
}

@inproceedings{tan2025membench,
  author       = {Haoran Tan and
                  Zeyu Zhang and
                  Chen Ma and
                  Xu Chen and
                  Quanyu Dai and
                  Zhenhua Dong},
  editor       = {Wanxiang Che and
                  Joyce Nabende and
                  Ekaterina Shutova and
                  Mohammad Taher Pilehvar},
  title        = {MemBench: Towards More Comprehensive Evaluation on the Memory of LLM-based
                  Agents},
  booktitle    = {Findings of the Association for Computational Linguistics, {ACL} 2025,
                  Vienna, Austria, July 27 - August 1, 2025},
  series       = {Findings of {ACL}},
  volume       = {{ACL} 2025},
  pages        = {19336--19352},
  publisher    = {Association for Computational Linguistics},
  year         = {2025},
  url          = {https://doi.org/10.18653/v1/2025.findings-acl.989},
  doi          = {10.18653/V1/2025.FINDINGS-ACL.989},
  bibsource    = {dblp computer science bibliography, https://dblp.org}
}

@article{jiang2025personamemv2,
  author       = {Bowen Jiang and
                  Yuan Yuan and
                  Maohao Shen and
                  Zhuoqun Hao and
                  Zhangchen Xu and
                  Zichen Chen and
                  Ziyi Liu and
                  Anvesh Rao Vijjini and
                  Jiashu He and
                  Hanchao Yu and
                  Radha Poovendran and
                  Gregory W. Wornell and
                  Lyle H. Ungar and
                  Dan Roth and
                  Sihao Chen and
                  Camillo Jose Taylor},
  title        = {PersonaMem-v2: Towards Personalized Intelligence via Learning Implicit
                  User Personas and Agentic Memory},
  journal      = {CoRR},
  volume       = {abs/2512.06688},
  year         = {2025},
  url          = {https://doi.org/10.48550/arXiv.2512.06688},
  doi          = {10.48550/ARXIV.2512.06688},
  eprinttype   = {arXiv},
  eprint       = {2512.06688},
  bibsource    = {dblp computer science bibliography, https://dblp.org}
}

@article{he2026memoryarena,
  author       = {Zexue He and
                  Yu Wang and
                  Churan Zhi and
                  Yuanzhe Hu and
                  Tzu{-}Ping Chen and
                  Lang Yin and
                  Ze Chen and
                  Tong Arthur Wu and
                  Siru Ouyang and
                  Zihan Wang and
                  Jiaxin Pei and
                  Julian J. McAuley and
                  Yejin Choi and
                  Alex Pentland},
  title        = {MemoryArena: Benchmarking Agent Memory in Interdependent Multi-Session
                  Agentic Tasks},
  journal      = {CoRR},
  volume       = {abs/2602.16313},
  year         = {2026},
  url          = {https://doi.org/10.48550/arXiv.2602.16313},
  doi          = {10.48550/ARXIV.2602.16313},
  eprinttype   = {arXiv},
  eprint       = {2602.16313},
  bibsource    = {dblp computer science bibliography, https://dblp.org}
}

@article{ji2026clawarena,
  author       = {Haonian Ji and
                  Kaiwen Xiong and
                  Siwei Han and
                  Peng Xia and
                  Shi Qiu and
                  Yiyang Zhou and
                  Jiaqi Liu and
                  Jinlong Li and
                  Bingzhou Li and
                  Zeyu Zheng and
                  Cihang Xie and
                  Huaxiu Yao},
  title        = {ClawArena: Benchmarking {AI} Agents in Evolving Information Environments},
  journal      = {CoRR},
  volume       = {abs/2604.04202},
  year         = {2026},
  url          = {https://doi.org/10.48550/arXiv.2604.04202},
  doi          = {10.48550/ARXIV.2604.04202},
  eprinttype   = {arXiv},
  eprint       = {2604.04202},
  bibsource    = {dblp computer science bibliography, https://dblp.org}
}

@article{zhu2026semaclaw,
  author       = {Ningyan Zhu and
                  Huacan Wang and
                  Jie Zhou and
                  Feiyu Chen and
                  Shuo Zhang and
                  Ge Chen and
                  Chen Liu and
                  Jiarou Wu and
                  Wangyi Chen and
                  Xiaofeng Mou and
                  Yi Xu},
  title        = {SemaClaw: {A} Step Towards General-Purpose Personal {AI} Agents through
                  Harness Engineering},
  journal      = {CoRR},
  volume       = {abs/2604.11548},
  year         = {2026},
  url          = {https://doi.org/10.48550/arXiv.2604.11548},
  doi          = {10.48550/ARXIV.2604.11548},
  eprinttype   = {arXiv},
  eprint       = {2604.11548},
  bibsource    = {dblp computer science bibliography, https://dblp.org}
}

@article{workbuddy2026benchmark,
  author       = {Siqi Cai and
                  Shaopeng Chen and
                  Xiang Fei and
                  Yong Mao and
                  Zihan Xu and
                  Zhiheng Lyu and
                  Zhijian Shao and
                  Yuchen Shi and
                  Shuwen Zhang and
                  Chaofan Qiu and
                  Linjie Che and
                  Xiaoxi Zhao and
                  Feng Wu and
                  Kai Zhang and
                  Chaofan Zhu and
                  Yubin Qi and
                  Xiaoyun Liang and
                  Peijie Dong and
                  Yunhao Zhang and
                  Yuanjie Zhu and
                  Ling Jiang and
                  Xianjun Zhang and
                  Zhehang Chu and
                  Anyuan Sang and
                  Zhen Feng and
                  Sen Nie and
                  Shi Wu and
                  Yuanzhen Xu and
                  Xin Li and
                  Ning Yang and
                  Zhiqiang Dong and
                  Hande Dong and
                  Qiang Lin and
                  Yi Liu and
                  Yunsheng Wu and
                  Ke Li and
                  Xing Sun},
  title        = {Tencent WorkBuddy Bench: {A} Multi-Domain Coding-Agent Benchmark with
                  Contamination-Resistant Task Construction},
  journal      = {CoRR},
  volume       = {abs/2607.20911},
  year         = {2026},
  url          = {https://doi.org/10.48550/arXiv.2607.20911},
  doi          = {10.48550/ARXIV.2607.20911},
  eprinttype   = {arXiv},
  eprint       = {2607.20911},
  bibsource    = {dblp computer science bibliography, https://dblp.org}
}

@article{chen2026vitabench,
  author       = {Yuxin Chen and
                  Yi Zhang and
                  Zhengzhou Cai and
                  Yaorui Shi and
                  Zhiyuan Yao and
                  Chenhang Cui and
                  Jingnan Zheng and
                  Yaqi Huo and
                  Xi Su and
                  Qi Gu and
                  Xunliang Cai and
                  Xiang Wang and
                  An Zhang and
                  Tat{-}Seng Chua},
  title        = {VitaBench 2.0: Evaluating Personalized and Proactive Agents in Long-Term
                  User Interactions},
  journal      = {CoRR},
  volume       = {abs/2605.27141},
  year         = {2026},
  url          = {https://doi.org/10.48550/arXiv.2605.27141},
  doi          = {10.48550/ARXIV.2605.27141},
  eprinttype   = {arXiv},
  eprint       = {2605.27141},
  bibsource    = {dblp computer science bibliography, https://dblp.org}
}

@article{dou2026clbenchlife,
  author       = {Shihan Dou and
                  Yujiong Shen and
                  Chenhao Huang and
                  Junjie Ye and
                  Jiayi Chen and
                  Junzhe Wang and
                  Qianyu He and
                  Shichun Liu and
                  Changze Lv and
                  Jiahang Lin and
                  Jiazheng Zhang and
                  Ming Zhang and
                  Shaofan Liu and
                  Tao Ji and
                  Zhangyue Yin and
                  Cheng Zhang and
                  Huaibing Xie and
                  Jianglu Hu and
                  Jingcheng Deng and
                  Lincheng Li and
                  Minda Hu and
                  Shaolei Wang and
                  Syrus Zhao and
                  Weichao Wang and
                  Yan Lei and
                  Yang Liu and
                  Yanling Xiao and
                  Yiting Liu and
                  Zenan Xu and
                  Zhen Guo and
                  Ziliang Zhao and
                  Pluto Zhou and
                  Tao Gui and
                  Qi Zhang and
                  Xuanjing Huang and
                  Yu{-}Gang Jiang and
                  Di Wang and
                  Shunyu Yao},
  title        = {CL-bench Life: Can Language Models Learn from Real-Life Context?},
  journal      = {CoRR},
  volume       = {abs/2604.27043},
  year         = {2026},
  url          = {https://doi.org/10.48550/arXiv.2604.27043},
  doi          = {10.48550/ARXIV.2604.27043},
  eprinttype   = {arXiv},
  eprint       = {2604.27043},
  bibsource    = {dblp computer science bibliography, https://dblp.org}
}

@inproceedings{wang2024workflow,
  author       = {Zora Zhiruo Wang and
                  Jiayuan Mao and
                  Daniel Fried and
                  Graham Neubig},
  editor       = {Aarti Singh and
                  Maryam Fazel and
                  Daniel Hsu and
                  Simon Lacoste{-}Julien and
                  Felix Berkenkamp and
                  Tegan Maharaj and
                  Kiri Wagstaff and
                  Jerry Zhu},
  title        = {Agent Workflow Memory},
  booktitle    = {Forty-second International Conference on Machine Learning, {ICML}
                  2025, Vancouver, BC, Canada, July 13-19, 2025},
  series       = {Proceedings of Machine Learning Research},
  volume       = {267},
  publisher    = {{PMLR} / OpenReview.net},
  year         = {2025},
  url          = {https://proceedings.mlr.press/v267/wang25bx.html},
  bibsource    = {dblp computer science bibliography, https://dblp.org}
}

@inproceedings{shen2026mem2actbench,
  author       = {Yiting Shen and
                  Kun Li and
                  Wei Zhou and
                  Songlin Hu},
  editor       = {Maria Liakata and
                  Viviane P. Moreira and
                  Jiajun Zhang and
                  David Jurgens},
  title        = {Mem2ActBench: {A} Benchmark for Evaluating Long-Term Memory Utilization
                  in Task-Oriented Autonomous Agents},
  booktitle    = {Proceedings of the 64th Annual Meeting of the Association for Computational
                  Linguistics (Volume 1: Long Papers), {ACL} 2026, San Diego, California,
                  United States, July 2-7, 2026},
  pages        = {8173--8190},
  publisher    = {Association for Computational Linguistics},
  year         = {2026},
  url          = {https://doi.org/10.18653/v1/2026.acl-long.370},
  doi          = {10.18653/V1/2026.ACL-LONG.370},
  bibsource    = {dblp computer science bibliography, https://dblp.org}
}

@article{li2026skillsbench,
  author       = {Xiangyi Li and
                  Wenbo Chen and
                  Yimin Liu and
                  Shenghan Zheng and
                  Xiaokun Chen and
                  Yifeng He and
                  Yubo Li and
                  Bingran You and
                  Haotian Shen and
                  Jiankai Sun and
                  Shuyi Wang and
                  Qunhong Zeng and
                  Di Wang and
                  Xuandong Zhao and
                  Yuanli Wang and
                  Roey Ben Chaim and
                  Zonglin Di and
                  Yipeng Gao and
                  Junwei He and
                  Yizhuo He and
                  Liqiang Jing and
                  Luyang Kong and
                  Xin Lan and
                  Jiachen Li and
                  Songlin Li and
                  Yijiang Li and
                  Yueqian Lin and
                  Xinyi Liu and
                  Xuanqing Liu and
                  Haoran Lyu and
                  Ze Ma and
                  Bowei Wang and
                  Runhui Wang and
                  Tianfu Wang and
                  Wengao Ye and
                  Yue Zhang and
                  Hanwen Xing and
                  Yiqi Xue and
                  Steven Dillmann and
                  Han{-}Chung Lee},
  title        = {SkillsBench: Benchmarking How Well Agent Skills Work Across Diverse
                  Tasks},
  journal      = {CoRR},
  volume       = {abs/2602.12670},
  year         = {2026},
  url          = {https://doi.org/10.48550/arXiv.2602.12670},
  doi          = {10.48550/ARXIV.2602.12670},
  eprinttype   = {arXiv},
  eprint       = {2602.12670},
  bibsource    = {dblp computer science bibliography, https://dblp.org}
}

@article{packer2023memgpt,
  author       = {Charles Packer and
                  Vivian Fang and
                  Shishir G. Patil and
                  Kevin Lin and
                  Sarah Wooders and
                  Joseph E. Gonzalez},
  title        = {MemGPT: Towards LLMs as Operating Systems},
  journal      = {CoRR},
  volume       = {abs/2310.08560},
  year         = {2023},
  url          = {https://doi.org/10.48550/arXiv.2310.08560},
  doi          = {10.48550/ARXIV.2310.08560},
  eprinttype   = {arXiv},
  eprint       = {2310.08560},
  bibsource    = {dblp computer science bibliography, https://dblp.org}
}

@inproceedings{chhikara2025mem0,
  author       = {Prateek Chhikara and
                  Dev Khant and
                  Saket Aryan and
                  Taranjeet Singh and
                  Deshraj Yadav},
  editor       = {In{\^{e}}s Lynce and
                  Nello Murano and
                  Mauro Vallati and
                  Serena Villata and
                  Federico Chesani and
                  Michela Milano and
                  Andrea Omicini and
                  Mehdi Dastani},
  title        = {Mem0: Building Production-Ready {AI} Agents with Scalable Long-Term
                  Memory},
  booktitle    = {{ECAI} 2025 - 28th European Conference on Artificial Intelligence,
                  25-30 October 2025, Bologna, Italy - Including 14th Conference on
                  Prestigious Applications of Intelligent Systems {(PAIS} 2025)},
  series       = {Frontiers in Artificial Intelligence and Applications},
  volume       = {413},
  pages        = {2993--3000},
  publisher    = {{IOS} Press},
  year         = {2025},
  url          = {https://doi.org/10.3233/FAIA251160},
  doi          = {10.3233/FAIA251160},
  bibsource    = {dblp computer science bibliography, https://dblp.org}
}

@article{rasmussen2025zep,
  author       = {Preston Rasmussen and
                  Pavlo Paliychuk and
                  Travis Beauvais and
                  Jack Ryan and
                  Daniel Chalef},
  title        = {Zep: {A} Temporal Knowledge Graph Architecture for Agent Memory},
  journal      = {CoRR},
  volume       = {abs/2501.13956},
  year         = {2025},
  url          = {https://doi.org/10.48550/arXiv.2501.13956},
  doi          = {10.48550/ARXIV.2501.13956},
  eprinttype   = {arXiv},
  eprint       = {2501.13956},
  bibsource    = {dblp computer science bibliography, https://dblp.org}
}

@inproceedings{xu2025amem,
  author       = {Wujiang Xu and
                  Zujie Liang and
                  Kai Mei and
                  Hang Gao and
                  Juntao Tan and
                  Yongfeng Zhang},
  editor       = {Danielle Belgrave and
                  Cheng Zhang and
                  Laura N. Montoya and
                  Hsuan{-}Tien Lin and
                  Razvan Pascanu and
                  Piotr Koniusz and
                  Marzyeh Ghassemi and
                  Nancy Chen and
                  Iv{\'{a}}n Vladimir Meza Ru{\'{\i}}z and
                  Arturo Loaiza{-}Bonilla},
  title        = {A-Mem: Agentic Memory for {LLM} Agents},
  booktitle    = {Advances in Neural Information Processing Systems 38: Annual Conference
                  on Neural Information Processing Systems 2025, NeurIPS 2025, San Diego,
                  CA, USA, December 2-7, 2025 / Mexico City, Mexico, November 30 - December
                  5, 2025},
  year         = {2025},
  url          = {http://papers.nips.cc/paper\_files/paper/2025/hash/19909c36f51abc4856b4560aff3d36d6-Abstract-Conference.html},
  bibsource    = {dblp computer science bibliography, https://dblp.org}
}

@inproceedings{cao2026reme,
  author       = {Zouying Cao and
                  Jiaji Deng and
                  Li Yu and
                  Weikang Zhou and
                  Zhaoyang Liu and
                  Bolin Ding and
                  Hai Zhao},
  editor       = {Maria Liakata and
                  Viviane P. Moreira and
                  Jiajun Zhang and
                  David Jurgens},
  title        = {Remember Me, Refine Me: {A} Dynamic Procedural Memory Framework for
                  Experience-Driven Agent Evolution},
  booktitle    = {Findings of the Association for Computational Linguistics, {ACL} 2026,
                  San Diego, California, United States, July 2-7, 2026},
  pages        = {16803--16822},
  publisher    = {Association for Computational Linguistics},
  year         = {2026},
  url          = {https://doi.org/10.18653/v1/2026.findings-acl.829},
  doi          = {10.18653/V1/2026.FINDINGS-ACL.829},
  bibsource    = {dblp computer science bibliography, https://dblp.org}
}

@article{li2025memos,
  author       = {Zhiyu Li and
                  Shichao Song and
                  Chenyang Xi and
                  Hanyu Wang and
                  Chen Tang and
                  Simin Niu and
                  Ding Chen and
                  Jiawei Yang and
                  Chunyu Li and
                  Qingchen Yu and
                  Jihao Zhao and
                  Yezhaohui Wang and
                  Peng Liu and
                  Zehao Lin and
                  Pengyuan Wang and
                  Jiahao Huo and
                  Tianyi Chen and
                  Kai Chen and
                  Kehang Li and
                  Zhen Tao and
                  Junpeng Ren and
                  Huayi Lai and
                  Hao Wu and
                  Bo Tang and
                  Zhenren Wang and
                  Zhaoxin Fan and
                  Ningyu Zhang and
                  Linfeng Zhang and
                  Junchi Yan and
                  Mingchuan Yang and
                  Tong Xu and
                  Wei Xu and
                  Huajun Chen and
                  Haofeng Wang and
                  Hongkang Yang and
                  Wentao Zhang and
                  Zhi{-}Qin John Xu and
                  Siheng Chen and
                  Feiyu Xiong},
  title        = {MemOS: {A} Memory {OS} for {AI} System},
  journal      = {CoRR},
  volume       = {abs/2507.03724},
  year         = {2025},
  url          = {https://doi.org/10.48550/arXiv.2507.03724},
  doi          = {10.48550/ARXIV.2507.03724},
  eprinttype   = {arXiv},
  eprint       = {2507.03724},
  bibsource    = {dblp computer science bibliography, https://dblp.org}
}

@inproceedings{yan2025memoryr1,
  author       = {Sikuan Yan and
                  Xiufeng Yang and
                  Zuchao Huang and
                  Ercong Nie and
                  Zifeng Ding and
                  Zonggen Li and
                  Xiaowen Ma and
                  Jinhe Bi and
                  Kristian Kersting and
                  Jeff Z. Pan and
                  Hinrich Sch{\"{u}}tze and
                  Volker Tresp and
                  Yunpu Ma},
  editor       = {Maria Liakata and
                  Viviane P. Moreira and
                  Jiajun Zhang and
                  David Jurgens},
  title        = {Memory-R1: Enhancing Large Language Model Agents to Manage and Utilize
                  Memories via Reinforcement Learning},
  booktitle    = {Proceedings of the 64th Annual Meeting of the Association for Computational
                  Linguistics (Volume 1: Long Papers), {ACL} 2026, San Diego, California,
                  United States, July 2-7, 2026},
  pages        = {12805--12825},
  publisher    = {Association for Computational Linguistics},
  year         = {2026},
  url          = {https://doi.org/10.18653/v1/2026.acl-long.583},
  doi          = {10.18653/V1/2026.ACL-LONG.583},
  bibsource    = {dblp computer science bibliography, https://dblp.org}
}

@article{zhang2026deltamem,
  author       = {Qi Zhang and
                  Shen Huang and
                  Chu Liu and
                  Shouqing Yang and
                  Junbo Zhao and
                  Haobo Wang and
                  Pengjun Xie},
  title        = {DeltaMem: Towards Agentic Memory Management via Reinforcement Learning},
  journal      = {CoRR},
  volume       = {abs/2604.01560},
  year         = {2026},
  url          = {https://doi.org/10.48550/arXiv.2604.01560},
  doi          = {10.48550/ARXIV.2604.01560},
  eprinttype   = {arXiv},
  eprint       = {2604.01560},
  bibsource    = {dblp computer science bibliography, https://dblp.org}
}

@article{yan2026memoryr2,
  author       = {Sikuan Yan and
                  Ahmed Bahloul and
                  Ercong Nie and
                  Susanna Schwarzmann and
                  Riccardo Trivisonno and
                  Volker Tresp and
                  Yunpu Ma},
  title        = {Memory-R2: Fair Credit Assignment for Long-Horizon Memory-Augmented
                  {LLM} Agents},
  journal      = {CoRR},
  volume       = {abs/2605.21768},
  year         = {2026},
  url          = {https://doi.org/10.48550/arXiv.2605.21768},
  doi          = {10.48550/ARXIV.2605.21768},
  eprinttype   = {arXiv},
  eprint       = {2605.21768},
  bibsource    = {dblp computer science bibliography, https://dblp.org}
}

@inproceedings{cheng2026amemgym,
  author       = {Cheng Jiayang and
                  Dongyu Ru and
                  Lin Qiu and
                  Yiyang Li and
                  Xuezhi Cao and
                  Yangqiu Song and
                  Xunliang Cai},
  editor       = {Carl Vondrick and
                  Bharath Hariharan and
                  Colin Raffel and
                  Lerrel Pinto and
                  Diyi Yang and
                  Aleksandra Faust},
  title        = {AMemGym: Interactive Memory Benchmarking for Assistants in Long-Horizon
                  Conversations},
  booktitle    = {The Fourteenth International Conference on Learning Representations,
                  {ICLR} 2026, Rio de Janeiro, Brazil, April 23-27, 2026},
  publisher    = {proceedings.iclr.cc},
  year         = {2026},
  url          = {https://proceedings.iclr.cc/paper\_files/paper/2026/hash/0856bc553d3e3b9827e5140d0ad3bf8d-Abstract-Conference.html},
  bibsource    = {dblp computer science bibliography, https://dblp.org}
}

@misc{alibaba2026qwenpricing,
  author       = {{Alibaba Cloud}},
  title        = {Model Studio Model Pricing},
  year         = {2026},
  howpublished = {\url{https://www.alibabacloud.com/help/en/model-studio/model-pricing}},
  note         = {Accessed: 2026-09-03}
}
\bibliographystyle{iclr2027_conference}

\appendix
\section{Related Work}
\label{sec:related-work}

\subsection{Long-Term Memory and Agent Benchmarks}

Long-term memory benchmarks have progressed from question answering over fixed
interaction histories to memory-conditioned agent execution. LoCoMo and
LongMemEval test retrieval and reasoning over multi-session conversations
\citep{maharana2024locomo,wu2025longmemeval}, while Mem2ActBench adds
task-oriented interaction traces \citep{shen2026mem2actbench}. More recent
benchmarks place memory inside executable loops: MemoryArena constructs
causally dependent tasks across sessions, and VitaBench 2.0 evaluates how
long-term user information affects tool-mediated decisions
\citep{he2026memoryarena,chen2026vitabench}. These settings move beyond
post-hoc recall, but vary in whether they use a general agent harness, natural
requests, and agent-created state as an explicit dependency of later tasks.

A complementary distinction concerns \emph{what} the agent must remember.
Personalization benchmarks emphasize user facts, preferences, and their
evolution, whereas procedural evaluations represent reusable ways of acting as
workflows or skills \citep{chen2026vitabench,wang2024workflow,li2026skillsbench}.
Harness benchmarks such as ClawArena and WorkBuddy Bench broaden executable
coverage across evolving environments and application domains
\citep{ji2026clawarena,workbuddy2026benchmark}, yet do not systematically test
both memory forms. For persistent personal agents, however, the challenge is to \textbf{continuously maintain personal and procedural memory together over long-running interaction in a real harness}, where both evolve and must remain usable for later tasks.

\subsection{Memory Systems and Learning-Based Management}

Training-free memory systems externalize long histories through context
management, retrieval, and structured updates. MemGPT introduces explicit
movement between in-context and external memory, while Mem0 extracts compact
memories and retrieves relevant entries for later interactions
\citep{packer2023memgpt,chhikara2025mem0}. Procedural knowledge is commonly
stored separately as induced workflows or portable skill packages
\citep{wang2024workflow,li2026skillsbench}. This division provides practical
interfaces for user-specific knowledge and reusable action patterns, but gives
limited support for their joint formation and revision when a user's changing
constraints alter how a procedure should be executed.

Recent work learns memory-management policies with reinforcement
learning. Memory-R1 optimizes memory operations and answer-time memory use from
downstream QA rewards \citep{yan2025memoryr1}; DeltaMem learns end-to-end memory
state transitions with state-level rewards \citep{zhang2026deltamem}; and
Memory-R2 introduces global--local optimization to improve credit assignment
across long session sequences \citep{yan2026memoryr2}. Their rewards primarily encourage correct memory writing and retrieval, but provide little supervision for \textbf{maintaining a well-organized store over long trajectories}. Our \textsc{PraxisMem} addresses this maintenance problem with location-aware GRPO and dynamic residual distillation against a well-organized target memory, selectively providing token-level guidance to the memory chunks responsible for unmet criteria.

\section{Additional Details of \textsc{PraxisMem}}
\label{app:memrsd}

Persistent memory management requires more than writing correct information once. As interaction history accumulates, the manager must \textbf{continuously revise memory} so that relevant information remains \emph{organized and retrievable} over long trajectories. \textsc{PraxisMem} therefore uses location-aware GRPO as its primary outcome objective, rewarding not only whether required content is written, but also whether it is maintained in the \textbf{appropriate part} of the persistent store.

GRPO alone provides limited learning signal when every rollout in a group misses the same requirement. We use OPSD to recover this group-level exploration failure. \textsc{PraxisMem} constructs a dynamic semantic residual from the student's realized memory state and the contemporaneous target state. This gives the privileged teacher information about \textbf{what the current rollout still needs to add, revise, move, or remove}. The residual is routed at both the criterion and chunk levels: supervision is applied only to unresolved criteria and only where the corresponding defect persists.

SFT would require complete, high-quality edit trajectories from each current store to its target state. Automatically deriving such trajectories from target states cannot ensure reliable operation-level supervision, while expert annotation at this scale is costly. Multiple add, update, and delete sequences can also yield the same valid target state, so selecting a single canonical trajectory imposes one teacher's action choices. SFT on such trajectories conditions later actions on gold intermediate states. In contrast, \textsc{PraxisMem} trains on the states reached by its own rollouts: the dynamic residual identifies missing, misplaced, outdated, conflicting, or redundant information and guides the operations that can repair it after earlier updates were incomplete or incorrect.

\begin{table}[t]
    \centering
    \small
    \caption{Persistent Markdown state used by \textsc{PraxisMem}.}
    \label{tab:markdown-state}
    \setlength{\tabcolsep}{5pt}
    \begin{tabular}{@{}p{0.33\linewidth}p{0.61\linewidth}@{}}
        \toprule
        Location & Maintained content \\
        \midrule
        \texttt{USER.md} & Stable user attributes and current preferences or constraints. \\
        \texttt{MEMORY.md} & Topic index linking current personal state to dated events. \\
        \texttt{events/*.md} & Time-indexed changes with optional task and skill links. \\
        \texttt{skills/<domain>/SKILL.md}
        & Domain rules and reusable procedural workflows. \\
        Raw interaction archive & Source evidence retained for targeted search and inspection. \\
        \bottomrule
    \end{tabular}
\end{table}

\subsection{Persistent Markdown State and Notation}
\label{app:memrsd-state}

Consider a persistent agent serving the same user across sessions. At session $n$, the user issues request $q_n$, the fixed task-agent policy $\pi$ produces interaction trace $H_n$, and the learned memory-manager policy $f_\theta$ maintains memory state $M_n$. The manager is trained for two tasks: \emph{Update}, which incorporates new interaction evidence into the persistent store, and \emph{Retrieval}, which extracts a compact memory context for a memory-dependent task.

The memory state comprises personal memory, procedural memory, and their association structure, 
\begin{equation}
M_n=\bigl(M_n^{\mathrm{pers}},M_n^{\mathrm{proc}},\mathcal I_n\bigr),
\end{equation}
and is realized as a shared directory of structured Markdown files. Table~\ref{tab:markdown-state} summarizes the placement contract. Personal and procedural memories are maintained by the same manager, while their organization, indexing, and cross-references induce the association structure $\mathcal I_n$ that supports subsequent maintenance and joint recovery. Importantly, $\mathcal I_n$ is an internal property of the maintained store and need not be explicitly serialized in the retrieval output. The Markdown schema is one implementation of this structured state, not a component of PraxisMemBench itself. PraxisMem only assumes that training-time reference states can be constructed under the chosen memory representation; the representation itself need not be unique, and alternative organizations can be used with analogous state-level targets.

The files form an indirect personal--event--procedure relation. For each
topic, \texttt{MEMORY.md} links the current value in \texttt{USER.md} to its
dated history in \texttt{events/*.md}. When an interaction establishes or
revises a reusable procedure, the corresponding event entry records the
associated task and \texttt{SKILL.md} workflow. Thus, \texttt{MEMORY.md}
contains no direct skill links: procedural memory is connected to personal
state through the event that established their relation.

Let $\mathcal{U}(M)$ denote the structured units parsed from store $M$. Each unit $u$ has a coarse type $c(u)$, textual content $\operatorname{text}(u)$, and location $\operatorname{loc}(u)$. The types include user information, memory-index entries, events, domain rules, and reusable workflows. For \texttt{USER.md} and \texttt{MEMORY.md}, location is defined by the file and second-level heading; procedural units are located by their domain-specific \texttt{SKILL.md} path. Event units are treated as location-neutral because their dates and headings primarily preserve provenance rather than determine future access.

For an Update sample, the interaction trace $H_n$ is divided into $K$ ordered chunks $\chi_{n,1:K}$, and each rollout starts from the canonical pre-session store $M_n^\star$. This canonical initialization is used only to construct controlled memory-manager training rollouts. During benchmark evaluation, the realized memory store and application state instead evolve freely across the complete trajectory and are never restored to their canonical pre-session states. We sample a group of $B$ manager rollouts from the frozen pre-update policy:
\begin{equation}
\mathcal{Y}_n=\{Y_i\}_{i=1}^{B},
\qquad
Y_i\sim
f_{\theta^-}
\bigl(\cdot\mid M_n^\star,\chi_{n,1:K}\bigr).
\label{app:eq:manager-rollouts}
\end{equation}
If $a_{i,k}$ denotes the memory operations produced by rollout $i$ after chunk $k$, its realized store evolves as
\begin{equation}
M_{i,1}^{-}=M_n^\star,
\qquad
M_{i,k}^{+}=T(M_{i,k}^{-},a_{i,k}),
\qquad
M_{i,k+1}^{-}=M_{i,k}^{+},
\label{app:eq:state-transition}
\end{equation}
where $T$ executes the memory operations in an isolated store. We denote the canonical memory state after chunk $k$ by $M_{n,k}^\star$, with $M_{n,0}^\star=M_n^\star$ and $M_{n,K}^\star=M_{n+1}^\star$. These intermediate targets specify a reference state under the chosen schema after each observed chunk and are used only to construct training residuals; they are unavailable at inference.

\begin{table}[t]
    \centering
    \small
    \caption{Update and retrieval training tasks.}
    \label{tab:update-retrieval}
    \setlength{\tabcolsep}{4pt}
    \begin{tabular}{@{}p{0.10\linewidth}p{0.24\linewidth}p{0.28\linewidth}p{0.28\linewidth}@{}}
        \toprule
        Task & Input & Allowed behavior & Supervised outcome \\
        \midrule
        Update & Store $M_n$ and ordered interaction chunks $\chi_{n,1:K}$
        & Search and read the store, archive evidence, and create or edit structured Markdown
        & The final state must contain the required changes in their designated locations while avoiding redundant units \\
        Retrieval & Store $M_n$ and current request $q_n$
        & Use read-only list, search, and file-reading operations; return concise Markdown bullets
        & The returned context must cover the required personal and procedural points while excluding irrelevant content \\
        \bottomrule
    \end{tabular}
\end{table}

\begin{table}[t]
    \centering
    \footnotesize
    \caption{Memory operations and task-specific availability.}
    \label{tab:memory-operations}
    \setlength{\tabcolsep}{3.5pt}
    \begin{tabular}{@{}p{0.24\linewidth}p{0.56\linewidth}cc@{}}
        \toprule
        Function & Purpose & Update & Retrieval \\
        \midrule
        \path{save_raw}
        & Archive the interaction evidence visible in the current chunk.
        & \cmark & \xmark \\
        \path{list_memory_files}
        & Inspect the store structure without reading file contents.
        & \cmark & \cmark \\
        \path{grep_memory}
        & Search memory and return matches with file locations.
        & \cmark & \cmark \\
        \path{read_memory}
        & Read a selected file or line range for focused inspection.
        & \cmark & \cmark \\
        \path{edit_memory}
        & Create, replace, or insert structured Markdown content.
        & \cmark & \xmark \\
        \bottomrule
    \end{tabular}
\end{table}

\subsection{Update and Retrieval Tasks}
\label{app:memrsd-tasks}

The same memory-manager policy $f_\theta$ is trained for both \emph{Update} and \emph{Retrieval}. They share the structured Markdown store and outcome-based supervision, but differ in whether the persistent state may change. Table~\ref{tab:update-retrieval} summarizes their task contracts, and Table~\ref{tab:memory-operations} lists the available memory operations.

\paragraph{Update task.}
Given the current store $M_n$ and ordered interaction chunks $\chi_{n,1:K}$, the manager may inspect the store, archive new evidence, and create or edit structured Markdown. Update is scored from the realized state transition rather than agreement with a canonical edit sequence: different operation paths receive equal credit when they produce semantically equivalent and correctly organized stores. This allows the manager to consolidate related content, revise existing memories, or repair an omission in a later chunk, reflecting the continual maintenance required by a persistent memory store.
The runtime system and instance-specific user prompt templates are shown in Figure~\ref{fig:memrsd-update-prompt}.

\paragraph{Retrieval task.}
Given $M_n$ and the current request $q_n$, the manager uses read-only operations to identify the personal and procedural memories needed by the task agent while leaving $M_n$ unchanged. The output is a compact assistant-facing set of memory points, without file paths, benchmark identifiers, or other annotation metadata. Retrieval therefore evaluates whether relevant information remains accessible from the maintained store as interaction history accumulates. Retrieval need not explicitly serialize personal--procedural relation edges; the maintained organization is successful when it enables the relevant personal and procedural evidence to be jointly recovered for the current task.
The corresponding retrieval prompts are shown in Figure~\ref{fig:memrsd-retrieval-prompt}.

\paragraph{Memory-operation interface.}
Update exposes the full interface in Table~\ref{tab:memory-operations}: \texttt{save\_raw}, \texttt{list\_memory\_files}, \texttt{grep\_memory}, \texttt{read\_memory}, and \texttt{edit\_memory}. Retrieval exposes only the read-only list, search, and read operations. The \texttt{scope} argument selects \texttt{all}, \texttt{user}, \texttt{events}, \texttt{skills}, \texttt{raw}, or a specific file; raw evidence must be accessed explicitly through the \texttt{raw} scope. \texttt{grep\_memory} performs lexical search only, without an implicit embedding or BM25 retriever. During Update, each observed chunk is archived once through \texttt{save\_raw}, while \texttt{edit\_memory} supports \texttt{create}, \texttt{str\_replace}, and \texttt{insert}. Invalid or failed operations are returned to the manager and contribute to the format reward.

Update contains multiple state-changing segments and therefore uses both criterion- and chunk-level OPSD routing. Retrieval has a single read-only output segment and uses only the criterion-level gate.

\subsection{Location-Aware GRPO}
\label{app:memrsd-grpo}

For persistent memory management, writing the correct content is \emph{not} sufficient: as the store accumulates information over long trajectories, memories must also be \textbf{placed and organized} so that they remain maintainable and retrievable. We therefore score Update from both the semantic correctness and the location of the realized state change, rather than rewarding content that is merely appended somewhere in the store.

The Content-only GRPO variant in Table~\ref{tab:memrsd-ablation} uses the
DeltaMem-style soft semantic reward for realized memory-state changes.
Location-aware GRPO adds placement credit to this content reward under the
same training and storage setup.

\paragraph{Location-aware update reward.}
For rollout $i$, we compare its realized state change with the canonical change from the same pre-session store:
\begin{equation}
\widehat{\Delta}_i
=
\operatorname{diff}(M_n^\star,M_{i,K}^{+}),
\qquad
\Delta_n^\star
=
\operatorname{diff}(M_n^\star,M_{n+1}^\star),
\label{app:eq:state-deltas}
\end{equation}
where \texttt{diff} returns added, deleted, or modified memory units. Let
$\widehat{\Delta}_i=\{\widehat u_a\}$ and
$\Delta_n^\star=\{u_b^\star\}$. We compute
\begin{equation}
s_{ab}
=
\operatorname{sim}_{\mathrm{emb}}
\left(
\operatorname{text}(\widehat u_a),
\operatorname{text}(u_b^\star)
\right),
\end{equation}
and retain only type-compatible pairs satisfying
$s_{ab}\geq\tau_{\mathrm{sem}}$.

For each candidate pair, location correctness is
\begin{equation}
\ell_{ab}=
\begin{cases}
1, & \text{event units},\\
1, & \text{user or index units under the same target heading},\\
1, & \text{procedural units in the same domain-specific skill path},\\
0, & \text{otherwise}.
\end{cases}
\label{app:eq:location-indicator}
\end{equation}
We then compute a one-to-one matching $\gamma_i^\star$ that lexicographically maximizes
\begin{equation}
\left(
\sum_{(a,b)\in\gamma}\ell_{ab},
\;|\gamma|,
\;\sum_{(a,b)\in\gamma}s_{ab}
\right).
\label{app:eq:location-matching}
\end{equation}
This prioritizes correctly located matches, followed by coverage and semantic similarity, preventing a highly similar but misplaced unit from displacing a correctly organized one.

Let $K_b$ denote the required keywords of gold unit $u_b^\star$. For a matched pair, its content coverage and location weight are
\begin{equation}
c_{ab}
=
\frac{
\sum_{v\in K_b}
\mathbb{I}[v\in\operatorname{text}(\widehat u_a)]
}{
|K_b|+\epsilon
},
\qquad
w_{ab}^{\mathrm{loc}}
=
\begin{cases}
1, & \ell_{ab}=1,\\
0.2, & \ell_{ab}=0.
\end{cases}
\end{equation}
The resulting location-aware soft true-positive count is
\begin{equation}
\mathrm{TP}_i^{\mathrm{loc}}
=
\sum_{(a,b)\in\gamma_i^\star}
w_{ab}^{\mathrm{loc}}c_{ab}.
\end{equation}
Precision and recall are then
\begin{equation}
P_i^{\mathrm{state}}
=
\frac{\mathrm{TP}_i^{\mathrm{loc}}}
{|\widehat{\Delta}_i|+\epsilon},
\qquad
R_i^{\mathrm{state}}
=
\frac{\mathrm{TP}_i^{\mathrm{loc}}}
{|\Delta_n^\star|+\epsilon},
\end{equation}
with
\begin{equation}
F_i^{\mathrm{state}}
=
\frac{
2P_i^{\mathrm{state}}R_i^{\mathrm{state}}
}{
P_i^{\mathrm{state}}+R_i^{\mathrm{state}}+\epsilon
}.
\label{app:eq:state-f1}
\end{equation}
Unmatched predictions reduce precision, including duplicate or redundant units, while unmatched gold units reduce recall. If both deltas are empty, $F_i^{\mathrm{state}}=1$; if only one is empty, $F_i^{\mathrm{state}}=0$.

We additionally use a format reward $R_i^{\mathrm{format}}$ for valid structured memory operations and an efficiency reward $R_i^{\mathrm{eff}}$ that penalizes excessive tool calls and unnecessary state growth. The complete Update reward is
\begin{equation}
R_i^{\mathrm{upd}}
=
0.8F_i^{\mathrm{state}}
+
0.1R_i^{\mathrm{format}}
+
0.1R_i^{\mathrm{eff}}.
\label{app:eq:update-reward}
\end{equation}

\paragraph{Retrieval reward.}
Let $\widehat{\mathcal C}_i$ denote the memory points returned by rollout $i$ and $\mathcal C_n^\star$ the required context for request $q_n$. Retrieval uses the same thresholded one-to-one semantic matching and keyword coverage, but no location penalty because the store is not modified. This yields $F_i^{\mathrm{ret}}$ analogously to Eq.~\ref{app:eq:state-f1}. Its total reward is
\begin{equation}
R_i^{\mathrm{ret}}
=
0.8F_i^{\mathrm{ret}}
+
0.1R_i^{\mathrm{format}}
+
0.1R_i^{\mathrm{eff}},
\label{app:eq:retrieval-reward}
\end{equation}
where the auxiliary terms evaluate output format and retrieval efficiency.

\paragraph{GRPO optimization.}
Each rollout group contains only one task type. Given task reward $R_i$, the group-relative advantage is
\begin{equation}
A_i
=
\frac{
R_i-\overline{R}
}{
\operatorname{Std}(R_1,\ldots,R_B)+\epsilon_A
}.
\label{app:eq:grpo-advantage}
\end{equation}
For a valid manager-generated token $y_{i,t}$ with online context $h_{i,t}^{S}$, define
\begin{equation}
\rho_{i,t}(\theta)
=
\frac{
f_\theta(y_{i,t}\mid h_{i,t}^{S})
}{
f_{\theta^-}(y_{i,t}\mid h_{i,t}^{S})
}.
\end{equation}
For the set $\mathcal V$ of valid manager-generated tokens, the clipped GRPO loss is
\begin{equation}
\mathcal L_{\mathrm{GRPO}}(\mathcal V)
=
-\frac{1}{|\mathcal V|}
\sum_{(i,t)\in\mathcal V}
\min\!\left[
\rho_{i,t}(\theta)A_i,\,
\operatorname{clip}
\bigl(\rho_{i,t}(\theta),1-\epsilon_c,1+\epsilon_c\bigr)A_i
\right].
\label{app:eq:grpo-loss}
\end{equation}
Prompt tokens, padding, tool observations, and environment outputs are excluded from $\mathcal V$. Thus, the task-level outcome reward supervises the complete manager trajectory while gradients are applied only to manager-generated tokens.

\subsection{Dynamic Residual-Routed OPSD}
\label{app:memrsd-opsd}

GRPO becomes ineffective for a criterion when every rollout in the group misses it, leaving no successful behavior for relative learning. \textsc{PraxisMem} addresses this with residual-guided OPSD: criterion-level routing identifies unresolved requirements, while chunk-level routing applies privileged token guidance only where the realized memory state still contains the corresponding defect.

\paragraph{Criterion activation.}
Let $\kappa_j$ denote a required gold criterion represented by memory unit $u_j^\star$. Using the location-aware matching $\gamma_i^\star$ from Eq.~\ref{app:eq:location-matching}, its final coverage in rollout $i$ is
\begin{equation}
\operatorname{cov}_{ij}
=
\begin{cases}
w_{aj}^{\mathrm{loc}}c_{aj},
& (a,j)\in\gamma_i^\star,\\
0,
& \text{otherwise}.
\end{cases}
\label{app:eq:criterion-coverage}
\end{equation}
Thus, missing units receive zero coverage, while incomplete or misplaced units receive partial coverage. We activate criterion $j$ only if no rollout satisfies it:
\begin{equation}
u_j
=
\mathbb{I}\!\left[
\max_{1\le i\le B}\operatorname{cov}_{ij}
<
\tau_{\mathrm{cov}}
\right],
\qquad
\mathcal U=\{j:u_j=1\}.
\label{app:eq:criterion-gate}
\end{equation}
Redundant predictions do not define gold criteria and therefore do not independently activate OPSD; they remain penalized by state precision.

\paragraph{Dynamic residual and chunk routing.}
Before rollout $i$ acts on chunk $k$, we compare its realized store with the canonical memory state after that chunk:
\begin{equation}
z_{i,k}
=
\operatorname{SemDiff}
\bigl(
M_{i,k}^{-},
M_{n,k}^\star;
M_{n,k-1}^\star
\bigr).
\label{app:eq:dynamic-residual}
\end{equation}
The residual is \textbf{privileged training information} and contains three types of discrepancy:
\begin{equation}
z_{i,k}
=
\bigl(
Z_{i,k}^{\mathrm{missing}},
Z_{i,k}^{\mathrm{revise}},
Z_{i,k}^{\mathrm{redundant}}
\bigr).
\label{app:eq:residual-components}
\end{equation}
$Z^{\mathrm{missing}}$ contains absent required units; $Z^{\mathrm{revise}}$ contains incomplete, outdated, conflicting, or misplaced units; and $Z^{\mathrm{redundant}}$ contains unmatched predictions. For misplaced units, the residual also records the current and target locations.

The preceding canonical state $M_{n,k-1}^\star$ identifies requirements introduced by the current chunk without replacing the rollout's actual state. Earlier errors therefore remain visible until repaired, reflecting persistent memory management in which each realized state becomes the starting point for subsequent updates.

Let $\mathcal R_{i,k}$ denote the required criteria represented by the missing or revise components of $z_{i,k}$. The chunk gate is
\begin{equation}
g_{i,k}
=
\mathbb{I}
\left[
\mathcal R_{i,k}\cap\mathcal U\neq\varnothing
\right].
\label{app:eq:chunk-gate}
\end{equation}
Only residual information associated with activated criteria is exposed to the self-teacher. Retrieval has no intermediate state transitions, so it uses a single residual and criterion-level gate over the final output.

\paragraph{Residual-conditioned self-teacher.}
The student samples token $y_{i,k,t}$ from its ordinary context $h_{i,k,t}^{S}$. The frozen policy snapshot $f_{\theta^-}$ then evaluates the same sampled token with and without access to the residual:
\begin{align}
\ell_{i,k,t}^{S}
&=
\log f_{\theta^-}
(y_{i,k,t}\mid h_{i,k,t}^{S}),
\\
\ell_{i,k,t}^{T}
&=
\log f_{\theta^-}
(y_{i,k,t}\mid h_{i,k,t}^{S},z_{i,k}),
\\
\delta_{i,k,t}
&=
\operatorname{sg}
\left[
\ell_{i,k,t}^{T}-\ell_{i,k,t}^{S}
\right].
\label{app:eq:teacher-signal}
\end{align}
The teacher does not generate a replacement trajectory; it scores the student's own sampled tokens under the same prefix while additionally observing the training-only residual.

This corresponds to reverse-KL guidance on student-visited states:
\begin{equation}
D_{\mathrm{KL}}(f_S\,\|\,f_T)
=
\mathbb E_{y\sim f_S}
\left[
\log f_S(y\mid\xi)
-
\log f_T(y\mid\xi,z)
\right].
\label{app:eq:reverse-kl}
\end{equation}
Using the sampled token and treating $\delta_{i,k,t}$ as detached yields the surrogate
\begin{equation}
-\delta_{i,k,t}
\log f_\theta
(y_{i,k,t}\mid h_{i,k,t}^{S}).
\end{equation}

Let $\mathcal V$ denote all valid manager-generated tokens and
\begin{equation}
\mathcal A
=
\{(i,k,t)\in\mathcal V:g_{i,k}=1\}
\end{equation}
the subset inside activated chunks. The OPSD loss is
\begin{equation}
\mathcal L_{\mathrm{OPSD}}(\mathcal A)
=
-
\frac{
\sum_{(i,k,t)\in\mathcal A}
\delta_{i,k,t}
\log f_\theta
(y_{i,k,t}\mid h_{i,k,t}^{S})
}{
|\mathcal A|+\epsilon
}.
\label{app:eq:opsd-loss}
\end{equation}
Normalizing by active tokens keeps the OPSD scale independent of rollout length and gate density. If $\mathcal A=\varnothing$, the OPSD loss is zero and the privileged teacher forward is skipped.

\subsection{Joint Objective and Training Details}
\label{app:memrsd-training}

The two training signals are combined as
\begin{equation}
\mathcal L
=
\mathcal L_{\mathrm{GRPO}}(\mathcal V)
+
\lambda_{\mathrm{OPSD}}(s)
\mathcal L_{\mathrm{OPSD}}(\mathcal A)
+
\beta\mathcal L_{\mathrm{refKL}}(\mathcal V),
\label{app:eq:joint-objective}
\end{equation}
where the reference KL term is included once following the underlying GRPO implementation. GRPO remains active on all valid rollout tokens, while OPSD contributes only within activated segments.

For training step $s$ out of $S$, we linearly warm up the OPSD coefficient over the first 10\% of training:
\begin{equation}
\lambda_{\mathrm{OPSD}}(s)
=
0.05
\min\!\left(
\frac{s}{0.1S},
1
\right).
\label{app:eq:opsd-warmup}
\end{equation}

We use $\tau_{\mathrm{sem}}=0.7$ for semantic matching, $\tau_{\mathrm{cov}}=0.7$ for criterion satisfaction, and a location weight of $0.2$ for semantically compatible content stored at an incorrect heading or skill path. Retrieval uses no location penalty.

For Retrieval, the efficiency reward penalizes excessive memory calls:
\begin{equation}
R_i^{\mathrm{eff,ret}}
=
\max\!\left(
0,\,
1-
\frac{\max(0,N_i^{\mathrm{call}}-12)}{12}
\right),
\end{equation}
with a runtime cap of 30 calls.

For Update, efficiency accounts for both operation count and unnecessary memory growth:
\begin{equation}
\begin{aligned}
&R_i^{\mathrm{acts}}
=
\max\!\left(
0,\,
1-
\frac{\max(0,N_i^{\mathrm{act}}-20)}{20}
\right),\\
&R_i^{\mathrm{len}}
=
\min\!\left(
1,\,
\frac{L_i^{\mathrm{allowed}}}
{L_i^{\mathrm{pred}}+\epsilon}
\right),\\
&R_i^{\mathrm{eff,upd}}
=
0.5R_i^{\mathrm{acts}}
+
0.5R_i^{\mathrm{len}}.
\end{aligned}
\end{equation}
Here,
\[
L_i^{\mathrm{allowed}}
=
\max(1.5L_i^{\mathrm{gold}},128),
\]
which limits unnecessary state growth and discourages redundant accumulation that would make long-horizon memory harder to maintain and retrieve from. The format reward is zero when no memory operation is issued; otherwise, it reflects the fraction of successfully parsed and executed operations, together with the required Retrieval output format when applicable.

Update and Retrieval share the same memory-manager backbone but form separate rollout groups, so group-relative advantages are computed only among rollouts of the same task type.

\section{Benchmark Construction Details}
\label{app:construction}

\subsection{Tool Domains and SOP Grounding}
\label{app:domain-sop-grounding}

\begin{table}[t]
    \centering
    \caption{Application domains used in the training and test splits. Descriptions summarize the executable
    functionality used to instantiate SOPs.}
    \label{tab:application-domains}
    \footnotesize
    \setlength{\tabcolsep}{4pt}
    \renewcommand{\arraystretch}{0.98}
    \begin{tabular}{@{}>{\raggedright\arraybackslash}p{0.24\linewidth}
        >{\raggedright\arraybackslash}p{0.72\linewidth}@{}}
        \toprule
        \textbf{Domain} & \textbf{Description} \\
        \midrule
        \rowcolor{gray!20}\multicolumn{2}{@{}l}{\textbf{Training split}} \\
        Availability Planner & Finds time windows and records schedule blocks. \\
        Care Coordination & Selects services and records plans and check-ins. \\
        Case Note & Maintains case notes, decisions, tags, and actions. \\
        Client Correspondence & Prepares client replies and handoff messages. \\
        Community Posting & Creates board posts, notices, and comments. \\
        Home Maintenance & Records repairs, service contacts, and recurring tasks. \\
        Household Inventory & Tracks items and creates restock plans. \\
        Job Search Pipeline & Matches leads and records applications and follow-ups. \\
        Learning Plan & Selects resources and creates plans and review checkpoints. \\
        Local Resource Finder & Searches services and records shortlists and outreach. \\
        Meal Supply Planner & Selects supplies and creates meal and shopping plans. \\
        Media Queue & Maintains viewing, listening, and reading queues. \\
        Personal Finance Ledger & Records expenses, budget envelopes, and reimbursement notes. \\
        Pet Care & Records routine, activity, and backup-care plans. \\
        Project Board & Creates, moves, and annotates project cards. \\
        Purchase Support & Compares options and records requests and approvals. \\
        Reservation Planner & Finds availability and records reservations. \\
        Service Case & Routes cases and records resolutions and closures. \\
        Subscription Billing & Reviews plans and records subscription changes. \\
        Team Bulletin & Reviews updates and publishes notices and action prompts. \\
        \midrule
        \rowcolor{gray!20}\multicolumn{2}{@{}l}{\textbf{Test split}} \\
        Ticket Support & Creates, updates, resolves, and closes tickets. \\
        Travel Booking & Books, reviews, and cancels travel itineraries. \\
        Calendar & Creates, updates, reschedules, and cancels events. \\
        E-commerce & Manages products, carts, checkout, and returns. \\
        Education & Selects resources and creates learning plans. \\
        Email & Searches messages and creates replies and summaries. \\
        Entertainment & Searches options and creates themed watchlists. \\
        Fitness & Records appointments, activities, and wellness plans. \\
        Food & Searches recipes and restaurants and creates meal plans. \\
        Gaming & Searches games and creates personalized shortlists. \\
        Location & Creates commute, neighborhood, and accessibility plans. \\
        Media & Creates topic- and effort-aware reading lists. \\
        Social Posting & Creates posts, comments, mentions, and follows. \\
        Team Chat & Reviews channels and sends, edits, or deletes messages. \\
        Tasks and Documents & Manages tasks, notes, and procedure updates. \\
        Airline Service & Manages flights, baggage, cancellations, and compensation. \\
        Retail Service & Manages orders, returns, exchanges, and payments. \\
        Telecom Service & Manages lines, roaming, payments, and data refills. \\
        Trading & Manages accounts, orders, funds, and watchlists. \\
        Web Research & Searches pages and creates grounded research notes. \\
        \bottomrule
    \end{tabular}
\end{table}

Before constructing user trajectories, we select application domains that
reflect \textbf{recurring work and daily-life activities in persistent personal
assistance} and can be naturally mediated through a \textbf{general-purpose harness}.
Candidate domains include communication, scheduling, shopping, travel, health,
and personal finance. We retain a domain only when it supports
(i) persistent application objects,
(ii) both information-seeking and state-changing actions, and
(iii) arguments, outputs, or application states through which \textbf{personal and
procedural memory can affect execution}.
The held-out domains adapt tool families from existing executable-agent
environments to a deterministic offline setting. We separately author the
training domains around analogous high-level activities while using distinct
domain names, tool interfaces, fixture schemas, and workflows.

For each selected domain, we implement a \textbf{stateful mock application}. Its data
model specifies the application objects and valid state transitions; typed tool
interfaces expose search or read operations together with actions such as
creating, updating, sending, booking, or recording; and fixture files initialize
the state for reproducible sessions. Every state-changing action must be
observable through a subsequent read or final-state export, which allows the
same environment to support both task execution and programmatic evaluation. A
domain is admitted only if its fixtures load correctly, the exposed tool schemas
match their implementations, read and write calls succeed through the MCP
interface, writes produce the expected state transition, invalid tool calls
fail explicitly, and independent runs do not share mutable state. These applications are implemented by experienced engineers, with coding agents used to assist development and testing.

For each validated domain, we use a three-stage pipeline to construct \textbf{reusable
SOPs} before introducing a specific persona. An SOP specifies a common task goal,
the action dependencies required to complete it, and the expected application
state or artifact, while leaving all user-specific values uninstantiated.
First, a proposal annotator inspects the domain's objects and exposed tools and
proposes three to five tasks that a user would naturally perform in that
application. We reject goals that collapse to a single tool call or require an
artificial sequence added only to increase task length. For every retained
proposal, the annotator also marks which tool arguments, output fields, or
artifact slots could later carry personal memory.

Second, a tool engineer makes each proposal executable. The engineer creates a
fixture containing the prerequisite objects, target objects, and distractors;
runs a reference sequence against the application; and records the required
tool calls, returned object identifiers, and resulting state changes. For an
artifact workflow, this stage also fixes the artifact location and defines
deterministic checks for its existence and structure together with semantic
requirements for manual review. Third, an independent tester receives only the
task goal and exposed tool schemas, without the reference sequence, fixture
contents, or evaluation checks. The tester completes the task and exports its
tool trace and final application state. A tool-use SOP is accepted only when
this blind execution reaches the intended state and covers the
required calls; an artifact SOP must additionally pass its deterministic and
semantic checks.

We maintain disjoint domain pools for training and held-out testing. Each split
contains 20 domains; the training pool contains 120 accepted SOPs and the test
pool contains 140. The pools share no domain names, and training domains use
separate capability names, tool surfaces, fixture schemas, and SOP identifiers.
Table~\ref{tab:application-domains} summarizes the application domains and the
executable functionality represented in each split.
These accepted SOPs provide the split-specific pool for trajectory construction.

\subsection{Construction Pipeline}
\label{app:construction-pipeline}

\begin{algorithm}[t]
    \caption{Sequential construction of a PraxisMemBench trajectory.}
    \label{alg:praxismem-construction}
    \small
    \begin{algorithmic}[1]
        \Require Initial canonical state $\Omega_1^\star=(M_1^\star,S_1^\star)$,
        domain--SOP pool $\mathcal{P}$, number of sessions $N$
        \For{$n=1,\ldots,N$}
            \Repeat
                \State Reconstruct the current canonical state
                $\Omega_n^\star=(M_n^\star,S_n^\star)$ from previously accepted
                sessions
                \State Sample application domain $d_n$ and SOP
                $\sigma_n$ from $\mathcal{P}$ when the session is executable
                \State Select personal fields and active procedural memories
                from $M_n^\star$ that can jointly shape the session
                \State Synthesize a user event $e_n$ that evolves the current
                personal state
                \State Construct natural dialogue that discloses newly
                introduced personal or procedural information
                \State Select historical personal--procedural dependencies
                required by the current task
                \State Instantiate candidate task $X_n$, including the public
                request, application fixture, and gold checks
                \State Independently execute and validate $X_n$
                \If{$X_n$ passes validation}
                    \State Accept $X_n$ and commit
                    $(X_n,\Omega_{n+1}^\star)=\operatorname{Annotate}
                    (\Omega_n^\star;e_n,d_n,\sigma_n)$
                \Else
                    \State Repair or resample the failed construction step and
                    revalidate
                \EndIf
            \Until{$X_n$ passes validation}
        \EndFor
        \State \Return Accepted trajectory $\{X_n\}_{n=1}^{N}$ and final
        canonical state $\Omega_{N+1}^\star$
    \end{algorithmic}
\end{algorithm}

PraxisMemBench constructs trajectories in temporal order, as summarized in Algorithm~\ref{alg:praxismem-construction}. At session $n$, construction starts from the canonical state $\Omega_n^\star=(M_n^\star,S_n^\star)$ established by previously accepted sessions. This builder-side canonical state records the trajectory facts and application state needed for construction; it does not prescribe a memory-system representation or organization. The selected event, domain, and SOP shape a candidate task; only independently validated sessions advance the canonical state. Changes in personal state can affect which procedure applies and how it is carried out in later tasks.

\paragraph{Persona initialization.}
We sample a short persona seed from PersonaHub~\citep{ge2024personahub} and
expand it into a structured, internally consistent profile using the prompt in
Figure~\ref{fig:persona-enrichment-prompt}. As summarized in
Table~\ref{tab:persona-ontology}, the schema covers six top-level categories,
ranging from stable identity and social context to preferences and practical
constraints. The expanded profile is builder-side state; its fields become
available to the assistant only when disclosed in accepted sessions.

\begin{table}[t]
    \centering
    \caption{Persona attribute ontology used to initialize PraxisMemBench
    trajectories. Each gray row denotes a top-level field group; the following
    row lists the information represented within that group.}
    \label{tab:persona-ontology}
    \resizebox{\linewidth}{!}{%
    \begin{tabular}{p{\textwidth}}
        \toprule
        \rowcolor{gray!20}\textbf{Basic Identity} \\
        Age, gender, location, languages, education, occupation, income band,
        and cultural background. \\
        \midrule
        \rowcolor{gray!20}\textbf{Professional and Daily Role} \\
        Job title, sector, responsibilities, common tasks, work schedule,
        digital-tool proficiency, and frequent collaborators. \\
        \midrule
        \rowcolor{gray!20}\textbf{Personality and Values} \\
        Core values, social orientation, work and planning styles, privacy
        awareness, sense of responsibility, and positive, neutral, and negative
        traits. \\
        \midrule
        \rowcolor{gray!20}\textbf{Relationship Network} \\
        Household, family, friends, colleagues, managers, students, community
        contacts, frequent contacts, and relationship strengths. \\
        \midrule
        \rowcolor{gray!20}\textbf{Preferences} \\
        Food, entertainment, communication, learning, work, shopping, health,
        schedule, and privacy preferences, together with explicit aversions. \\
        \midrule
        \rowcolor{gray!20}\textbf{Constraints and Boundaries} \\
        Budget, time, health, family, geography, device, policy and compliance,
        and career constraints. \\
        \bottomrule
    \end{tabular}}
\end{table}

\paragraph{Session sampling and event synthesis.}
To create opportunities for personal context and domain-specific procedures to
interact, each scheduled session is sampled from the user and procedural state
established by earlier accepted sessions. \emph{Tool-use and artifact sessions} draw
an accepted SOP from the trajectory's training or test pool, while encouraging
occasional reuse of previously visited domains and avoiding excessive
short-term repetition. After selecting the domain and SOP, we retrieve the
active procedural preferences for that domain and sample a mutable field from
the current persona snapshot, allowing the\textbf{ evolving personal state and
procedural history to jointly shape the session context}. \emph{Memory QA} sessions
instead sample stable and changing personal fields. 
Together with prior events, these selections define the inputs for session
construction.

To model an \textbf{evolving user} rather than a fixed persona, the event generator then
synthesizes a concrete change for each selected mutable field using the prompt
in Figure~\ref{fig:event-synthesis-prompt}. It receives the field's current
value, the complete persona snapshot, earlier events, the session time, and,
for executable tasks, the selected domain and SOP. Starting from the current
value, the generator proposes a new value together with a plausible event that
explains the transition and remains relevant to the upcoming task. We require
the old and new values to differ substantively, the event to remain consistent
with the current profile and history, and its wording not to copy entire
persona values or explicitly expose a name, age, gender, or occupation.
Invalid events are repaired or resampled before the session is instantiated.

\paragraph{Memory disclosure and dependency construction.}
Beyond constructing the executable task itself, each session must also establish
the conversational history from which later memory dependencies arise. Because
user-specific information in persistent assistance is typically \emph{revealed
incidentally across everyday interactions} rather than restated at task time, we
construct preceding user turns that naturally disclose new personal events and
procedural rules before they are needed.

\emph{Personal updates} describe changes in the user's state, while \emph{procedural updates}
introduce reusable, domain-specific requirements for how an application action
should be performed or an output should be formatted. New procedural rules are
stated explicitly when first introduced; later sessions may reuse them without
restating their exact content. In this way, both personal and procedural memory
are accumulated through natural interaction rather than exposed directly as
task specifications.

Given this accumulated history, we then select previously disclosed personal
facts together with procedural memories that remain applicable in the current
domain and user state. Recent information introduced immediately before the
task is normally excluded from the historical target set, since it does not
require long-term retrieval. The selected memories are instantiated into
measurable task dependencies, such as query constraints, object selection,
workflow decisions, output requirements, or final-state conditions. In joint
tasks, \textbf{personal context can further affect which procedural requirement applies
or how it should be instantiated}. The public request preserves the task goal
and necessary non-memory details while omitting the historical personal and
procedural information that the memory method must recover. With these
dependencies established, the next stage instantiates the concrete task and
its corresponding gold checks.

\begin{table}[t]
\centering
\caption{A complete test trajectory spanning nearly one year. QA
rows name the personal fields queried; executable rows summarize the task
presented to the assistant.}
\label{tab:representative-trajectory}
\scriptsize
\setlength{\tabcolsep}{3pt}
\renewcommand{\arraystretch}{0.88}
\begin{tabular}{@{}p{0.065\linewidth}p{0.09\linewidth}p{0.18\linewidth}p{0.61\linewidth}@{}}
\toprule
\textbf{Session} & \textbf{Date} & \textbf{Family / domain} & \textbf{Task} \\
\midrule
s001 & Jan. 8 & Tool / entertainment & Create separate watchlists of relaxing movies or shows and games. \\
s002 & Jan. 25 & Memory QA & Recall changes in sense of responsibility. \\
s003 & Feb. 6 & Artifact / travel & Set a travel budget limit and save context grounded in the user's travel preferences. \\
s004 & Feb. 22 & Tool / education & Find beginner Python resources and save a teaching-oriented plan for a pilot. \\
s005 & Mar. 9 & Memory QA & Recall changes in social tendency. \\
s006 & Mar. 25 & Artifact / travel & Create an SFO--JFK trip and save its booking receipt. \\
s007 & Apr. 7 & Tool / entertainment & Create separate calm screen and game watchlists using remembered preferences. \\
s008 & Apr. 23 & Memory QA & Recall changes in work-time structure and wake pattern. \\
s009 & May 9 & Artifact / tasks & Review active priority tasks and create notes for a family travel. \\
s010 & May 24 & Tool / e-commerce & Inspect a delivered compact-shelf order and request its return. \\
s011 & Jun. 21 & Memory QA & Recall changes in sense of responsibility and shopping avoids. \\
s012 & Jul. 4 & Artifact / web & Research and save a family-travel spending action note. \\
s013 & Jul. 21 & Tool / social & Find and repost a relevant community post. \\
s014 & Aug. 6 & Memory QA & Recall changes in social tendency. \\
s015 & Aug. 20 & Artifact / web & Research a local community resource and save an orientation-handoff note. \\
s016 & Sep. 5 & Tool / entertainment & Create an entertainment watchlist suited to the user's travel-planning needs. \\
s017 & Sep. 18 & Memory QA & Recall changes in household composition. \\
s018 & Oct. 3 & Artifact / location & Find accessible cultural venues and compare public-transit routes to them, \\
s019 & Oct. 18 & Tool / telecom & Verify an overdue bill and send a payment request. \\
s020 & Oct. 30 & Memory QA & Recall changes in professional sector and planning tendency. \\
s021 & Nov. 6 & Tool / education & Create a teaching-oriented Python plan for a community orientation helper. \\
s022 & Nov. 13 & Tool / e-commerce & Return a delivered wall itinerary board. \\
s023 & Nov. 20 & Tool / social & Find and repost a relevant community post highlighting a local place. \\
s024 & Nov. 27 & Tool / telecom & Check the overdue bill for a family phone line and send a payment request. \\
s025 & Dec. 4 & Artifact / location & Find two nearby accessible cultural services, compare public-transit routes to them. \\
\bottomrule
\end{tabular}
\end{table}

\paragraph{Task and gold instantiation.}
Given an accepted SOP and the personal--procedural dependencies selected above,
an experienced engineer, assisted by a coding agent, instantiates a concrete
session in which the public request remains \textbf{natural} while the targeted memories
have observable effects on task completion. The request, application fixture,
reference execution or artifact requirements, and gold checks are refined
jointly so that \textbf{personal context} and \textbf{procedural requirements} interact through
measurable task outcomes.

\begin{table*}[t]
    \centering
    \small
    \setlength{\tabcolsep}{10pt}
    \caption{Long-horizon scale of representative memory benchmarks and
    PraxisMemBench.}
    \label{tab:long-horizon-scale}
    \begin{tabular}{@{}llrr@{}}
        \toprule
        Benchmark & Setting & Sessions / turns & History tokens \\
        \midrule
        \multirow{2}{*}{LongMemEval} & S & $\sim$40--50 sessions & 115K \\
        & M & 500 sessions & 1,500K \\
        \multirow{2}{*}{MemBench} & Participation normal & Varies by trajectory & 10K \\
        & Participation long & Varies by trajectory & 100K \\
        \multirow{2}{*}{PersonaMem-v2} & Effective history & Not fixed & 32K \\
        & Extended history & Not fixed & 128K \\
        \multirow{2}{*}{AMemGym} & Base & 11 periods / 47.15 turns & 60K--140K average \\
        & Extra & 21 periods / 81.60 turns & 512K+ required \\
        MemoryArena & -- & 2--16 sessions & 14.1K--122.4K \\
        \rowcolor{gray!12}
        PraxisMemBench & Test trajectory & 25 sessions & 663.6K average \\
        \bottomrule
    \end{tabular}
\end{table*}

For \emph{Memory QA}, the session asks about previously disclosed attribute histories.
Gold answers record the relevant earlier value, change event, and current value,
including recent intermediate transitions when necessary, with each required
memory point separately annotated.

For \emph{tool-use tasks}, the selected SOP is instantiated in a session-specific
fixture containing prerequisite objects, targets, and plausible distractors.
A reference execution determines the required tool calls and final application
state. Gold annotations identify the tool arguments, object selections, or state
conditions that depend on personal memory, while procedural requirements are
checked against the tool trace or resulting state.

For \emph{artifact tasks}, the builder specifies a generated file or application-state
artifact together with deterministic checks for its existence, location,
structure, and procedural format. Semantic slots identify content that must
reflect historical personal memory. Across all three task families, every
claimed memory dependency must map to an observable question answer, tool
argument, artifact field, or application outcome; dependencies without a
measurable destination are removed or resampled.

\paragraph{Independent validation and acceptance.}
Before an executable session is accepted, an independent tester receives the
public request, only the personal and procedural memories required for that
task, and the exposed application tool schemas. The tester can run the same
application interface but cannot inspect the fixture, reference tool sequence,
gold artifact requirements, or grading code. We compare its resulting trace,
state, and artifact with the recorded execution and the same programmatic and
rubric-based checks used in evaluation. \emph{Tool tasks} must cover the
required reference steps, reach the intended state, and demonstrably use a
targeted personal memory point; \emph{artifact tasks} must pass their programmatic,
semantic-memory, and procedural checks. \emph{Memory QA} questions undergo gold-answer
consistency and answer-leakage checks. Failed sessions are repaired and
retested or resampled. Only after acceptance do the disclosed user changes,
procedural updates, and application outcome enter the state replayed for the
next session.

\subsection{Benchmark Scale}
\label{app:benchmark-validation}

Session count alone does not determine the volume or difficulty of persistent
history. Existing memory benchmarks therefore report complementary measures
such as sessions, turns, interaction periods, and history tokens. As summarized
in Table~\ref{tab:long-horizon-scale}, LongMemEval, MemBench, PersonaMem-v2,
and AMemGym explicitly evaluate multiple history-length settings
\citep{wu2025longmemeval,tan2025membench,jiang2025personamemv2,cheng2026amemgym}.
MemoryArena further shows that executable traces can reach 122.4K tokens with
only 2--16 sessions because each session contains multi-step actions and
observations \citep{he2026memoryarena}.

Motivated by the multi-session evaluation designs in Table~\ref{tab:long-horizon-scale}, PraxisMemBench uses 25 temporally ordered sessions per trajectory. This length provides repeated opportunities for personal state changes to affect which procedures apply and how they are executed, while keeping the construction, independent validation, and full evaluation of executable sessions tractable. Across the complete OpenClaw, Claude Code, and Hermes runs, the observable user, assistant, tool-call, and tool-observation history averages an estimated 663.6K tokens per trajectory (606.7K--742.7K). This scale supports persistent-memory evaluation, although the benchmark’s primary focus is user-conditioned procedural use. Upon acceptance, we will release the benchmark and construction tools so that future evaluations can extend trajectories with additional sessions.

\subsection{Benchmark Examples}
\label{app:benchmark-examples}

\paragraph{Example trajectory.}
Table~\ref{tab:representative-trajectory} follows one year in the life of Maya
Chen, a fictional community partnerships coordinator in Greater Vancouver.
Each row summarizes one session. Figures~\ref{fig:qa-task-example}--
\ref{fig:artifact-task-example} show one case from each task family, with
lightly abridged requests and the original dated evidence, application state,
reference execution, and grading criteria.

\paragraph{Representative tasks.}
Figures~\ref{fig:qa-task-example}--\ref{fig:artifact-task-example} present one
detailed example from each task family. The \emph{Memory QA} example isolates the
evolving personal state that persistent assistance must preserve. The two
executable examples demonstrate the benchmark's central premise that
procedures are inherently user-conditioned. In the \emph{Tool example}, procedural
memory requires the agent to inspect the pending order and its owner's
available payment methods, while personal memory determines which candidate
payment method fits the user's work, privacy, and compliance context. In the
\emph{Artifact example}, procedural memory determines how the ticket must be reviewed,
resolved, and verified, while the user's evolving regional commitments,
follow-through preferences, and trusted working relationship determine the
concrete resolution plan. In both cases, the procedure alone does not determine
the correct execution: it must be selected or instantiated under the current
personal context.

\begin{figure}[t]
\centering
\begin{tcolorbox}[
    enhanced,
    title=\textbf{Memory QA Example: Recalling a Changing Household},
    colframe=black!58,
    colback=white,
    colbacktitle=black!78,
    coltitle=white,
    boxrule=0.35mm,
    arc=2mm,
    width=\linewidth,
    fontupper=\small,
    left=3mm,right=3mm,top=2mm,bottom=2mm
]
\textbf{Name:} Maya Chen \qquad \textbf{Time:} September 18, 2026

\tcbline
\textbf{Public request}
\begin{tcolorbox}[colback=white,colframe=black!28,boxrule=0.2mm,
    leftrule=2mm,arc=0.8mm,left=2mm,right=2mm,top=1mm,bottom=1mm]
``How has my household changed over time? Include the earlier state, what
caused each change, and the current state.''
\end{tcolorbox}

\textbf{Relevant history}

\textbf{March 25, 2026 --- Personal.} Maya lived with her longtime partner
Alex in a compact apartment. Alex moved from expedition work into a stable
youth-program role and proposed, giving the household a more predictable shared
rhythm.

\textbf{May 24, 2026 --- Personal update.} After returning furniture that did
not fit their apartment, Maya and Alex began looking for a home that better fit
the household they were building.

\tcbline
\textbf{Gold answer}

\begin{enumerate}
    \setlength{\itemsep}{1pt}
    \setlength{\parskip}{0pt}
    \setlength{\topsep}{2pt}
    \item \textbf{Earlier state:} Maya lived with her longtime partner Alex in a
    compact apartment.
    \item \textbf{Change event 1:} Alex took a more stable youth-program role and
    proposed to Maya during a quiet hike.
    \item \textbf{Intermediate state:} Alex was now Maya's fiancé, and the couple
    had a more predictable shared rhythm for planning their life together.
    \item \textbf{Change event 2:} Furniture they ordered could not fit in the
    apartment, prompting them to acknowledge that they needed a larger home.
    \item \textbf{Current state:} Maya and Alex are actively apartment-hunting and
    sorting what they will take to a home better suited to their future household.
\end{enumerate}

\medskip
\textbf{Evaluation checkpoints}

\textbf{Recall:} fraction of the five gold points fully recovered.\\
\textbf{Precision:} fraction of the answer's claims supported by the gold points.\\
\textbf{F1:} harmonic mean of recall and precision.
\end{tcolorbox}
\caption{Memory QA task example.}
\label{fig:qa-task-example}
\end{figure}

\begin{figure}[t]
\centering
\begin{tcolorbox}[
    enhanced,
    title=\textbf{Tool-Use Example: Choosing the Correct Payment Method},
    colframe=black!58,
    colback=white,
    colbacktitle=black!78,
    coltitle=white,
    boxrule=0.35mm,
    arc=2mm,
    width=\linewidth,
    fontupper=\small,
    left=3mm,right=3mm,top=2mm,bottom=2mm
]
\textbf{Name:} Lauren Chen \qquad \textbf{Time:} June 16, 2026

\tcbline
\textbf{Public request}
\begin{tcolorbox}[colback=white,colframe=black!28,boxrule=0.2mm,
    leftrule=2mm,arc=0.8mm,left=2mm,right=2mm,top=1mm,bottom=1mm]
``Fix payment on my pending order \#GSD-SECURITY-2195. Move it off my
personal card to the available account method that matches work
context.''
\end{tcolorbox}

\textbf{Relevant history}

\textbf{February 26, 2026 --- Personal.} After a moderator was threatened when
their enforcement role was linked to personal accounts, the user became
especially protective of moderator identities and sensitive player information.

\textbf{May 12, 2026 --- Personal update.} Her work spans multiplayer game
development, live operations, and player trust and safety. She now incorporates
regional data-protection and youth-safety requirements during initial security
planning.

\textbf{June 16, 2026 --- Procedure.} Earlier in the conversation, she stated
that a payment change must first confirm that the order is pending and then
inspect the owner's available payment methods.

\tcbline
\textbf{Task-relevant MCP tools}

\texttt{get\_order\_details}: read an order, including its owner, status, and payment.\\
\texttt{get\_user\_details}: read a user's profile and available payment methods.\\
\texttt{modify\_pending\_order\_payment}: replace payment on a pending order.\\
\texttt{get\_product\_details}: read a product and its available variants.\\
\texttt{get\_item\_details}: read the details of a specific product variant.\\
\texttt{modify\_pending\_order\_address}: change the address of a pending order.\\
\texttt{modify\_pending\_order\_items}: exchange items within a pending order.\\
\texttt{cancel\_pending\_order}: cancel an order that has not yet shipped.\\
\textit{Eight additional tools are available in the retail application domain.}

\medskip
\textbf{Current application state}

Order \#GSD-SECURITY-2195 is pending, totals \$94.50, and is charged to a
personal Visa. The same account offers a general studio-operations card, a
live-operations event card, and a project card for player trust and safety,
moderator privacy, data protection, and youth-safety compliance. Other orders
and another user's account act as distractors.

\tcbline
\textbf{Gold execution}

Read order \(\rightarrow\) verify \texttt{pending} and obtain its owner
\(\rightarrow\) inspect that owner's payment methods \(\rightarrow\) select
the trust-and-safety project card \(\rightarrow\) update the target order.

\medskip
\textbf{Evaluation checkpoints}

\textbf{Payment method:} the project card for player trust and safety,
moderator privacy, data protection, and youth-safety compliance. \quad
\textbf{Procedure:} order and payment methods are read before modification.\quad
\textbf{Required steps:} all three calls target the correct order and owner.\quad
\textbf{Final state:} the new \$94.50 charge and corresponding refund are saved,
while distractor orders remain unchanged.
\end{tcolorbox}
\caption{Tool-use task example.}
\label{fig:tool-task-example}
\end{figure}

\begin{figure}[t]
\centering
\begin{tcolorbox}[
    enhanced,
    title=\textbf{Artifact Example: Resolving a Station-Accessibility Ticket},
    colframe=black!58,
    colback=white,
    colbacktitle=black!78,
    coltitle=white,
    boxrule=0.35mm,
    arc=2mm,
    width=\linewidth,
    fontupper=\small,
    left=3mm,right=3mm,top=2mm,bottom=2mm
]
\textbf{Name:} Marisa Alvarez \qquad \textbf{Time:} October 23, 2026

\tcbline
\textbf{Public request}
\begin{tcolorbox}[colback=white,colframe=black!28,boxrule=0.2mm,
    leftrule=2mm,arc=0.8mm,left=2mm,right=2mm,top=1mm,bottom=1mm]
``Resolve ticket 84 about the station-accessibility recommendation handoff.
Explain the practical fix in the resolution and reflect my regional work
geography, follow-through communication style, and close corridor relationship
there.''
\end{tcolorbox}

\textbf{Relevant history}

\textbf{January 21, 2026 --- Personal.} The user accepted a Eugene civic-mobility
collaboration, extending her work across Portland, Salem, and Eugene; Darren was
the Eugene transit coordinator involved in that work.

\textbf{February 6, 2026 --- Procedure.} In an earlier ticket task, she established
the process of reviewing open tickets before resolving the target, beginning the
saved text with ``Resolution Summary:'', and reading the ticket afterward.

\textbf{July 14, 2026 --- Personal update.} The Eugene work shifted to monthly
site visits with remote coordination. She also stated a preference for concrete
next steps with named owners and dates.

\textbf{October 16, 2026 --- Personal update.} After months of coordination,
Darren had become one of her trusted working relationships.

\tcbline
\textbf{Task-relevant MCP tools}

\texttt{ticket\_login}: authenticate to the ticket system.\\
\texttt{get\_user\_tickets}: list the current user's tickets, optionally by status.\\
\texttt{get\_ticket}: read one ticket and its current state.\\
\texttt{resolve\_ticket}: save resolution text and mark a ticket resolved.\\
\texttt{create\_ticket}: create a ticket with a title, description, and priority.\\
\texttt{edit\_ticket}: update fields on an existing ticket.\\
\texttt{close\_ticket}: close an existing ticket.\\
\texttt{logout}: end the authenticated ticket-system session.\\
\texttt{ticket\_get\_login\_status}: check whether a ticket-system user is authenticated.

\medskip
\textbf{Current application state}

Ticket 84 tracks the follow-up to a set of station-accessibility recommendations. The recommendations have been passed to the user’s team, but the ticket still needs a concrete plan specifying who will take each next step and when. The same user also has an unrelated open map-legend ticket
and a resolved elevator-outage ticket; an open procurement ticket belongs to
another user.

\tcbline
\textbf{Gold execution}

Log in \(\rightarrow\) list the user's \texttt{Open} tickets \(\rightarrow\)
resolve ticket 84 with a substantive saved resolution \(\rightarrow\) read
ticket 84 again to verify the result.

\medskip
\textbf{Expected artifact}

The resolution begins with ``Resolution Summary:'', assigns next steps with
named owners and dates, coordinates the Portland--Salem--Eugene work through
the Eugene collaboration, and asks Darren to confirm station-specific timing
before partner-facing updates.

\medskip
\textbf{Evaluation checkpoints}

\textbf{Regional work:} the Portland--Salem--Eugene corridor, coordinated
through the Eugene civic-mobility collaboration. \quad
\textbf{Follow-through:} concrete next steps with named owners and dates. \quad
\textbf{Trusted collaborator:} Eugene transit coordinator Darren confirms the
station-specific timing. \quad
\textbf{Procedure:} open-ticket review and required prefix. \quad
\textbf{Required steps:} login, review, resolve, and verification read. \quad
\textbf{Final artifact:} ticket 84 is resolved with substantive text containing
the practical plan.
\end{tcolorbox}
\caption{Artifact task example.}
\label{fig:artifact-task-example}
\end{figure}

\section{Evaluation Details}
\label{app:evaluation-details}

\begin{table*}[t]
    \centering
    \caption{Agreement of independent expert annotations and LLM judges on
    semantic grading. Human--Human reports QWK before adjudication; LLM rows
    are measured against the adjudicated human reference. Each rubric contains
    30 sampled scoring points.}
    \label{tab:judge-agreement}
    \small
    \setlength{\tabcolsep}{7pt}
    \begin{tabular}{lccccc}
        \toprule
        Evaluator & Overall & Tool Argument & Artifact Slot & QA Recall & QA Precision \\
        \midrule
        Human--Human        & 0.934 & 0.960 & 0.979 & 1.000 & 0.783 \\
        GLM-5               & 0.793 & 0.696 & 0.800 & 0.844 & 0.837 \\
        Qwen3.7-Plus        & 0.796 & 0.698 & 0.863 & 0.812 & 0.833 \\
        DeepSeek-V4-Flash   & 0.670 & 0.408 & 0.632 & 0.902 & 0.823 \\
        Kimi-K2.5           & 0.776 & 0.704 & 0.710 & 0.899 & 0.799 \\
        \bottomrule
    \end{tabular}
\end{table*}

\paragraph{Evaluation protocol.}
Each memory method is evaluated over a complete trajectory in chronological
order. Within each session, a user script specifies the sequence of
disclosures and requests. A fixed user message is sent directly when
provided; otherwise the user simulator generates the next public message
from the current and previous turn strategies and recent dialogue, using
the prompt in Figure~\ref{fig:user-simulator-prompt}. It either continues
an unmet request or advances to the next turn, without seeing future turn
strategies. The task agent then responds through its harness. In
OpenClaw-based runs, it receives the current session dialogue, available memory context, and application
tools exposed through MCP.

In the \emph{Scheduled} setting, controlled methods retrieve at predefined task
or targeted-question turns, using the current public user request as the
query, and append the returned context to that turn. In the \emph{Autonomous}
setting, requests are less specified; the task agent decides when to call
its available memory operation and supplies a query when that operation
accepts one. Controlled OpenClaw runs disable its native memory tools and
instruct the agent not to use workspace files as an alternate memory store;
native-memory methods retain their built-in interface. 

At session end,
controlled methods ingest the public dialogue
and observed tool events through their update interface; native methods
maintain memory through their agent behavior. The resulting memory store and application state are both carried into the next session. Evaluation is free-running with respect to these states: session-specific fixtures define or introduce the objects scheduled for the current session, but do not restore earlier application state to its canonical construction snapshot. The runner neither resets nor repairs application state after a failed execution. Consequently, earlier errors may cause state drift, including modified, missing, or already-consumed objects, and any resulting downstream failure remains part of the persistent evaluation. Canonical states and reference executions are used only for benchmark construction, validation, and grading; they are not used to overwrite the evaluated trajectory. OpenClaw assigns a distinct conversation
key to each session, so prior session dialogue is not directly carried into
the next task conversation.

\paragraph{Task-family grading.}
For Tool tasks, we compare the observed tool trace and final application state
with the required steps and procedural checks. Structured memory-bearing
arguments are checked against gold values; open-text arguments are scored by
the LLM judge for semantic coverage of the required personal facts using
the prompt in Figure~\ref{fig:judge-tool-memory}. Tool
Memory Slot averages these argument scores, Tool Procedural measures passed
procedural checks, and Tool Step SR measures required-step completion. For
Memory QA, the judge compares the answer with atomic gold facts: Recall is
the fraction of fully recovered facts using
Figure~\ref{fig:judge-memory-recall}, Precision measures the accuracy of
answer claims against the gold facts using
Figure~\ref{fig:judge-memory-precision-scheduled}, and F1 is their harmonic
mean. For Artifact
tasks, programmatic verifiers check the artifact's existence, structure, and
procedural requirements, while the judge scores whether its semantic slots
reflect the required personal facts using
Figure~\ref{fig:judge-artifact-slots}. Artifact Memory Slot averages the slot
scores, Artifact Procedural measures passed procedural checks, and Artifact
Assert measures passed non-procedural assertions.

\paragraph{Session success and aggregation.}
For results in the main paper, \emph{Session SR} denotes the
memory-conditioned session success rate. Let $m_{i,d}\in[0,1]$ be the score of
an applicable memory dimension $d\in\{\mathrm{pers},\mathrm{proc}\}$ for
session $i$, and let $\mathcal D_i$ contain the dimensions defined for that
session. Its binary success indicator is
\begin{equation}
z_i =
\begin{cases}
\mathbb I\!\left[\displaystyle\bigwedge_{d\in\mathcal D_i}
m_{i,d}\geq 0.75\right],
& i\text{ is Tool or Artifact},\\[2pt]
\mathbb I\!\left[F1_i\geq 0.75\ \land\ P_i\geq 0.60\ \land\ R_i\geq 0.60\right],
& i\text{ is Memory QA}.
\end{cases}
\label{eq:session-success}
\end{equation}

Session SR is a diagnostic, memory-conditioned measure and does not correspond to \emph{Strict E2E Session SR}, which additionally requires all successful task execution and outcomes.

For task family $t$, let $\mathcal E_t$ denote the evaluation-eligible sessions for which $z_i$ is defined. A small number of sessions are excluded from SR aggregation after final human validation to ensure consistent scoring. We compute the family-level and overall rates as
\begin{equation}
\operatorname{SR}_t
=
\frac{\sum_{i\in\mathcal E_t}z_i}{|\mathcal E_t|},
\qquad
\operatorname{SR}_{\mathrm{overall}}
=
\frac{\sum_t\sum_{i\in\mathcal E_t}z_i}
{\sum_t|\mathcal E_t|}.
\label{eq:session-sr-aggregation}
\end{equation}
Thus, overall Session SR is a pooled session-level rate rather than an
unweighted mean of the three family-level rates.

\paragraph{Judge agreement.}
To assess the reliability of the rubric-based scores, we sample 30 scored
points from each of the four semantic rubrics, for 120 points in total. The
sample is balanced across the original GLM-5 score levels and spans different
memory methods and sessions. Two independent expert annotators score the same
blinded evidence packets using the corresponding evaluation rubrics, without
access to method identities, the original GLM-5 scores, or other annotations.
We compute quadratic weighted Cohen's $\kappa$ (QWK) on their independent
ratings and adjudicate disagreements to form the human reference. We then
apply GLM-5, Qwen3.7-Plus, DeepSeek-V4-Flash, and Kimi-K2.5 to the same packets
with reasoning disabled and compare each judge with this reference.

As shown in Table~\ref{tab:judge-agreement}, the two expert annotators reach an
overall QWK of 0.934. GLM-5 reaches 0.793 against the adjudicated reference,
comparable to Qwen3.7-Plus (0.796) and Kimi-K2.5 (0.776), indicating that the
semantic evaluation is not specific to a single judge model. DeepSeek-V4-Flash
shows lower agreement on Tool Argument scoring, where partial coverage of
memory-bearing open-text arguments leaves more room for rubric interpretation.

\section{Implementation Details}
\label{app:configurations}

\subsection{Benchmark Construction and Evaluation}

\paragraph{Construction setup.}
Application SOPs and individual sessions were constructed by software
engineers working with coding agents and independently validated by a separate
group of engineers. We use the GPT-5.5 API to expand PersonaHub seeds into
structured profiles and to synthesize personal change events.

\paragraph{Models and evaluation.}
Qwen3.7-Plus serves as the task agent, while GLM-5 serves as both the user
simulator and the LLM judge. LLM-based memory baselines use Qwen3.5-9B for
memory management where applicable, and \textsc{PraxisMem} is trained from the
same Qwen3.5-9B base model. Memory-R2 is likewise retrained from Qwen3.5-9B
following its released repository and LoGo-GRPO training procedure.
Embedding-based components use Qwen3-Embedding-0.6B. The user simulator, LLM
judge, and memory LLM run with temperature 0 and reasoning disabled.

Programmatic verifiers evaluate required tool steps, procedural requirements,
artifact assertions, and observable application state, while the GLM-5 judge
scores semantic personal-memory requirements and Memory QA responses against
the gold annotations.

\paragraph{System implementations.}
OpenClaw 2026.4.15 provides the task harness for controlled memory methods and
the OpenClaw native-memory baseline, with the latter retaining its built-in
memory operations. The other native-memory baselines use Claude Code 2.1.153
with Agent SDK 0.2.128 and Hermes v0.20.0, respectively. In native-memory
systems, the task LLM performs memory retrieval and maintenance within the
agent loop rather than relying on a separate memory LLM.

We use mem0ai==2.0.7 for Mem0,
graphiti-core==0.29.2 for Zep,
MemoryOS==2.0.30 for MemOS, and AgentScope ReMe v0.2.0.6.
A-Mem follows its released repository implementation. For these systems, we
preserve their original memory representation and update--retrieval procedure,
while adding only the runner interface required for chronological session
execution, isolated trajectory storage, and usage logging.

All controlled memory methods use a 4,096-token final memory-injection budget.
DualMem combines retrieval from its personal and procedural stores within this
same budget. Long Context instead uses a 128,000-token total task context; when
the accumulated history exceeds this limit, the oldest history is truncated
first while preserving the current-session context and system instructions.
We report one complete-test-split inference run per method.

\subsection{DualMem Baseline}

\paragraph{Memory representation.}
DualMem is a dual-store baseline inspired by Mem0
\citep{chhikara2025mem0}. It instantiates two isolated memory stores with the
same Qwen3.5-9B memory model and Qwen3-Embedding-0.6B embedder. Each store has
its own local Qdrant vector collection, entity index, history database, and
user scope. The Personal store retains user identity, relationships,
preferences, goals, constraints, resources, commitments, and evolving state.
The Procedural store retains reusable steps, preconditions, branches,
tool-use rules, output conventions, verification checks, and recovery
procedures. Personal extraction follows the Mem0 extraction pipeline with a
personal-information scope, while procedural extraction uses a dedicated,
domain-general procedural instruction. No record or entity link is shared
between the two stores.

\paragraph{Memory update.}
At the end of each session, both stores receive the same user--assistant
interaction and visible tool evidence. DualMem updates them serially in the
fixed order Personal then Procedural. Within each store, the memory manager may
add, update, or delete memory records, while the entity-linking stage updates
the corresponding entity-index relations. Deduplication and conflict handling
remain local to that store: the Procedural updater cannot read the Personal
store or condition its decisions on a personal update, and the Personal
updater cannot link its records to a procedure. Thus, the two updates separate
what is remembered about the user from what is remembered about how to act.

\paragraph{Retrieval and injection.}
At each scheduled retrieval point, the same query is issued first to the
Personal store and then to the Procedural store. Each side returns up to 20
items in its own similarity order; DualMem performs no cross-store score
normalization, reranking, linking, or conflict resolution. The two result lists
are serialized under separate \texttt{personal\_memory} and
\texttt{procedural\_memory} sections and concatenated within the shared
4,096-token memory budget. Unused capacity from a shorter side is assigned to
the other side, while truncation preserves complete memory items whenever
possible. The resulting block is supplied directly to the task agent, which
must determine how the two sets of memories apply to the current task. In the
Autonomous setting, one agent-issued \texttt{memory\_search(query)} call
triggers the same sequential two-store retrieval and merged return.

\subsection{PraxisMem Training}

We derive 1,921 training examples from the 20 training trajectories: 481 Update examples (25.0\%) and 1,440 Retrieval examples (75.0\%). An Update example is a session-level rollout. For each training session, we instantiate a reference memory state under the PraxisMem Markdown schema from the trajectory and its annotated memory dependencies. This reference is one valid organization under our chosen schema, rather than a benchmark-provided or uniquely correct memory state. An Update example starts from the preceding reference state and targets the state after incorporating the current interaction. Chunk-level reference states are obtained by replaying the constructed updates in temporal order and are used only to form training residuals.

Reference Update units identify the target file and section together with three to six required keywords; Retrieval targets identify expected context supported by the preceding state together with one to three required keywords. The annotations are reviewed and automatically checked against the corresponding gold text. To match deployment, Update inputs use the same memory-ingest chunking routine as evaluation: visible user--assistant messages and tool events are kept in chronological order and packed into approximately 4,096-token chunks, preserving message boundaries where possible and splitting an overlong message first by paragraph. Retrieval examples use the public request at the checkpoint as the query, and their target evidence must already exist in the preceding reference state.

We train \textsc{PraxisMem} from the Qwen3.5-9B base model. All experiments are conducted on one node with 8 Nvidia A100 GPUs. Table~\ref{tab:memrsd-training-config}
summarizes the sampling, optimization, and method-specific training settings.

\begin{table}[t]
    \centering
    \caption{Training configuration for \textsc{PraxisMem}.}
    \label{tab:memrsd-training-config}
    \small
    \setlength{\tabcolsep}{5pt}
    \renewcommand{\arraystretch}{1.08}
    \begin{tabular}{@{}>{\raggedright\arraybackslash}p{0.39\linewidth}
        >{\raggedright\arraybackslash}p{0.55\linewidth}@{}}
        \toprule
        \textbf{Parameter} & \textbf{Value} \\
        \midrule
        \rowcolor{gray!20}\multicolumn{2}{@{}l}{\textbf{Sampling}} \\
        Rollouts per example & 8 \\
        Temperature / top-$p$ / top-$k$ & 1.0 / 1.0 / $-1$ \\
        Max. prompt / response tokens & 24,576 / 4,096 \\
        \midrule
        \rowcolor{gray!20}\multicolumn{2}{@{}l}{\textbf{Optimization}} \\
        Strategy & full-parameter FSDP2 \\
        Batch sizes & 8 \\
        Learning rate & $10^{-6}$ \\
        GRPO clip & 0.2 \\
        Reference-KL coefficient & 0.01 \\
        Number of Training Epochs & 1 \\
        \midrule
        \rowcolor{gray!20}\multicolumn{2}{@{}l}{\textbf{PraxisMem-specific}} \\
        Wrong-location match weight & 0.2 \\
        Semantic / coverage thresholds & 0.7 / 0.7 \\
        OPSD teacher / routing & Current actor / criterion$\times$chunk \\
        OPSD peak weight / warmup & 0.05 / first 10\% of steps \\
        \bottomrule
    \end{tabular}
\end{table}

For Update, location-aware matching discounts semantically correct memories written in the wrong file or section. Dynamic residual OPSD activates guidance for criteria missed by the rollout group and routes it to chunks where the defect remains unresolved. The self-teacher receives the chunk-specific residual, while the student follows the ordinary memory-management prompt.

\subsection{Additional Training Analysis}
\label{app:additional-training-analysis}

The controlled ablation in Table~\ref{tab:memrsd-ablation} uses the first
eight trajectories in the fixed test-set order. All variants share the same
training data, base model, memory-store format, and evaluation protocol.

\begin{figure*}[t]
    \centering
    \includegraphics[width=\textwidth]{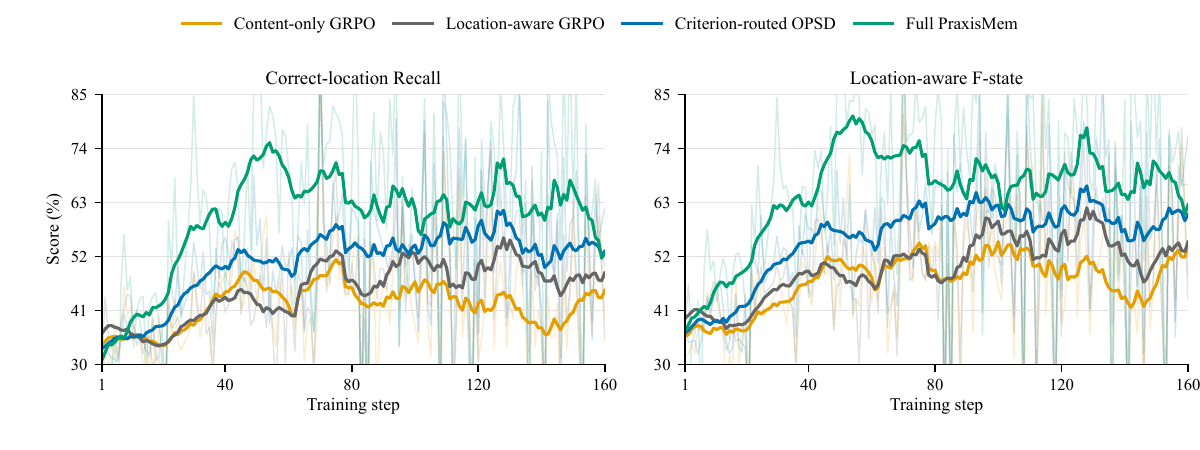}
    \caption{Training dynamics of four memory-training variants, measured by
    Correct-location Recall and location-aware F-state.}
    \label{fig:memrsd-training-dynamics}
\end{figure*}

\paragraph{Training dynamics.}
Figure~\ref{fig:memrsd-training-dynamics} compares four progressively
structured training objectives over the training process.
Correct-location Recall measures whether required information is maintained at
the intended memory location, while location-aware F-state jointly reflects
memory-state correctness and placement quality.

Content-only GRPO is consistently the weakest variant: its mean
Correct-location Recall and location-aware F-state over the final 20 steps are
41.1\% and 47.5\%, respectively, and both curves regress after improving in
the middle of training. Adding location-aware rewards raises these averages to
47.2\% and 51.6\%, showing that explicit placement supervision improves memory
organization but leaves substantial late-stage variation. Criterion-routed
OPSD further reaches 53.3\% and 58.5\%, indicating that routing supervision by
the failed criterion provides a more informative signal than location-aware
reward alone.

Full \textsc{PraxisMem} improves both metrics more rapidly and maintains the
strongest late-training averages. Correct-location Recall and location-aware
F-state first exceed 60\% at steps 35 and 29, respectively, and peak at 75.2\%
and 80.6\% around step 54. Their final-20-step averages remain 61.3\% and
66.8\%. These progressive gains show that accurate memory content alone is
insufficient: combining placement-aware optimization with fine-grained routing
more effectively maintains information in the organization from which later
tasks can recover user-conditioned procedures.

\begin{figure*}[t]
    \centering
    \includegraphics[width=\textwidth]{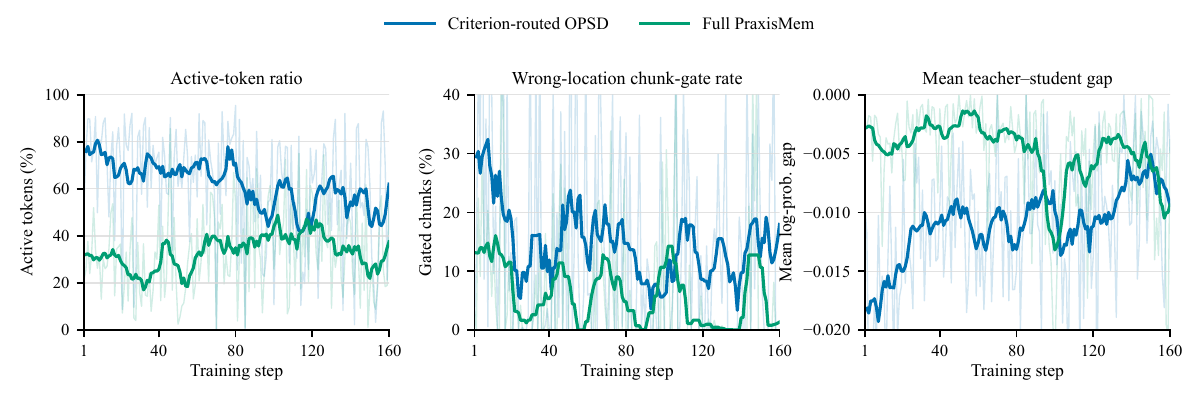}
    \caption{Routing and stability diagnostics for the two OPSD variants over training steps.}
    \label{fig:memrsd-routing-diagnostics}
\end{figure*}

\paragraph{Routing and stability diagnostics.}
We compare Criterion-routed OPSD and full \textsc{PraxisMem} under the same
initialization, training data, and optimization schedule.
Figure~\ref{fig:memrsd-routing-diagnostics} reports supervision coverage,
wrong-location routing, and teacher--student conflict over all 160 training
steps. The wrong-location chunk-gate rate records how often an Update chunk is
selected for teacher supervision because the rollout placed a memory in the
wrong file or section.

Criterion-routed OPSD activates 62.0\% of eligible assistant tokens on
average, whereas criterion$\times$chunk routing in full \textsc{PraxisMem}
activates 33.2\%. Its wrong-location chunk-gate rate also decreases from
14.6\% to 5.9\%, showing that a wrong-location residual is confined to fewer
parts of the generated update rather than propagated broadly across a
criterion group. The mean teacher--student log-probability gap remains
negative for both variants, consistent with prior observations that a
privileged teacher does not uniformly prefer student-sampled tokens. Full
\textsc{PraxisMem} has a smaller average gap magnitude ($-0.0048$ versus
$-0.0110$), indicating less conflict on the tokens selected for distillation.
 Thus,
\textsc{PraxisMem} applies more precise supervision: it reduces the range of
tokens and chunks exposed to the privileged teacher while improving the
quality of memory organization.

\section{Additional Results}
\label{app:additional-results}

\subsection{Oracle Diagnostics}
\label{app:oracle-analysis}

Oracle conditions in memory benchmarks typically isolate memory access while
leaving evidence integration, reasoning, and task execution to the same reader
or task agent. Consequently, providing gold memory does not necessarily yield
perfect performance. In LongMemEval, the strongest reader reaches 92.4\%
accuracy even when only the gold evidence sessions are provided, while
Llama-3.1-70B and Llama-3.1-8B reach 84.8\% and 75.6\%, respectively
\citep{wu2025longmemeval}. AMemGym directly supplies the current ground-truth
user state, yet Oracle performance across 17 models ranges from 81.6\% to
96.5\%, with a mean of 91.81\% \citep{cheng2026amemgym}. For action-oriented
evaluation, Mem2ActBench reports only 53.8\% parameter F1 and 88.2\%
tool-selection accuracy under Perfect Retrieval
\citep{shen2026mem2actbench}, while VitaBench 2.0 reaches at most 52.7\% even
after ground-truth user preferences are directly provided
\citep{chen2026vitabench}. These benchmarks differ in tasks and metrics, but
consistently show that correct memory access can still leave substantial errors
in reasoning, preference application, parameter grounding, and execution.

To further diagnose this gap in \textsc{PraxisMemBench}, we compare two Oracle
conditions. \emph{Oracle} provides the exact personal and procedural memories
required by the current task, while leaving the task agent to determine how
they should be grounded to tool arguments, artifact fields, action order, and
application state. \emph{Oracle-Action} additionally provides an action scaffold:
the required tools and their order, together with the source of task-relevant
arguments, such as personal memory, procedural memory, the user request, or a
previous tool result. It does not reveal runtime object identifiers, unseen
tool outputs, complete gold tool calls, final answers, or finished artifacts;
the agent must still interact with the application and execute the task.

\begin{table}[t]
    \centering
    \caption{Comparison of Oracle and Oracle-Action on the
    \textsc{PraxisMemBench} test set. }
    \label{tab:oracle-action}
    \resizebox{\linewidth}{!}{%
    \begin{tabular}{@{}lcccccccccccc@{}}
        \toprule
        Method & \multicolumn{4}{c}{Tool}
        & \multicolumn{4}{c}{Memory QA}
        & \multicolumn{4}{c}{Artifact} \\
        \cmidrule(lr){2-5}\cmidrule(lr){6-9}\cmidrule(l){10-13}
        & Pers. & Proc. & Step SR & SR
        & Recall & Precision & F1 & SR
        & Pers. & Proc. & Assert. & SR \\
        \midrule
        Oracle
        & 90.42 & 93.98 & 93.59 & 83.90
        & 98.45 & 99.29 & 98.82 & 99.29
        & 85.18 & 90.32 & 93.16 & 77.42 \\
        Oracle-Action
        & 95.42 & 97.30 & 96.67 & 94.10
        & 100.00 & 100.00 & 100.00 & 100.00
        & 100.00 & 95.24 & 97.67 & 95.65 \\
        \bottomrule
    \end{tabular}
    }
\end{table}

As shown in Table~\ref{tab:oracle-action}, action-level guidance substantially
narrows the remaining gap on executable tasks. Tool Session SR rises from
83.90\% to 94.10\%, while Artifact Session SR increases from 77.42\% to
95.65\%. Memory QA, by
contrast, is already nearly saturated under the standard Oracle and reaches
100\% with Oracle-Action. The larger gains on Tool and Artifact tasks indicate
that a substantial portion of the residual Oracle error comes after memory
access: the agent must still \textbf{combine personal and procedural requirements,
ground them to concrete parameters and application objects}, resolve runtime
tool outputs, and verify execution. Even Oracle-Action remains below perfect
end-to-end performance because these runtime decisions and outcomes are not
directly revealed. We therefore interpret Oracle as an upper bound on
\emph{memory supply} under a fixed task agent, rather than as a
programmatically perfect solver.
\subsection{Autonomous-Setting Results}
\label{app:autonomous-results}

Table~\ref{tab:autonomous-results} reports results in the Autonomous setting,
where requests are less specified and the task agent must decide whether, when,
and with what query to access memory.

\begin{table}[htbp]
    \centering
    \caption{Overall results in the Autonomous setting on the PraxisMemBench test set.
    Tool tasks report personal-memory (Pers.), procedural-memory (Proc.), step
    success rate (Step SR), and session success rate (Session SR). Memory QA reports
    recall, precision, F1, and SR. Artifact tasks report Pers., Proc.,
    artifact-assertion score (Assert.), and SR. }
    \label{tab:autonomous-results}
    \providecommand{\tbdscore}{\makebox[2.05em][c]{\textcolor{gray}{--}}}
    \resizebox{\linewidth}{!}{%
    \begin{tabular}{@{}lcccccccccccc@{}}
        \toprule
        Method & \multicolumn{4}{c}{Tool}
        & \multicolumn{4}{c}{Memory QA}
        & \multicolumn{4}{c}{Artifact} \\
        \cmidrule(lr){2-5}\cmidrule(lr){6-9}\cmidrule(l){10-13}
        & Pers. & Proc. & Step SR & SR
        & Recall & Precision & F1 & SR
        & Pers. & Proc. & Assert. & SR \\
        \midrule
        \rowcolor{gray!12}
        \multicolumn{13}{@{}l}{\textit{Diagnostic}} \\
        Oracle
        & 84.05 & 93.55 & 89.96 & 82.93
        & 97.02 & 97.77 & 97.03 & 100.00
        & 85.67 & 84.62 & 92.36 & 74.19 \\
        Long Context
        & 35.42 & 34.35 & 59.46 & 16.67
        & 36.10 & 44.21 & 36.79 & 14.29
        & 30.47 & 35.94 & 64.71 & 10.26 \\
        DualMem
        & 25.47 & 19.11 & 61.26 & 12.50
        & 24.07 & 40.53 & 27.53 & 5.71
        & 29.69 & 19.27 & 73.82 & 15.38 \\
        \midrule
        \rowcolor{gray!12}
        \multicolumn{13}{@{}l}{\textit{Basic baselines}} \\
        BM25
        & 18.57 & 27.42 & 76.15 & 14.63
        & 26.76 & 44.51 & 29.33 & 0.00
        & 21.33 & 23.72 & 79.39 & 19.35 \\
        Embedding RAG
        & 22.97 & 18.75 & 65.92 & 9.30
        & 28.52 & 45.95 & 30.56 & 9.38
        & 13.24 & 12.50 & 69.26 & 14.29 \\
        \midrule
        \rowcolor{gray!12}
        \multicolumn{13}{@{}l}{\textit{Harness-native memory}} \\
        OpenClaw
        & 17.52 & 12.60 & 60.85 & 6.25
        & 17.21 & 31.55 & 19.86 & 0.00
        & 25.26 & 27.78 & 71.49 & 20.51 \\
        Hermes
        & 25.66 & 35.77 & 62.58 & 10.42
        & 17.29 & 46.86 & 21.05 & 2.86
        & 5.73 & 18.75 & 46.39 & 10.26 \\
        Claude Code
        & 29.83 & 18.50 & 62.50 & 8.33
        & 17.05 & 37.53 & 20.43 & 0.00
        & 12.89 & 22.40 & 64.93 & 10.26 \\
        \midrule
        \rowcolor{gray!12}
        \multicolumn{13}{@{}l}{\textit{Advanced memory systems}} \\
        Mem0
        & 23.96 & 16.26 & 65.37 & 10.42
        & 30.17 & 42.09 & 31.37 & 2.94
        & 14.84 & 14.06 & 67.02 & 7.89 \\
        Zep
        & 16.86 & 20.33 & 63.86 & 2.08
        & 13.62 & 30.29 & 15.46 & 0.00
        & 15.10 & 10.94 & 74.05 & 15.38 \\
        A-Mem
        & 27.27 & 21.54 & 63.84 & 10.42
        & 20.07 & 35.33 & 22.10 & 0.00
        & 17.58 & 13.02 & 76.11 & 15.38 \\
        ReMe
        & 22.86 & 16.20 & 62.56 & 9.30
        & 8.33 & 22.77 & 9.47 & 0.00
        & 10.71 & 12.50 & 66.68 & 11.43 \\
        MemOS
        & 20.36 & 21.34 & 62.80 & 8.33
        & 21.21 & 33.12 & 21.38 & 0.00
        & 14.58 & 18.23 & 74.55 & 12.82 \\
        \midrule
        \rowcolor{gray!12}
        \multicolumn{13}{@{}l}{\textit{Learned memory managers}} \\
        Memory-R2
        & 17.38 & 12.90 & 61.39 & 12.20
        & 18.10 & 43.35 & 21.65 & 0.00
        & 19.50 & 15.33 & 70.39 & 19.35 \\
        \textbf{PraxisMem (Ours)}
        & \textbf{36.00} & \textbf{38.52} & \textbf{79.33} & \textbf{23.00}
        & \textbf{42.87} & \textbf{47.21} & \textbf{43.21} & \textbf{22.22}
        & \textbf{30.98} & \textbf{37.29} & \textbf{80.37} & \textbf{31.64} \\
        \bottomrule
    \end{tabular}%
    }
\end{table}

Compared with the Scheduled results in Table~\ref{tab:overall-results},
Autonomous interaction reduces task success for most methods,
particularly on Tool and Memory QA sessions. Hermes falls from 26.34\% to
10.42\% Tool Session SR and from 10.71\% to 2.86\% Memory QA Session SR;
MemOS falls from 19.12\% to 8.33\% and from 3.57\% to 0\%, respectively.
Component-level execution remains considerably higher: Hermes and MemOS still
reach 62.58\% and 62.80\% Tool Step SR, yet complete few sessions. Under less
specified requests, an agent therefore makes partial progress while failing
to recover the user state and procedural knowledge that jointly determine the
correct target, action, or execution path.

Autonomous performance also depends on whether the task agent invokes memory.
We define the retrieval-call rate as the fraction of sessions in which the
agent makes at least one \texttt{memory\_search} call. The advanced memory systems have an average retrieval-call
rate of 57.3\%. Scheduled evaluation supplies
retrieval at predefined turns, whereas Autonomous agents leave many sessions
without any memory call. They must infer that an underspecified request depends
on prior interaction, formulate an effective query, and then connect the
retrieved personal context to the applicable procedure. Under the same setting,
\textsc{PraxisMem} attains the strongest non-oracle Session SR for Tool, Memory QA,
and Artifact tasks at 23.00\%, 22.22\%, and 31.64\%, compared with the next-best
16.67\%, 14.29\%, and 20.51\%. Its advantage is consistent with maintaining
both memory content and the organization needed to apply procedures under the
current user state, while the remaining gap from Scheduled performance shows
that reliable autonomous memory access remains an open challenge.

\subsection{Task Completion by Memory Dependency Type}
\label{app:joint-task-completion}

\begin{figure}[t]
    \centering
    \includegraphics[width=\linewidth]{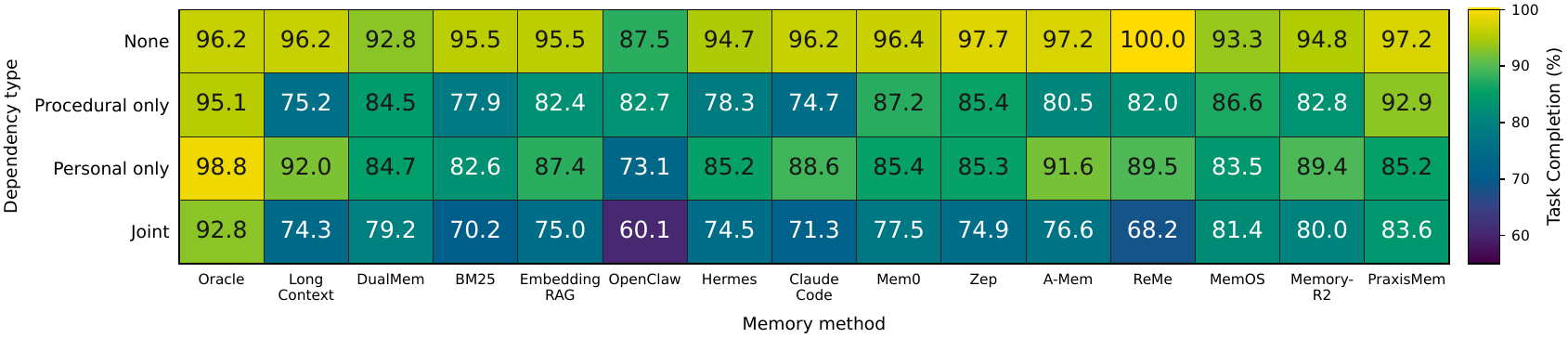}
    \caption{Task Completion by memory dependency type.}
    \label{fig:joint-task-completion}
\end{figure}

Figure~\ref{fig:joint-task-completion} reports task-completion performance after
grouping executable tasks by their required memory composition. We define
\emph{Task Completion} as the fraction of task-specific executable
requirements satisfied: required tool steps for Tool tasks and verifiable
artifact assertions for Artifact tasks. The rows distinguish tasks requiring
neither memory type, procedural memory only, personal memory only, or both.
The Joint row directly tests user-conditioned procedures, since correct
execution requires procedural knowledge to be applied under the relevant
personal context.

Across non-oracle methods, Task Completion on Joint tasks averages 74.8\%, 9.4
points below the 84.2\% mean over the Personal-only and Procedural-only rows.
This degradation appears across retrieval baselines, harness-native memory,
and advanced memory systems. DualMem makes the limitation especially clear:
despite reaching 84.7\% on Personal-only tasks and 84.5\% on Procedural-only
tasks, it falls to 79.2\% when both memories must be used together.
Independently maintaining the two stores and concatenating their retrieved
contents therefore preserves much of their isolated utility, but does not
reliably preserve the relation needed to determine which procedure applies to
the current user state.

Existing non-oracle baselines reach at most 81.4\% on Joint tasks, indicating
that their main weakness lies in maintaining actionable personal--procedural
relations over persistent interaction. \textsc{PraxisMem} achieves 83.6\%, the
strongest non-oracle result, by learning both memory content and its
organization. This result supports the central premise of PraxisMemBench:
procedures in persistent personal assistance are inherently user-conditioned,
and effective memory systems must keep the relevant procedure connected to the
evolving personal context that governs its selection and execution.

\subsection{DualMem Access and Failure Attribution}
\label{app:dualmem-failure-attribution}

To examine whether DualMem's degradation under stronger dependency reflects
evidence access or joint use, we first measure whether its retrieved context
contains the required personal and procedural evidence.
Table~\ref{tab:dualmem-retrieval-coverage} covers all 349 Tool and Artifact
tasks that require historical memory.

\begin{table}[t]
    \centering
    \caption{DualMem required-evidence retrieval by dependency type. Cells
    report retrieved sessions over eligible sessions, followed by percentage.}
    \label{tab:dualmem-retrieval-coverage}
    \small
    \setlength{\tabcolsep}{3.2pt}
    \begin{tabular}{@{}lccc@{}}
        \toprule
        Task group & Sessions & Personal retrieved & Procedural retrieved \\
        \midrule
        Personal-only & 22 & 15/22 (68.2\%) & -- \\
        Procedural-only & 24 & -- & 18/24 (75.0\%) \\
        Coordinated & 181 & 112/181 (61.9\%) & 140/181 (77.3\%) \\
        Interdependent & 122 & 96/122 (78.7\%) & 98/122 (80.3\%) \\
        \bottomrule
    \end{tabular}%
\end{table}

Interdependent tasks have higher marginal retrieval coverage than Coordinated
tasks for both personal evidence (78.7\% versus 61.9\%) and procedural
evidence (80.3\% versus 77.3\%). Their lower Task Completion is therefore not
accompanied by lower marginal evidence retrieval.

We next assign one mutually exclusive primary cause to each failed DualMem
joint session. Table~\ref{tab:dualmem-failure-attribution} separates missing
evidence from failures that arise when the two memory types must be related or
reconciled.

\begin{table}[t]
    \centering
    \caption{Primary failure attribution for failed DualMem joint sessions.
    Cells report counts followed by row percentages.}
    \label{tab:dualmem-failure-attribution}
    \scriptsize
    \setlength{\tabcolsep}{2.5pt}
    \resizebox{\linewidth}{!}{%
    \begin{tabular}{@{}lccccc@{}}
        \toprule
        & \multicolumn{3}{c}{Access-related} & \multicolumn{2}{c}{Relation-related} \\
        \cmidrule(lr){2-4}\cmidrule(l){5-6}
        Joint type & Personal missing & Procedural missing & Both missing
        & Fragmented relation & Competing memory \\
        \midrule
        Coordinated ($n=172$)
        & 54 (31.4\%) & 26 (15.1\%) & 15 (8.7\%) & 71 (41.3\%) & 6 (3.5\%) \\
        Interdependent ($n=103$)
        & 17 (16.5\%) & 15 (14.6\%) & 9 (8.7\%) & 59 (57.3\%) & 3 (2.9\%) \\
        \bottomrule
    \end{tabular}%
    }
\end{table}

Access-related errors account for 55.2\% of Coordinated failures and 39.8\%
of Interdependent failures. Conversely, relation-related errors increase from
44.8\% to 60.2\%, driven primarily by fragmented relations (41.3\% versus
57.3\%). Together with Table~\ref{tab:dualmem-retrieval-coverage}, this shift
is consistent with stronger interdependence placing greater demands on
preserving and applying the relation between personal and procedural memory,
rather than merely retrieving either type in isolation.

\subsection{Representative Failure Cases}
\label{app:failure-cases}

\newcommand{\CaseTraceEntry}[1]{%
  \par\noindent\hangindent=1.5em\hangafter=1 #1\par}

\paragraph{Analysis protocol.}
The failure analyses cover all failed joint-memory tasks in
Table~\ref{tab:dualmem-failure-attribution} and for the methods shown in the
failure-by-stage comparison. Two annotators independently assign one mutually
exclusive failure category without seeing the task's dependency label or each
other's judgments, and disagreements are resolved through adjudication. For
the evidence-age analysis, we group scored dependency points by the elapsed
time between their source evidence and target task, and compute the mean
point-level score within each time-distance bin.

\paragraph{Representative cases.}
The case studies trace how failures arise when a memory system cannot preserve
the connection between a user's evolving context and the procedure applied on
the user's behalf. Each case records the earlier evidence, the method's
observable memory state, the context supplied to the task agent, and the
resulting execution. 

Figure~\ref{fig:failure-dualmem-escalation} shows a fragmented-relation
failure in DualMem. Its two stores surface the brief-writing procedure and
several records about delegated revision work, but the retrieved fragments do
not reveal that a third email determines who should prepare the revision and
when that work can occur.
Figure~\ref{fig:failure-return} shows a fragmented-relation failure in Hermes.
The return procedure exposes several valid items, but the correct choice
depends on the remembered relation between Aino's family members. Hermes
recovers the procedure and the family members separately, then follows a stale
return precedent because their operative relation is missing.
Figure~\ref{fig:failure-openclaw} shows an
access failure in which neither the collaborator context nor the acknowledgment
procedure reaches the agent. Figure~\ref{fig:failure-status-update} shows an
asymmetric update that preserves personal context but loses the associated
document rules. Figure~\ref{fig:failure-email} shows the converse joint-use
problem: the output procedure is recovered, but competing personal memory
causes it to prioritize the wrong actions. These cases illustrate why
persistent assistance requires maintaining user-conditioned procedures beyond
storing individually plausible personal and procedural facts. 

\makeatletter
\setlength{\@fptop}{0pt}
\setlength{\@dblfptop}{0pt}
\makeatother

\begin{tcolorbox}[
    enhanced,
    breakable,
    title=\textbf{Fragmented Relation: DualMem Misses a Delegation Source},
    colframe=black!58,colback=white,colbacktitle=black!78,coltitle=white,
    boxrule=0.35mm,arc=2mm,width=\textwidth,
    fontupper=\small\linespread{0.96}\selectfont,
    left=3mm,right=3mm,top=2mm,bottom=2mm
]
\textbf{Task.} On February 14, 2026, Morgan asks the assistant to review recent
emails concerning whether a student's survey information may be shared with
the teaching team, then prepare a concise Markdown brief covering the related
privacy, file-export, and revision-planning issues.

\textbf{Earlier evidence.}
\begin{itemize}
    \setlength{\itemsep}{1pt}
    \setlength{\parskip}{0pt}
    \setlength{\topsep}{2pt}
    \setlength{\leftmargin}{4mm}
    \item \textbf{January 11, 2026}: After a wrist diagnosis imposed typing
    limits, Morgan began asking trained interns to prepare initial revisions and
    reviewing their drafts in short sessions.
    \item \textbf{January 23, 2026}: Morgan established a preference for tools
    and services with stable export guarantees and citation or audit trails.
    \item \textbf{January 25, 2026}: New teaching responsibilities meant that
    revision work with the interns had to be scheduled outside Morgan's classes
    and student meetings.
    \item \textbf{February 4, 2026}: The earlier escalation task established the
    procedure: list the inbox, read only the relevant source messages, write
    \texttt{escalation\_brief.md}, and neither draft nor send an email.
\end{itemize}

\textbf{Current application state.} The inbox contains three relevant messages.
The first asks whether a student's survey information can be shared with the
teaching team. The second warns that the temporary reference tool may no longer
export files reliably after Friday. The third says that trained interns can
prepare the required revision the following day, provided that the work is
scheduled outside Morgan's teaching commitments. A campus newsletter and a
software receipt are unrelated. Morgan's existing work arrangement makes the
third message necessary for deciding who should prepare the revision and when.

\textbf{Memory state.} DualMem maintains physically separate personal and
procedural stores, each with its own version history and local vector index.

\begin{tcolorbox}[colback=gray!4,colframe=gray!38,boxrule=0.25mm,arc=0.8mm,
    left=2mm,right=2mm,top=1mm,bottom=1mm]
\ttfamily
\fontsize{7.5pt}{8.5pt}\selectfont
memory/dualmem/\par
\quad|-- storage\_context.json \quad {\rmfamily\itshape store and namespace configuration}\par
\quad|-- ingest\_ledger.jsonl \quad {\rmfamily\itshape append-only update log}\par
\quad|-- personal/\par
\quad|\quad|-- history.db \quad {\rmfamily\itshape versioned personal-memory history}\par
\quad|\quad`-- qdrant/collection/dualmem\_personal/storage.sqlite\par
\quad|\qquad\quad {\rmfamily\itshape personal-memory vector index}\par
\quad`-- procedural/\par
\qquad\quad|-- history.db \quad {\rmfamily\itshape versioned procedural-memory history}\par
\qquad\quad`-- qdrant/collection/dualmem\_procedural/storage.sqlite\par
\qquad\qquad\quad {\rmfamily\itshape procedural-memory vector index}
\end{tcolorbox}

\par\medskip
\noindent\textbf{Memory access.}
\begin{tcolorbox}[colback=gray!4,colframe=gray!38,boxrule=0.25mm,arc=0.8mm,
    left=2mm,right=2mm,top=1mm,bottom=1mm,
    fontupper=\fontsize{7.5pt}{8.5pt}\selectfont]
\textbf{Personal retrieval (top 20).}\par
\CaseTraceEntry{\textbf{Ranks 1--9:} near-duplicate records restating the
current request, filename, and four required headings.}
\CaseTraceEntry{\textbf{Rank 15:} a delegation decision made after Morgan's
wrist diagnosis.}
\CaseTraceEntry{\textbf{Rank 17:} a task update describing a related work
handoff.}
\CaseTraceEntry{\textbf{Rank 19:} a project mentor taking a later coordination
pass.}
\CaseTraceEntry{\emph{The remaining records concern inbox flagging, unrelated
task handoffs, and other request summaries.}}
\vspace{1mm}
\textbf{Procedural retrieval (19 returned).}\par
\CaseTraceEntry{\textbf{Rank 1:} use the email tools to create a concise
disclosure/export escalation brief with the four required headings.}
\CaseTraceEntry{\textbf{Rank 4:} a generic procedure for listing, reading, and
flagging email.}
\CaseTraceEntry{\emph{The remaining 17 records concern task handoffs,
research notes, media lists, shopping, health updates, and other workflows.}}
\end{tcolorbox}

The two result lists are concatenated before inference. They provide the brief
structure and several individually relevant records about delegated work, but
do not reconstruct their practical connection: the third email must be read to
determine that the interns should prepare the revision outside Morgan's
scheduled teaching activities. The retrieved flagging procedure also competes
with the established instruction to write a file without drafting or sending
email.

\par\medskip
\noindent\textbf{Task-LLM execution.}
\begin{tcolorbox}[colback=gray!4,colframe=gray!38,boxrule=0.25mm,arc=0.8mm,
    left=2mm,right=2mm,top=1mm,bottom=1mm,
    fontupper=\fontsize{7.5pt}{8.5pt}\selectfont]
\CaseTraceEntry{\texttt{list\_emails(limit=20)} \(\rightarrow\) three relevant
messages and two distractors.}
\CaseTraceEntry{\texttt{read\_email(msg\_1801)} \(\rightarrow\) the urgent
student-disclosure question.}
\CaseTraceEntry{\texttt{read\_email(msg\_1802)} \(\rightarrow\) the unstable
export warning.}
\CaseTraceEntry{\texttt{read\_email(msg\_1803)} \(\rightarrow\) not called;
the concrete division of work and its scheduling constraint are not read.}
\CaseTraceEntry{\texttt{draft\_email(...)} \(\rightarrow\) an escalation brief
is drafted to the advisor.}
\CaseTraceEntry{\texttt{flag\_email(msg\_1801, important)} \(\rightarrow\)
the disclosure message is flagged.}
\end{tcolorbox}

\par\medskip
\noindent\textbf{Outcome.} The resulting brief correctly identifies the privacy risk
and recommends moving the materials from the temporary reference tool to an
approved system that can reliably export files. However, it gives only a
generic instruction to coordinate with the project mentor. It never assigns
the revision to the interns or schedules their work outside Morgan's teaching
commitments. It also drafts an email despite the established instruction to
write a file. The privacy and export requirements pass, the delegation
requirement fails, and the session is unsuccessful under Session SR.
\end{tcolorbox}
\begin{minipage}{\textwidth}
\captionsetup{hypcap=false}
\captionof{figure}{DualMem fragmented-relation failure. Separate retrieval
surfaces the escalation format and adjacent delegation history, yet fails to
connect that history to the third email. The assistant consequently misses the
email that determines who should prepare the revision and when, then produces
an incomplete action plan.}
\label{fig:failure-dualmem-escalation}
\end{minipage}

\begin{tcolorbox}[
    enhanced,
    breakable,
    title=\textbf{Fragmented Relation: Hermes Returns the Wrong Item},
    colframe=black!58,colback=white,colbacktitle=black!78,coltitle=white,
    boxrule=0.35mm,arc=2mm,width=\textwidth,
    fontupper=\small\linespread{0.96}\selectfont,
    left=3mm,right=3mm,top=2mm,bottom=2mm
]
\textbf{Task.} On February 20, 2026, Aino asks the assistant to return exactly
one item from delivered order \#TAU-RET-115 that matches her family-support
context, following the refund process used for an earlier fair-inventory
return.

\textbf{Earlier evidence.}
\begin{itemize}
    \setlength{\itemsep}{1pt}
    \setlength{\parskip}{0pt}
    \setlength{\topsep}{2pt}
    \setlength{\leftmargin}{4mm}
    \item \textbf{January 10, 2026}: Aino disclosed that her mother Helena still
    visits craft fairs, remembers regular customers, and is among the people she
    relies on most.
    \item \textbf{January 10, 2026}: Aino disclosed that her younger brother Eetu
    works in logistics and occasionally helps with van transport on market
    weekends.
    \item \textbf{January 14, 2026}: An earlier fair-inventory return established
    the workflow: read the delivered order, inspect the account's saved payment
    methods, return exactly one matching item, and use the family-market-support
    gift card when available.
\end{itemize}

\textbf{Current application state.} The order contains three plausible
logbooks: Helena--Eetu for market-weekend customer tracking and van loading
(\$43), Jarmo--Helena for studio repairs (\$38), and Veera--gallery team for
gallery scheduling (\$42). The account offers a studio card, a
family-market-support gift card, and a kiln-maintenance card. The correct action
is to return the Helena--Eetu logbook to the family-market-support gift card.

\textbf{Memory state.} Hermes maintains user information in
\texttt{USER.md} and \texttt{MEMORY.md}, indexes prior sessions in
\texttt{state.db}, and stores reusable procedures as skill files.

\begin{tcolorbox}[colback=gray!4,colframe=gray!38,boxrule=0.25mm,arc=0.8mm,
    left=2mm,right=2mm,top=1mm,bottom=1mm]
\ttfamily
\fontsize{7.5pt}{8.5pt}\selectfont
home/\par
\quad|-- memories/\par
\quad|\quad|-- USER.md \quad {\rmfamily\itshape  maintained user profile}\par
\quad|\quad`-- MEMORY.md \quad {\rmfamily\itshape curated long-term summary}\par
\quad|-- state.db \quad {\rmfamily\itshape searchable native-session history}\par
\quad|-- .skills\_prompt\_snapshot.json \quad {\rmfamily\itshape skills exposed to the agent}\par
\quad`-- skills/\par
\qquad\quad|-- apple, autonomous-ai-agents, creative, email\par
\qquad\quad|-- github, media, mlops, note-taking\par
\qquad\quad`-- productivity/retail-operations, research, smart-home,\par
\qquad\qquad social-media, software-development
\end{tcolorbox}

\texttt{USER.md} records Eetu only under \emph{Family} as ``Brother Eetu,''
while \texttt{MEMORY.md} lists Helena only under \emph{Network} alongside
other close contacts. Neither file records that Eetu handles market-weekend
transport with Helena's support. The \texttt{state.db} database stores prior
native-agent sessions. At the target session, the
\texttt{skills/} directory contains 71 bundled skills and five skills created
during this trajectory, including \texttt{retail-operations}.

\par\medskip
\noindent\textbf{Memory access.}
\begin{tcolorbox}[colback=gray!4,colframe=gray!38,boxrule=0.25mm,arc=0.8mm,
    left=2mm,right=2mm,top=1mm,bottom=1mm,
    fontupper=\fontsize{7.5pt}{8.5pt}\selectfont]
\textbf{Automatically injected into the system prompt.}\par
\CaseTraceEntry{\texttt{MEMORY} \(\leftarrow\) the contents of
\texttt{MEMORY.md}, including ``Network: ... Helena'' among several close
contacts.}
\CaseTraceEntry{\texttt{USER PROFILE} \(\leftarrow\) the contents of
\texttt{USER.md}, including ``Family: Brother Eetu''; no stored relation
connects Eetu with Helena.}
\vspace{1mm}
\textbf{Active retrieval.}\par
\CaseTraceEntry{\texttt{skill\_view(retail-operations)}
\(\rightarrow\) read the order, select an item from remembered family context,
inspect saved payment methods, and route the refund to a matching gift card.}
\CaseTraceEntry{\texttt{session\_search(``delivered order return retail
refund pattern'')} \(\rightarrow\) a result dominated by the current
\#TAU-RET-115 request.}
\CaseTraceEntry{\texttt{session\_search(``return\_delivered\_order\_items
payment\_method gift card'')} \(\rightarrow\) the earlier \#TAU-RET-114
execution: ``father Jarmo helping with studio repairs'' and refund to
\texttt{family\_market\_support\_gift\_card}.}
\end{tcolorbox}

Hermes places both maintained Markdown files directly in the system prompt;
the Task LLM therefore sees the isolated Eetu and Helena entries before making
any tool call. \texttt{retail-operations/SKILL.md} supplies the generic return
workflow, while the second \texttt{session\_search} over \texttt{state.db}
surfaces a specific but stale item-selection precedent. Neither access path
recovers that Eetu supports Helena's market-weekend transport.

\par\medskip
\noindent\textbf{Task-LLM execution.}
\begin{tcolorbox}[colback=gray!4,colframe=gray!38,boxrule=0.25mm,arc=0.8mm,
    left=2mm,right=2mm,top=1mm,bottom=1mm,
    fontupper=\fontsize{7.5pt}{8.5pt}\selectfont]
\CaseTraceEntry{\texttt{get\_order\_details(\#TAU-RET-115)}
\(\rightarrow\) delivered order with three candidate logbooks.}
\CaseTraceEntry{\texttt{get\_user\_details(finnish\_ceramic\_artist\_001)}
\(\rightarrow\) studio card, family-market-support gift card, and
kiln-maintenance card.}
\CaseTraceEntry{\texttt{return\_delivered\_order\_items(order=\#TAU-RET-115,
item=Jarmo--Helena logbook, payment=family-market-support gift card)}
\(\rightarrow\) success; return requested.}
\end{tcolorbox}

\par\medskip
\noindent\textbf{Outcome.} The procedural checks pass, but the returned item does not
match the remembered family-support relation. The task completes two of three
required steps, so the session is unsuccessful under Session SR.
\end{tcolorbox}
\begin{minipage}{\textwidth}
\captionsetup{hypcap=false}
\captionof{figure}{Hermes fragmented-relation failure. The agent recovers the
return workflow and separately sees Eetu and Helena in maintained notes, but
does not recover that Eetu supports Helena's market-weekend transport. A stale
earlier return therefore leads it to select the wrong item.}
\label{fig:failure-return}
\end{minipage}

\begin{tcolorbox}[
    enhanced,
    breakable,
    title=\textbf{Retrieval Failure: OpenClaw Cannot Recover a Corridor Collaborator},
    colframe=black!58,colback=white,colbacktitle=black!78,coltitle=white,
    boxrule=0.35mm,arc=2mm,width=\textwidth,fontupper=\small,
    left=3mm,right=3mm,top=2mm,bottom=2mm
]
\textbf{Task.} On October 20, 2026, the user asks the assistant to read
\texttt{\#station-access}, find Darren's message about consent before sharing
rider observations, add the established acknowledgment, and reply using
Darren's remembered role and location.

\textbf{Earlier evidence.}
\begin{itemize}
    \setlength{\itemsep}{1pt}
    \setlength{\parskip}{0pt}
    \setlength{\topsep}{2pt}
    \setlength{\leftmargin}{4mm}
    \item \textbf{February 18, 2026}: A privacy-sensitive Slack acknowledgment
    established the workflow: read the relevant channel, add exactly one
    \texttt{eyes} reaction, and send a brief reply beginning with ``Consent
    check:''. 
    \item \textbf{October 16, 2026}: The user disclosed that Darren is the
    Eugene transit coordinator and had become a close collaborator through
    sustained work on the Portland--Salem--Eugene corridor.
\end{itemize}

Together, the February workflow and October relationship update determine both
the target message and the required acknowledgment.

\textbf{Memory state.} OpenClaw maintains two Markdown files, two native
recall-accounting files, a hybrid SQLite index over the Markdown content, and
a separate archive of session transcripts.

\begin{tcolorbox}[colback=gray!4,colframe=gray!38,boxrule=0.25mm,arc=0.8mm,
    left=2mm,right=2mm,top=1mm,bottom=1mm]
\ttfamily
\fontsize{7.5pt}{8.5pt}\selectfont
workspace/memory/\par
\quad|-- 2026-01-08.md \quad {\rmfamily\itshape dated maintained note}\par
\quad|-- session\_tracking.md \quad {\rmfamily\itshape cross-session summaries}\par
\quad`-- .dreams/\par
\qquad\quad|-- events.jsonl \quad {\rmfamily\itshape recall-event audit log}\par
\qquad\quad`-- short-term-recall.json \quad {\rmfamily\itshape per-chunk recall accounting}\par
openclaw\_memory.sqlite \quad {\rmfamily\itshape hybrid index over maintained notes}\par
\quad|-- files \quad {\rmfamily\itshape indexed-source registry}\par
\quad|-- chunks \quad {\rmfamily\itshape searchable text and metadata}\par
\quad|-- chunks\_fts \quad {\rmfamily\itshape full-text search index}\par
\quad|-- chunks\_vec \quad {\rmfamily\itshape semantic vector index}\par
\quad`-- embedding\_cache \quad {\rmfamily\itshape cached source and query embeddings}\par
openclaw\_home/.openclaw/agents/\textless agent\textgreater/sessions/\par
\quad|-- *.jsonl \quad {\rmfamily\itshape historical session transcripts}\par
\quad`-- sessions.json \quad {\rmfamily\itshape session registry}
\end{tcolorbox}

\texttt{memory/2026-01-08.md} contains entertainment dislikes and an unrelated
ticket rule. \texttt{memory/session\_tracking.md} contains the user's
bus-corridor household, mobility work, and a broad consent boundary, but no
Darren entry, Eugene role, or reusable Slack acknowledgment. Consequently,
\texttt{openclaw\_memory.sqlite} indexes four source chunks that already lack
both target memories.
The separate session archive retains prior interaction records, but those
records are absent from the observed SQLite index and are not accessed in this
task.

\par\medskip
\noindent\textbf{Memory access.}
\begin{tcolorbox}[colback=gray!4,colframe=gray!38,boxrule=0.25mm,arc=0.8mm,
    left=2mm,right=2mm,top=1mm,bottom=1mm,
    fontupper=\fontsize{7.5pt}{8.5pt}\selectfont]
\CaseTraceEntry{\texttt{memory\_search(``close corridor collaborator role
location place'')} \(\rightarrow\) 0 hits.}
\CaseTraceEntry{\texttt{memory\_search(``privacy-forward update community
mobility channel acknowledgment task'')} \(\rightarrow\)
\texttt{memory/2026-01-08.md}, an unrelated entertainment-preference note.}
\CaseTraceEntry{\texttt{memory\_search(``corridor consent accessibility notes
circulated'')} \(\rightarrow\) the same unrelated note.}
\CaseTraceEntry{\texttt{memory\_get(MEMORY.md)} \(\rightarrow\) empty.}
\CaseTraceEntry{\texttt{Get-ChildItem(memory)} \(\rightarrow\)
\texttt{.dreams/}, \texttt{2026-01-08.md}, and
\texttt{session\_tracking.md}.}
\CaseTraceEntry{\texttt{read(memory/session\_tracking.md)} \(\rightarrow\)
the bus-corridor household, mobility meetings, and consent boundary; no Darren,
Eugene, or Slack acknowledgment rule.}
\CaseTraceEntry{\texttt{read(memory/2026-01-08.md)} \(\rightarrow\)
entertainment dislikes and an unrelated ticket-resolution rule.}
\CaseTraceEntry{\emph{Subsequent searches for Darren, Eugene, the corridor
relationship, and the acknowledgment workflow return 0 hits.}}
\end{tcolorbox}

The broad mobility and consent context enters the Task LLM only through
the explicit read of \texttt{memory/session\_tracking.md}. It supports the
topic of Darren's message but cannot identify him as the Eugene transit
coordinator or recover the exact one-reaction, prefixed-reply procedure. The
separate session-search interface is not invoked in this task.

\par\medskip
\noindent\textbf{Task-LLM execution.}
\begin{tcolorbox}[colback=gray!4,colframe=gray!38,boxrule=0.25mm,arc=0.8mm,
    left=2mm,right=2mm,top=1mm,bottom=1mm,
    fontupper=\fontsize{7.5pt}{8.5pt}\selectfont]
\CaseTraceEntry{\texttt{list\_channels()} \(\rightarrow\)
\texttt{\#station-access} and \texttt{\#community-mobility}.}
\CaseTraceEntry{\texttt{read\_messages(channel=\#station-access, limit=20)}
\(\rightarrow\) Darren's consent-sensitive target message and two distractors.}
\CaseTraceEntry{\texttt{read\_messages(channel=\#community-mobility, limit=30)}
\(\rightarrow\) one unrelated coalition reminder.}
\CaseTraceEntry{\texttt{add\_reaction(...)} \(\rightarrow\) not called.}
\CaseTraceEntry{\texttt{send\_message(...)} \(\rightarrow\) not called.}
\end{tcolorbox}

After finding Darren's target message, the agent asks the user for the missing
history rather than performing the required acknowledgment.

\par\medskip
\noindent\textbf{Outcome.} Current application access succeeds while the long-range
personal and procedural evidence remains unavailable. Only the message-reading
step passes.
\end{tcolorbox}
\begin{minipage}{\textwidth}
\captionsetup{hypcap=false}
\captionof{figure}{OpenClaw retrieval failure after roughly nine months.
\texttt{memory/2026-01-08.md} and \texttt{memory/session\_tracking.md} contain
accumulated notes, but repeated searches over their SQLite index do not recover
the two memories needed to act on the current Slack message.}
\label{fig:failure-openclaw}
\end{minipage}

\begin{tcolorbox}[
    enhanced,
    breakable,
    title=\textbf{Update Failure: PraxisMem Omits a Reusable Document Rule},
    colframe=black!58,colback=white,colbacktitle=black!78,coltitle=white,
    boxrule=0.35mm,arc=2mm,width=\textwidth,fontupper=\small,
    left=3mm,right=3mm,top=2mm,bottom=2mm
]
\textbf{Task.} On January 11, 2026, a rural-health career consultant asks the
assistant to find all critical tasks already in progress and create a brief
status document for the coming advising week. The document must connect those
tasks to the consultant's current partner network and schedule.

\textbf{Earlier evidence.}
\begin{itemize}
    \setlength{\itemsep}{1pt}
    \setlength{\parskip}{0pt}
    \setlength{\topsep}{2pt}
    \setlength{\leftmargin}{4mm}
    \item \textbf{January 6, 2026}: The consultant disclosed a new professional
    network centered on first-generation students from rural counties, TRIO
    coordinators, rural healthcare recruiters, and community-college transfer
    advisors.
    \item \textbf{January 6, 2026}: The first task-status workflow established
    that document titles begin with ``Status Snapshot -'' and that
    schedule-sensitive documents contain a section named ``Schedule impact.''
\end{itemize}

\textbf{Current application state.} Three critical tasks concern Tuesday's
Pine Valley advising packets, a shortfall in covered-student advising hours,
and Thursday's transfer-advising documentation protocol.

\textbf{Memory state.} PraxisMem uses a structured Markdown store containing a
current profile, an index, dated events, reusable procedures, and raw source
transcripts.

\begin{tcolorbox}[colback=gray!4,colframe=gray!38,boxrule=0.25mm,arc=0.8mm,
    left=2mm,right=2mm,top=1mm,bottom=1mm]
\ttfamily
\fontsize{7.5pt}{8.5pt}\selectfont
PraxisMem store/\par
\quad|-- USER.md \quad {\rmfamily\itshape current user profile}\par
\quad|-- MEMORY.md \quad {\rmfamily\itshape index linking current values to events}\par
\quad|-- events/ \quad {\rmfamily\itshape dated personal-state changes}\par
\quad|\quad`-- 2026-01-06.md \quad {\rmfamily\itshape dated event record}\par
\quad|-- skills/ \quad {\rmfamily\itshape reusable procedural memories}\par
\quad`-- raw/ \quad {\rmfamily\itshape source transcript archive}\par
\qquad\quad`-- undated.jsonl \quad {\rmfamily\itshape archived source-session dialogue}
\end{tcolorbox}

\texttt{USER.md} records the current professional network---including
first-generation rural students, TRIO coordinators, rural recruiters, and
transfer advisors---and \texttt{events/2026-01-06.md} records the corresponding
change. \texttt{MEMORY.md} links that current value to the event, but contains
no task-status rule, and \texttt{skills/} contains no
\texttt{tasks\_docs/SKILL.md} or equivalent workflow. The source dialogue in
\texttt{raw/undated.jsonl} preserves both ``Status Snapshot -'' and ``Schedule
impact,'' but raw transcripts are excluded from ordinary retrieval and do not
count as organized long-term-memory updates.

\par\medskip
\noindent\textbf{Memory access.}
\begin{tcolorbox}[colback=gray!4,colframe=gray!38,boxrule=0.25mm,arc=0.8mm,
    left=2mm,right=2mm,top=1mm,bottom=1mm,
    fontupper=\fontsize{7.5pt}{8.5pt}\selectfont]
\textbf{Retrieval-agent operations.}\par
\CaseTraceEntry{\texttt{list\_memory\_files(scope=skills)} \(\rightarrow\)
the available skill inventory; no task-status document workflow is listed.}
\CaseTraceEntry{\texttt{read\_memory(
path=skills/docs/tasks/SKILL.md)} \(\rightarrow\) error: unsafe or unsupported
path.}
\CaseTraceEntry{\texttt{read\_memory(path=USER.md)} \(\rightarrow\) the current
user profile.}
\vspace{1mm}
\textbf{Context injected into the Task LLM.}\par
\CaseTraceEntry{\textbf{Core values:} educational access, accuracy, student
confidence, informed consent, and realistic planning.}
\CaseTraceEntry{\textbf{Work style:} structured, barrier-aware, and
documentation-focused.}
\CaseTraceEntry{\textbf{Preferred resources:} offline-capable advising
templates, mobile reference sheets, consent checklists, concise summaries, and
cross-device trackers.}
\end{tcolorbox}

The retrieval agent reads the profile after its attempted skill read fails,
then supplies the three profile excerpts above as a single memory-context block
to the Task LLM. It does not read \texttt{MEMORY.md}, the dated event, or the
raw transcript. Consequently, the injected context contains neither the
task-status title prefix nor the required schedule-section name.

\par\medskip
\noindent\textbf{Task-LLM execution.}
\begin{tcolorbox}[colback=gray!4,colframe=gray!38,boxrule=0.25mm,arc=0.8mm,
    left=2mm,right=2mm,top=1mm,bottom=1mm,
    fontupper=\fontsize{7.5pt}{8.5pt}\selectfont]
\CaseTraceEntry{\texttt{query\_tasks(status=in\_progress, priority=critical)}
\(\rightarrow\) HCC-401, HCC-402, and HCC-403.}
\CaseTraceEntry{\texttt{create\_document(title=``Task-Status Review---Sunday,
January 11, 2026'', ...)} \(\rightarrow\) document created with a section
named ``Schedule Implications.''}
\end{tcolorbox}

\par\medskip
\noindent\textbf{Outcome.} The task selection and personal grounding are useful, but
both exact procedural checks fail: the title lacks ``Status Snapshot -'' and
the document lacks the required ``Schedule impact'' section. Because neither
rule was written to \texttt{MEMORY.md}, \texttt{events/}, or \texttt{skills/},
both remain unrecoverable at inference time.
\end{tcolorbox}
\begin{minipage}{\textwidth}
\captionsetup{hypcap=false}
\captionof{figure}{PraxisMem update failure. Personal context from the source session is
stored in \texttt{USER.md} and \texttt{events/2026-09-07.md} and later
retrieved, while two disclosed procedural rules remain only in
\texttt{raw/} and never enter \texttt{MEMORY.md}, \texttt{events/}, or
\texttt{skills/}.}
\label{fig:failure-status-update}
\end{minipage}

\begin{tcolorbox}[
    enhanced,
    breakable,
    title=\textbf{Competing Memory: A Well-Formed but Mis-personalized Email Digest},
    colframe=black!58,colback=white,colbacktitle=black!78,coltitle=white,
    boxrule=0.35mm,arc=2mm,width=\textwidth,fontupper=\small,
    left=3mm,right=3mm,top=2mm,bottom=2mm
]
\textbf{Task.} On February 11, 2026, the user asks MemOS to read four current
emails and create a Markdown triage digest using the established format and the
relevant personal and work context. The inbox includes an urgent studio note
from roommate Dani, a same-day therapeutic-illustration call, an invitation to
a loud networking mixer, and a licensing newsletter.

\textbf{Earlier evidence.}
\begin{itemize}
    \setlength{\itemsep}{1pt}
    \setlength{\parskip}{0pt}
    \setlength{\topsep}{2pt}
    \setlength{\leftmargin}{4mm}
    \item \textbf{January 7, 2026}: The user disclosed that her long-trusted
    friend Dani had become her roommate after Dani's building was sold.
    \item \textbf{January 13, 2026}: The user identified loud networking mixers
    as an uncomfortable form of professional socializing.
    \item \textbf{January 22, 2026}: The user retained a preference for calm,
    uninterrupted drawing blocks while adding short written response procedures
    for customer and order problems.
    \item \textbf{January 31, 2026}: An earlier inbox-triage task established
    the digest title, the \emph{Urgent Emails}, \emph{Non-Urgent Emails}, and
    \emph{Recommended Actions} sections, sender and subject coverage, and a
    compact priority table.
\end{itemize}

\textbf{Current application state.} The inbox contains Dani's shared-studio
note, a same-day therapeutic-illustration call, an invitation to a loud
networking mixer, and a licensing newsletter. The remembered relationship
makes Dani's note personally important, the user's social preference lowers
the mixer's priority, and the call must be scheduled without fragmenting the
user's focused drawing time.

\textbf{Memory state.} 

\begin{tcolorbox}[colback=gray!4,colframe=gray!38,boxrule=0.25mm,arc=0.8mm,
    left=2mm,right=2mm,top=1mm,bottom=1mm]
\ttfamily
\fontsize{7.5pt}{8.5pt}\selectfont
memory/memos/\par
\quad|-- storage\_context.json \quad {\rmfamily\itshape backend and collection binding}\par
\quad|-- .memos/logs/memos.log \quad {\rmfamily\itshape extraction, update, and search trace}\par
\quad`-- qdrant/ \quad {\rmfamily\itshape persisted vector store}\par
\qquad\quad|-- meta.json \quad {\rmfamily\itshape collection metadata}\par
\qquad\quad`-- collection/openclaw\_memory/storage.sqlite\par
\qquad\qquad`-- {\rmfamily\itshape memory records and their cosine-vector index}\par
record fields \quad {\rmfamily id, text, status, version, history, successor, source, tags, and update time}
\end{tcolorbox}

\par\medskip
\noindent\textbf{Memory access.}
\begin{tcolorbox}[colback=gray!4,colframe=gray!38,boxrule=0.25mm,arc=0.8mm,
    left=2mm,right=2mm,top=1mm,bottom=1mm,
    fontupper=\fontsize{7.5pt}{8.5pt}\selectfont]
\CaseTraceEntry{\textbf{Ranks 1--3:} three near-duplicate records restating the
current request to create an inbox digest using earlier personal and work
context.}
\CaseTraceEntry{\textbf{Rank 4:} January 7 digest instruction---write
\texttt{s015\_email\_triage\_digest.md} with urgent, non-urgent, and recommended
action sections plus sender and subject coverage.}
\CaseTraceEntry{\textbf{Rank 6:} ``preference for automatic reassignment without
reviewing relationship or workload context.''}
\CaseTraceEntry{\textbf{Rank 7:} earlier inbox-triage convention---include the
digest title, three required sections, sender addresses, subject lines, and
remembered relationship/work-rhythm context.}
\CaseTraceEntry{\textbf{Rank 16:} preference for visual examples, short
checklists, reusable templates, and compact triage tables with priority labels.}
\CaseTraceEntry{\emph{The remaining 13 records concern other task reminders,
Slack updates, travel-budget workflow, and unrelated history.}}
\end{tcolorbox}

The top-20 retrieval recovers
useful output structure and one directly competing preference. None of its
returned records identifies Dani as the close roommate or the need to protect calm drawing blocks.

\par\medskip
\noindent\textbf{Task-LLM execution.}
\begin{tcolorbox}[colback=gray!4,colframe=gray!38,boxrule=0.25mm,arc=0.8mm,
    left=2mm,right=2mm,top=1mm,bottom=1mm,
    fontupper=\fontsize{7.5pt}{8.5pt}\selectfont]
\CaseTraceEntry{\texttt{list\_emails(limit=50)} \(\rightarrow\)
four current messages.}
\CaseTraceEntry{\texttt{read\_email(msg\_s024\_001)} \(\rightarrow\)
Dani's urgent studio-corner note.}
\CaseTraceEntry{\texttt{read\_email(msg\_s024\_002)} \(\rightarrow\)
loud networking-mixer invitation.}
\CaseTraceEntry{\texttt{read\_email(msg\_s024\_003)} \(\rightarrow\)
same-day therapeutic-illustration call.}
\CaseTraceEntry{\texttt{read\_email(msg\_s024\_004)} \(\rightarrow\)
licensing newsletter.}
\CaseTraceEntry{\texttt{write(tester\_artifacts/s024\_email\_triage\_digest.md)}
\(\rightarrow\) Markdown digest created successfully.}
\end{tcolorbox}

The artifact explicitly cites the conflicting preference, recommends automatic
reassignment for Dani and the client call, and treats the mixer as a
context-free yes/no decision.

\par\medskip
\noindent\textbf{Outcome.} Every programmatic and procedural assertion passes, including
the filename, headings, sender/subject coverage, and priority column. All three
personal semantic slots score zero. The artifact is structurally convincing,
yet its recommendations apply the wrong user model.
\end{tcolorbox}
\begin{minipage}{\textwidth}
\captionsetup{hypcap=false}
\captionof{figure}{MemOS competing-memory failure. Retrieved procedural structure yields
a polished artifact, while an incompatible personal record reverses how the
current emails should be prioritized.}
\label{fig:failure-email}
\end{minipage}

\clearpage
\makeatletter
\setlength{\@fptop}{0pt plus 1fil}
\setlength{\@dblfptop}{0pt plus 1fil}
\makeatother

\subsection{Sensitivity to the Task-Agent Backbone}
\label{app:task-llm-sensitivity}

To examine whether the benchmark conclusions depend on the task-agent backbone,
we rerun the Scheduled setting on the first eight test trajectories with three
task LLMs: Qwen3.7-Plus, GLM-5.2, and DeepSeek-V4-Pro. All other settings,
including the memory method, user simulator, judge, verifier, and trajectory
subset, remain unchanged.

\begin{table}[t]
    \centering
    \caption{Session SR under different task-agent backbones on eight fixed
    test trajectories in the Scheduled setting.}
    \label{tab:task-llm-sensitivity}
    \small
    \setlength{\tabcolsep}{5pt}
    \begin{tabular}{@{}lccc@{}}
        \toprule
        Memory method & Qwen3.7-Plus & GLM-5.2 & DeepSeek-V4-Pro \\
        \midrule
        OpenClaw native memory & 10.0 & 26.4 & 31.0 \\
        MemOS                  & 18.1 & 22.6 & 24.2 \\
        \textsc{PraxisMem}     & \textbf{43.1} & \textbf{46.7} & \textbf{49.8} \\
        \bottomrule
    \end{tabular}
\end{table}

As shown in Table~\ref{tab:task-llm-sensitivity}, stronger task agents improve
all three systems, but the sensitivity differs substantially. OpenClaw native
memory varies by 21.0 Session-SR points across backbones, consistent with its
task LLM jointly handling memory maintenance, retrieval, and task execution.
In contrast, MemOS and \textsc{PraxisMem}, whose memory modules remain fixed
across task agents, vary by only 6.1 and 6.7 points, respectively.
\textsc{PraxisMem} also remains the strongest method under all three task
backbones. These results suggest that the main advantage of
\textsc{PraxisMem} is not specific to the Qwen3.7-Plus task agent.

\subsection{Runtime and Token Efficiency}
\label{app:runtime-efficiency}

Persistent personal assistants operate inside general-purpose harnesses over
repeated sessions, making both monetary cost and memory-maintenance latency
practical constraints. Task-agent inference accumulates API charges throughout
the trajectory, while constructing and updating memory consumes additional
compute and determines how quickly newly disclosed information becomes
available to later tasks. We therefore evaluate cost and latency alongside
Session SR.

We record task-agent calls, input and output tokens, and wall-clock time for
each executed session. Task-LLM API cost uses the standard international
real-time Qwen3.7-Plus rates of USD~0.40 per million input tokens and USD~1.60
per million output tokens \citep{alibaba2026qwenpricing}. Memory-GPU cost is
computed from measured update or construction time using an accounting rate of
USD~1.59 per A100 GPU-hour. Complete-split cost is reported over all 20 test
trajectories. For ReMe, we report the cost and latency of its memory
construction stage. For harness-native memory, whose maintenance time is
interleaved with task execution, we estimate memory latency as the excess
wall-clock time over the average runtime of stable task-LLM baselines. Overall
Session SR follows the pooled aggregation in
Eq.~\ref{eq:session-sr-aggregation}. Detailed task-agent resource statistics are reported in
Table~\ref{tab:integrated-runtime}, while Table~\ref{tab:memory-runtime}
separately reports observable memory-side update or construction cost, latency,
and retrieval context.

\begin{figure}[t]
    \centering
    \begin{subfigure}[t]{0.48\linewidth}
        \centering
        \includegraphics[width=\linewidth]{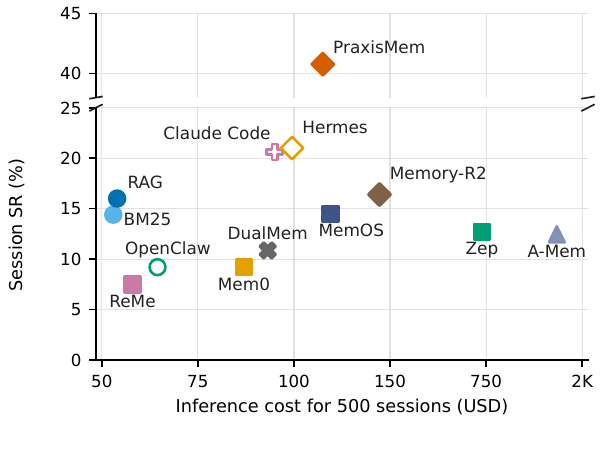}
        \caption{Inference cost}
    \end{subfigure}\hfill
    \begin{subfigure}[t]{0.48\linewidth}
        \centering
        \includegraphics[width=\linewidth]{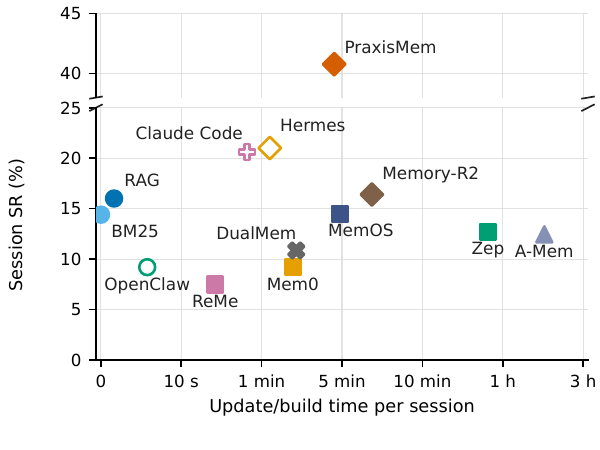}
        \caption{Memory update latency}
    \end{subfigure}
\caption{Effectiveness and runtime trade-offs.
(a) Session SR versus inference cost for completing the 500-session test split,
including Task-LLM API and memory-GPU costs.
(b) Session SR versus per-session memory update/build latency; harness-native
memory latency is estimated from excess wall-clock time over stable task-LLM
baselines.}
    \label{fig:runtime-tradeoffs}
\end{figure}

Figure~\ref{fig:runtime-tradeoffs} shows that higher memory cost does not imply
better persistent assistance. Zep performs 425.29 Memory-LLM calls and requires
approximately 0.85 hours per session to construct and update its temporal
graph, while A-Mem performs 1,397.38 calls and requires approximately 2.03
hours through its neighbor-based memory evolution and update chain. Such
latencies are difficult to accommodate in a practical harness-based personal
assistant, yet their Session SRs remain only 12.70\% and 12.45\%,
respectively. Harness-native systems offer a better trade-off, although their
gains over lightweight retrieval remain limited: Embedding RAG reaches
16.00\% Session SR at an estimated \$54 for the complete test split, compared
with 20.60\% at \$95 for Claude Code and 21.00\% at \$99.5 for Hermes.
In contrast, \textsc{PraxisMem} achieves substantially higher Session SR with
modest additional cost and memory-maintenance latency, yielding a stronger
cost--effectiveness trade-off.

\begin{table}[t]
    \centering
    \caption{Per-session task-agent resources across memory settings.
    Input and output tokens include all Qwen3.7-Plus calls made by the task agent,
    with API cost computed at USD~0.40 per million input tokens and USD~1.60 per
    million output tokens. Memory-GPU cost uses a single A100 GPU at
    USD~1.59 per GPU-hour.}
    \label{tab:integrated-runtime}
    \footnotesize
    \setlength{\tabcolsep}{3.75pt}
    \providecommand{\tbdscore}{\makebox[2.05em][c]{\textcolor{gray}{--}}}
    \begin{tabular*}{\linewidth}{@{\extracolsep{\fill}}lrrrrrrr@{}}
        \toprule
        Method & Calls & Input tok. & Output tok. & Time (s) & Task API (\$) & Mem. GPU (\$) & Session SR \\
        \midrule
        \rowcolor{gray!12}
        \multicolumn{8}{@{}l}{\textit{Diagnostic}} \\
        Oracle & 7.66 & 268.51K & 5.27K & 87.55  & 0.116 & -- & 86.20 \\
        Long Context & 7.61 & 473.03K & 5.56K & 99.79  & 0.198 & -- & 27.40 \\
        DualMem & 7.83 & 264.86K & 5.09K & 75.24 & 0.114 & 0.072 & 10.86 \\
        \midrule
        \rowcolor{gray!12}
        \multicolumn{8}{@{}l}{\textit{Basic baselines}} \\
        BM25          & 7.32 & 243.94K & 4.95K & 84.46 & 0.106 & -- & 14.40 \\
        Embedding RAG & 7.45 & 249.22K & 4.94K & 89.71 & 0.108 & -- & 16.00 \\
        \midrule
        \rowcolor{gray!12}
        \multicolumn{8}{@{}l}{\textit{Harness-native memory}} \\
        OpenClaw    & 9.22 & 302.42K & 4.84K & 95.62  & 0.129 & -- & 9.20 \\
        Hermes      & 15.83 & 464.60K & 8.18K & 174.24 & 0.199 & -- & 21.00 \\
        Claude Code & 16.99 & 449.18K & 6.72K & 140.76 & 0.190 & -- & 20.60 \\
        \midrule
        \rowcolor{gray!12}
        \multicolumn{8}{@{}l}{\textit{Advanced memory systems}} \\
        Mem0  & 7.57 & 245.43K & 5.08K & 93.24 & 0.106 & 0.068 & 9.20 \\
        Zep & 7.38 & 234.72K & 4.81K & 99.79 & 0.102 & 1.346 & 12.70 \\
        A-Mem & 7.45 & 247.37K & 5.03K & 93.89 & 0.107 & 3.233 & 12.45 \\
        ReMe  & 7.31 & 234.39K & 4.87K & 81.79 & 0.102 & 0.014 & 7.50 \\
        MemOS & 7.52 & 249.15K & 5.20K & 89.35 & 0.108 & 0.130 & 14.46 \\
        \midrule
        \rowcolor{gray!12}
        \multicolumn{8}{@{}l}{\textit{Learned memory managers}} \\
        Memory-R2 & 7.42 & 247.82K & 5.14K & 87.48 & 0.107 & 0.182 & 16.40 \\
        PraxisMem (Ours)
        & 7.47 & 248.87K & 5.16K & 88.91 & 0.108 & 0.122 & 40.78 \\
        \bottomrule
    \end{tabular*}
\end{table}

\begin{table}[t]
    \centering
\caption{Per-session resources of separately observable memory methods,
averaged over executed sessions. Input and output tokens refer to the
generative Memory LLM, while context tokens measure memory text returned to the task agent. For ReMe, we report the resources and latency of its memory
construction stage.}
    \label{tab:memory-runtime}
    \small
    \begin{tabular*}{\linewidth}{@{\extracolsep{\fill}}lrrrrrrr@{}}
        \toprule
        Method & Ops. & Calls & Input tok. & Output tok. & Time (s) & Context tok. & Session SR \\
        \midrule
        \rowcolor{gray!12}
        \multicolumn{8}{@{}l}{\textit{Diagnostic}} \\
        Oracle        & 1.00 & --   & --     & --     & 0.00    & 1.14K  & 86.20 \\
        Long Context  & 1.00 & --   & --     & --     & 0.00    & 81.44K & 27.40 \\
        DualMem       & 14.07 & 13.99 & 125.97K & 8.07K & 163.58 & 3.39K & 10.86 \\
        \midrule
        \rowcolor{gray!12}
        \multicolumn{8}{@{}l}{\textit{Retrieval baselines}} \\
        BM25          & 1.00 & --   & --     & --     & 0.02    & 3.88K & 14.40 \\
        Embedding RAG & 1.00 & --   & --     & --     & 1.64    & 3.87K & 16.00 \\
        \midrule
        \rowcolor{gray!12}
        \multicolumn{8}{@{}l}{\textit{Advanced memory systems}} \\
        Mem0          & 6.96  & 6.96   & 84.89K & 8.12K  & 153.22  & 1.55K & 9.20 \\
        Zep           & 60.32 & 425.29 & 1.28M  & 87.07K & 3046.82 & 0.98K & 12.70 \\
        A-Mem         & 1.00  & 1397.38 & 1.24M  & 108.85K & 7320.53 & 3.44K & 12.45 \\
        ReMe & 3.13 & 2.89 & 8.42K & 1.77K & 31.18 & 1.03K & 7.50 \\
        MemOS         & 6.99  & 6.21   & 29.35K & 4.69K  & 294.12  & 2.67K & 14.46 \\
        \midrule
        \rowcolor{gray!12}
        \multicolumn{8}{@{}l}{\textit{Learned memory managers}} \\
        Memory-R2     & 7.06  & 10.99  & 73.95K  & 28.91K & 411.62 & 3.13K & 16.40 \\
        PraxisMem (Ours) & 7.33 & 35.82 & 262.17K & 5.18K & 276.62 & 0.34K & 40.78 \\
        \bottomrule
    \end{tabular*}
\end{table}

\section{Prompt Templates}
\label{app:prompts}

We collect the prompts used across benchmark construction, memory management,
task execution, and evaluation in this section. Braced expressions denote
instance-specific inputs.

\begin{minipage}{\linewidth}
\begin{tcolorbox}[
        colback=gray!6,
        colframe=gray!45,
        boxrule=0.5pt,
        arc=1pt,
        left=6pt,
        right=6pt,
        top=6pt,
        bottom=6pt]
    \small
    You are constructing a stable user persona for a persistent personal-agent
    benchmark. Expand the supplied persona seed into a realistic, specific, and
    internally consistent profile. Preserve the central identity expressed by
    the seed while adding enough detail to support a long sequence of personal,
    tool-use, and artifact-creation interactions.

    \medskip
    \textbf{Input}

    \texttt{source\_id}: \{source\_id\}\\
    \texttt{seed\_text}: \{seed\_text\}\\
    \texttt{persona\_id}: \{persona\_id\}

    \medskip
    \textbf{Output format}

    Return only valid YAML. Preserve the input metadata and organize the profile
    under exactly these six top-level groups:

    \texttt{basic\_identity}: full name, age, gender, location, languages,
    education level, occupation, income band, and cultural background.\\
    \texttt{professional\_and\_daily\_role}: job title, sector,
    responsibilities, common tasks, work-time structure, digital-tool
    proficiency, and frequent collaborators.\\
    \texttt{personality\_and\_values}: social orientation, core values, work
    style, planning tendency, spending style, privacy awareness, sense of
    responsibility, \texttt{positive\_traits}, \texttt{neutral\_tendencies},
    and \texttt{negative\_traits}.\\
    \texttt{relationship\_network}: household, extended family, friends,
    colleagues, managers, students, community contacts, frequent contacts, and
    relationship strengths.\\
    \texttt{preferences}: food, entertainment, communication, learning, work,
    shopping, health, sleep and schedule, privacy, and aversions.\\
    \texttt{constraints\_and\_boundaries}: budget, time, health, family,
    geographic, device, policy and compliance, and career constraints.
\end{tcolorbox}
\captionsetup{hypcap=false}
\captionof{figure}{Prompt template for expanding a PersonaHub seed into a
structured stable persona.}
\label{fig:persona-enrichment-prompt}
\end{minipage}

\begin{tcolorbox}[
        breakable,
        colback=gray!6,
        colframe=gray!45,
        boxrule=0.5pt,
        arc=1pt,
        left=6pt,
        right=6pt,
        top=6pt,
        bottom=6pt]
    \small
    \textbf{System message}\par
    You synthesize one persona change event for a long-term agent memory
    benchmark. Return only valid JSON. Do not include markdown.

    \medskip
    \textbf{User message}\par
    We are building one benchmark session.

    \medskip
    \textbf{Inputs}\par
    Persona basic identity, provided only for plausibility checks:
    \texttt{\{basic\_identity\}}.\par
    Current persona snapshot, provided for conflict checks:
    \texttt{\{persona\_snapshot\}}.\par
    Target persona field: \texttt{field\_path = \{field\_path\}};
    \texttt{current\_value = \{current\_value\}}.\par
    Optional tool/SOP context: \texttt{domain\_id = \{domain\_id\}},
    \texttt{capability = \{capability\}}, \texttt{sop\_id = \{sop\_id\}},
    \texttt{source\_paths = \{source\_paths\}}.\par
    Conversation time: \texttt{\{session\_time\_tag\}}.\par
    Already happened events: \texttt{\{past\_events\}}.\par
    Previous rejected attempt feedback: \texttt{\{repair\_feedback\}}.

    \medskip
    \textbf{Task}\par
    \begin{enumerate}
        \item Propose one new value for the exact same target persona field.
        Write one concrete event that naturally explains the change from the
        current value to the new value.
        \item Use the full persona snapshot to avoid contradictions with the
        user's other traits, preferences, constraints, relationships, and role
        details. Make the event compatible with the optional tool/SOP context
        if possible. If the sampled field cannot naturally fit that context,
        set \texttt{needs\_resample} to \texttt{true} and explain briefly.
        \item Make the new value a substantive change that affects future
        assistant behavior visibly. Preserve the current value's JSON type;
        for an object, preserve exactly the same keys. Do not keep the old
        value wholesale and append minor details. Prefer a meaningful
        redirect, reprioritization, narrowing, or restructuring while
        preserving persona continuity. If an additive change is necessary,
        explain in the rationale how it changes future behavior.
        \item Treat the conversation time as the time of this session; the
        event may have occurred shortly beforehand. Keep the event natural,
        concrete, and not overly dramatic. Do not mention tool names or reveal
        the user's name, age, gender, or occupation.
        \item Do not copy the current value, new value, or any complete persona
        field value verbatim into the event, and do not list field contents
        inside it. Write \texttt{event\_summary} from the user's first-person
        perspective, using ``I,'' ``my,'' or ``me'' rather than third-person
        pronouns for the user.
        \item If feedback from a rejected attempt is present, fix the stated
        issue without repeating the rejected pattern.
    \end{enumerate}

    \medskip
    \textbf{Output format}\par
    Return JSON with exactly these fields:
    \texttt{needs\_resample} (Boolean),
    \texttt{field\_path} (the sampled field),
    \texttt{old\_value} (the supplied current value),
    \texttt{new\_value} (the same JSON type as \texttt{old\_value}),
    \texttt{event\_summary} (one natural sentence), and
    \texttt{rationale} (one short sentence).
\end{tcolorbox}
\begin{minipage}{\linewidth}
\captionsetup{hypcap=false}
\captionof{figure}{Prompt template for synthesizing a persona value change
and the event that explains it.}
\label{fig:event-synthesis-prompt}
\end{minipage}

\begin{tcolorbox}[
        breakable,
        colback=gray!6,
        colframe=gray!45,
        boxrule=0.5pt,
        arc=1pt,
        left=6pt,
        right=6pt,
        top=6pt,
        bottom=6pt]
    \small
    You are simulating the user in a benchmark session. Generate exactly
    one next user response as JSON.

    \medskip
    \textbf{User profile}\par
    Name: \texttt{\{full\_name\}}; age: \texttt{\{age\}}; gender:
    \texttt{\{gender\}}; occupation: \texttt{\{occupation\}}.\par
    User simulator speaking style: \texttt{\{speaking\_style\}}.

    \medskip
    \textbf{Turn strategies}\par
    Previous turn strategy: \texttt{\{previous\_strategy\}}.\par
    Current turn strategy: \texttt{\{current\_strategy\}}.

    \medskip
    \textbf{Rules}\par
    \begin{itemize}
        \item Only speak as the simulated user. Follow only the active
        strategy selected by the action rules below; do not add another
        session task.
        \item Use only the information given in this prompt. Do not invent
        names, dates, IDs, emails, events, or preferences. Use the speaking
        style only to shape wording and tone.
        \item You may briefly acknowledge assistant suggestions, but do
        not adopt them as new personal facts, preferences, commitments, or
        future plans.
        \item When a turn discloses a memory point, disclose only the fact
        or change specified by the active strategy. If it specifies an
        earlier value, a change event or reason, and a current value,
        include all three.
        \item Preserve the active strategy's required content, including
        quoted text, tool or parameter names, exact phrases, headings,
        channels, statuses, priorities, formats, and item counts. Do not
        omit, weaken, paraphrase away, or merge task details.
        \item Do not introduce, preview, infer, or merge content from a
        future turn. Do not infer the overall session goal from the
        profile, speaking style, or conversation history.
        \item Make the wording natural while preserving every explicit
        non-memory requirement from the active strategy. The recent
        conversation is provided as native chat messages after this
        system prompt, with at most the most recent
        \texttt{\{recent\_rounds\}} rounds.
        \item The only valid action values are \texttt{maintain} and
        \texttt{advance}. If the previous turn strategy is None, choose
        \texttt{advance} and follow the current strategy. Choose
        \texttt{maintain} if the assistant did not satisfy the previous
        strategy, then follow that previous strategy. Otherwise choose
        \texttt{advance} and follow the current strategy. If the current
        strategy ends the session, choose \texttt{advance} and leave
        content empty.
    \end{itemize}

    \medskip
    \textbf{Output JSON only}\par
    Return \texttt{action} (\texttt{maintain} or \texttt{advance}) and
    \texttt{content} (a string).
\end{tcolorbox}
\begin{minipage}{\linewidth}
\captionsetup{hypcap=false}
\captionof{figure}{Prompt template for the user simulator's next turn.}
\label{fig:user-simulator-prompt}
\end{minipage}

\begin{tcolorbox}[
        breakable,
        colback=gray!6,
        colframe=gray!45,
        boxrule=0.5pt,
        arc=1pt,
        left=6pt,
        right=6pt,
        top=6pt,
        bottom=6pt]
    \small
    \textbf{System message}\par
    You are a long-term memory manager for a personal assistant. Use only the
    provided session context and the memory files you inspect with tools.
    Maintain concise Markdown memory files for personal facts, event timelines,
    and reusable skills. Do not invent hidden labels, internal identifiers,
    grading information, or source IDs. Use tools to inspect, edit, and save
    memory. Do not output JSON patches in the final response. Call
    \texttt{save\_raw(date)} exactly once when you can infer the session date;
    use \texttt{undated} only if no date is visible.

    \medskip
    \textbf{Memory store format}\par
    \begin{itemize}
        \item \texttt{USER.md} stores current stable personal facts and
        preferences. Use \texttt{\#\# snake\_case\_topic} sections followed by
        concise prose.
        \item \texttt{MEMORY.md} indexes topics to current state and timeline.
        Each \texttt{\#\# snake\_case\_topic} section contains
        \texttt{current: USER.md\#topic}, one or more chronological
        \texttt{timeline: events/YYYY-MM-DD.md\#topic} links, and short
        lowercase \texttt{tags}.
        \item \texttt{events/YYYY-MM-DD.md} stores dated changes. Each topic
        may contain \texttt{value}, \texttt{previous\_value}, \texttt{summary},
        \texttt{task}, \texttt{skill}, \texttt{notes}, and \texttt{tags};
        include only fields supported by the visible interaction.
        \item \texttt{skills/<tool\_area>/SKILL.md} stores reusable procedural
        memory under \texttt{\# tool\_area}, \texttt{\#\# Domain Rules}, and
        \texttt{\#\# Workflows}, with named workflows and numbered steps.
        \item \texttt{raw/YYYY-MM-DD.jsonl} is maintained only through
        \texttt{save\_raw(date)} and must not be edited directly.
    \end{itemize}
    \texttt{USER.md} and event headings use stable snake-case topic IDs.
    \texttt{MEMORY.md} sections contain only \texttt{current},
    \texttt{timeline}, and \texttt{tags}; skills are not stored there.
    \texttt{SKILL.md} contains reusable procedures without personal facts.
    Use dates and tool-area names only when visible in the interaction or tool
    results. Do not write experiment, session, task, source, or grading IDs.

    \medskip
    \textbf{Grep search tool}\par
    Use \texttt{grep\_memory(pattern, scope, match\_mode, case\_sensitive,
    context, max\_count)} for lexical search. Regex and literal modes correspond
    to \texttt{grep -E} and \texttt{grep -F}; case-insensitive search corresponds
    to \texttt{grep -i}; context corresponds to \texttt{grep -C}; and
    \texttt{max\_count} corresponds to \texttt{grep -m} per file. If a result
    is relevant, call \texttt{read\_memory(path, start\_line, end\_line)} to
    inspect the surrounding Markdown before acting.

    \medskip
    \textbf{Retention gate}\par
    Promote information only when it is a durable fact or preference about the
    current user, a reusable assistant workflow or user-specific procedure, or
    an outcome from a task performed for the current user. Do not promote
    quoted or forwarded third-party conversations, stories about other people,
    general knowledge, math or code exercises, or transient discussion.
    Third-party details remain third-party details when the user shares, reads,
    summarizes, or asks about them. If the visible chunk contains no durable
    information, call \texttt{save\_raw(date)} once and finish immediately
    without reading or editing structured memory files.

    \medskip
    \textbf{User message}\par
    Session work state from earlier chunks:
    \texttt{\{session\_work\_state\}}.\par
    Visible interaction context:
    \texttt{\{visible\_interaction\}}.\par
    Optional visible tool calls and results:
    \texttt{\{tool\_events\}}.

    \medskip
    Update the Markdown memory store. Use \texttt{save\_raw(date)} after
    identifying the date from the visible session context. Allowed editable
    memory files are \texttt{USER.md}, \texttt{MEMORY.md},
    \texttt{events/YYYY-MM-DD.md}, \texttt{events/undated.md}, and
    \texttt{skills/<tool\_area>/SKILL.md}. Use
    \texttt{edit\_memory(command=``create'' $|$ ``str\_replace'' $|$
    ``insert'')} to create or modify files.
\end{tcolorbox}
\begin{minipage}{\linewidth}
\captionsetup{hypcap=false}
\captionof{figure}{System and user prompt templates for \textsc{PraxisMem}
memory update.}
\label{fig:memrsd-update-prompt}
\end{minipage}

\begin{tcolorbox}[
        breakable,
        colback=gray!6,
        colframe=gray!45,
        boxrule=0.5pt,
        arc=1pt,
        left=6pt,
        right=6pt,
        top=6pt,
        bottom=6pt]
    \small
    \textbf{System message}\par
    You are a long-term memory retrieval module. Use tools to inspect the
    Markdown memory store and find evidence for the current request. The final
    response must contain only concise bullets in the form
    \texttt{- label: text}. Do not include headings, explanations, hidden
    labels, internal identifiers, source IDs, or irrelevant history.

    \medskip
    \textbf{Memory store format}\par
    \begin{itemize}
        \item \texttt{USER.md} stores current stable personal facts and
        preferences in \texttt{\#\# snake\_case\_topic} sections.
        \item \texttt{MEMORY.md} indexes each topic to its current
        \texttt{USER.md} section and chronologically ordered event links.
        \item \texttt{events/YYYY-MM-DD.md} stores dated values, previous
        values, summaries, tasks, skill links, notes, and tags when supported by
        the interaction.
        \item \texttt{skills/<tool\_area>/SKILL.md} stores domain rules and
        named reusable workflows with numbered steps.
        \item \texttt{raw/YYYY-MM-DD.jsonl} contains archived source
        interactions and is accessed only through an explicit raw scope.
    \end{itemize}
    \texttt{USER.md} and event headings use stable snake-case topic IDs;
    \texttt{MEMORY.md} contains only current links, timeline links, and tags;
    and skill files contain reusable procedures without personal facts. Do not
    use internal experiment, session, task, source, or grading identifiers.

    \medskip
    \textbf{Grep search tool}\par
    Use \texttt{grep\_memory(pattern, scope, match\_mode, case\_sensitive,
    context, max\_count)} for lexical search. Regex and literal modes correspond
    to \texttt{grep -E} and \texttt{grep -F}; case-insensitive search corresponds
    to \texttt{grep -i}; context corresponds to \texttt{grep -C}; and
    \texttt{max\_count} corresponds to \texttt{grep -m} per file. Relevant
    matches must be followed by \texttt{read\_memory} on the corresponding
    file or line range.

    \medskip
    \textbf{User message}\par
    Final user request:
    \texttt{\{current\_request\}}.

    \medskip
    Use memory tools to find only the information needed for this request.
    Return all necessary memory that fits within the available context budget.
\end{tcolorbox}
\begin{minipage}{\linewidth}
\captionsetup{hypcap=false}
\captionof{figure}{System and user prompt templates for \textsc{PraxisMem}
memory retrieval.}
\label{fig:memrsd-retrieval-prompt}
\end{minipage}

\begin{tcolorbox}[
        breakable,
        colback=gray!6,
        colframe=gray!45,
        boxrule=0.5pt,
        arc=1pt,
        left=6pt,
        right=6pt,
        top=6pt,
        bottom=6pt]
    \small
    You are a strict evaluator for personalized memory points reflected in
    tool arguments.

    \medskip
    For each item, compare the actual memory-bearing open-text tool arguments
    against the gold memory point. Each item is one personalized memory point.
    Multiple memory points may be reflected in the same tool argument.

    \medskip
    \textbf{Scoring}\par
    First identify every required semantic component in the gold memory
    point.\par
    2: The actual arguments cover all required semantic components. Exact
    wording is not required.\par
    1: The actual arguments cover a strict majority of the required semantic
    components, but omit at least one important component.\par
    0: The actual arguments cover half or fewer of the required semantic
    components, are merely topically related or vague, miss the point,
    contradict it, or use the wrong remembered user fact.\par
    A memory point with only one atomic semantic component can receive only
    2 or 0.

    \medskip
    \textbf{Rules}\par
    Evaluate only personalized memory points; do not evaluate procedural
    memory such as fixed section titles, default formatting rules, or
    workflow habits.\par
    Evaluate only the listed memory-bearing tool arguments. Structured
    identifiers and selections are scored deterministically before this
    judge is called; do not reinterpret them.\par
    Do not award credit for weak implication or general discussion of the
    same topic.\par
    If the task is to create a public-safe artifact or message, a safe
    category-level reference can fully satisfy a private memory point when
    it preserves the same operational boundary without exposing the exact
    private wording.\par
    Do not use outside knowledge. Return exactly one result per item,
    preserving \texttt{item\_id}.

    \medskip
    \textbf{Items}: \texttt{\{items\_json\}}\par
    Return only valid JSON with an \texttt{items} array. Each record contains
    \texttt{item\_id}, \texttt{score} (0, 1, or 2), and a brief \texttt{reason}.
\end{tcolorbox}
\begin{minipage}{\linewidth}
\captionsetup{hypcap=false}
\captionof{figure}{Judge prompt for personal-memory points in tool arguments.}
\label{fig:judge-tool-memory}
\end{minipage}

\begin{tcolorbox}[
        breakable,
        colback=gray!6,
        colframe=gray!45,
        boxrule=0.5pt,
        arc=1pt,
        left=6pt,
        right=6pt,
        top=6pt,
        bottom=6pt]
    \small
    You are a strict artifact evaluator. Evaluate whether the produced
    artifact satisfies each gold artifact slot.

    \medskip
    \textbf{Scoring rubric for each slot}\par
    First identify every required semantic component in the gold slot.\par
    2: The artifact covers all required semantic components.\par
    1: The artifact covers a strict majority of the required semantic
    components, but omits at least one important component.\par
    0: The artifact covers half or fewer of the required semantic
    components, is merely topically related or vague, is missing, is
    contradicted, or does the opposite.\par
    A slot with only one atomic semantic component can receive only 2 or 0.

    \medskip
    \textbf{Rules}\par
    Evaluate only the artifact content, not the assistant's surrounding
    chat. Semantic matches are acceptable; exact wording is not required.\par
    Do not award credit for weak implication or general discussion of the
    same topic. Do not use outside knowledge.\par
    Treat fields such as \texttt{target\_atomic\_fact},
    \texttt{fact\_path}, or \texttt{artifact\_path} as identifiers and
    location hints only. Values like \texttt{0} or \texttt{1} are list
    indexes, not Boolean instructions or positive/negative labels.\par
    Return exactly one result per gold slot, preserving \texttt{slot\_id}.

    \medskip
    \textbf{Gold artifact slots}: \texttt{\{gold\_slots\_json\}}\par
    \textbf{Artifact}: \texttt{\{artifact\_json\}}\par
    Return only valid JSON with a \texttt{slots} array. Each record contains
    \texttt{slot\_id}, \texttt{score} (0, 1, or 2), and a brief \texttt{reason}.
\end{tcolorbox}
\begin{minipage}{\linewidth}
\captionsetup{hypcap=false}
\captionof{figure}{Judge prompt for semantic personal-memory slots in
artifacts.}
\label{fig:judge-artifact-slots}
\end{minipage}

\begin{tcolorbox}[
        breakable,
        colback=gray!6,
        colframe=gray!45,
        boxrule=0.5pt,
        arc=1pt,
        left=6pt,
        right=6pt,
        top=6pt,
        bottom=6pt]
    \small
    You are a strict memory recall evaluator. Evaluate whether the
    assistant's final summary covers each gold memory point.

    \medskip
    \textbf{Scoring rubric for each gold memory point}\par
    2: Fully covered or logically implied by one or more assistant
    bullets.\par
    1: Partially covered, but some important value or detail is missing,
    uncertain, or slightly wrong.\par
    0: Not mentioned, contradicted, or matched to the wrong current value.

    \medskip
    \textbf{Rules}\par
    Semantic matches are acceptable; exact wording is not required.
    Natural phrasing, harmless qualifiers, or non-conflicting casual
    wording should not reduce the score.\par
    A gold point may ask for a current value, historical value,
    intermediate value, or change event. Judge the exact
    \texttt{target\_value\_text} and \texttt{point\_type} in that gold
    point.\par
    Extra assistant bullets do not affect recall. Do not use outside
    knowledge. Return exactly one result per gold point, preserving
    \texttt{gold\_index} exactly.

    \medskip
    \textbf{Gold memory points}: \texttt{\{gold\_points\_json\}}\par
    \textbf{Assistant bullets}: \texttt{\{candidate\_bullets\_json\}}\par
    \textbf{Assistant final answer}: \texttt{\{assistant\_answer\}}\par
    Return only valid JSON with a \texttt{points} array. Each record contains
    \texttt{gold\_index}, the original \texttt{field\_path},
    \texttt{score} (0, 1, or 2), \texttt{matched\_bullet\_indexes}, and a
    brief \texttt{reason}.
\end{tcolorbox}
\begin{minipage}{\linewidth}
\captionsetup{hypcap=false}
\captionof{figure}{Judge prompt for Memory QA recall.}
\label{fig:judge-memory-recall}
\end{minipage}

\begin{tcolorbox}[
        breakable,
        colback=gray!6,
        colframe=gray!45,
        boxrule=0.5pt,
        arc=1pt,
        left=6pt,
        right=6pt,
        top=6pt,
        bottom=6pt]
    \small
    You are a strict memory precision and accuracy evaluator. The program
    has already parsed the assistant's final summary into candidate
    bullets. Do not split, merge, or invent candidate units. Evaluate each
    candidate bullet exactly as one memory candidate.

    \medskip
    \textbf{For each candidate}\par
    Decide whether its field belongs to the target gold memory field set,
    and whether its content is supported by the gold memory points.

    \medskip
    \textbf{Accuracy score}\par
    2: The candidate's core content is supported by the gold memory points
    and has no contradictory detail.\par
    1: The candidate's core content is partly supported, but it misses an
    important detail, is uncertain, or is slightly wrong.\par
    0: The candidate is contradicted by the gold memory points, belongs to
    a wrong current value, or is unrelated to the target memory fields.

    \medskip
    \textbf{is\_target\_field}\par
    True if the candidate field semantically matches one of the gold
    memory fields; false if the field is outside the gold memory field set.

    \medskip
    \textbf{Rules}\par
    Use only the gold memory points, not outside knowledge. Gold points may
    include current values, historical values, intermediate values, and
    change events; any of those are supported when the candidate matches
    the same target field and content.\par
    Natural phrasing, harmless qualifiers, and non-conflicting casual
    wording should not reduce accuracy. Do not penalize a candidate
    merely because it contains a non-conflicting wording detail absent
    from the gold text.\par
    Preserve \texttt{candidate\_index} exactly and return one result per
    candidate bullet.

    \medskip
    \textbf{Gold memory points}: \texttt{\{gold\_points\_json\}}\par
    \textbf{Candidate bullets}: \texttt{\{candidate\_bullets\_json\}}\par
    Return only valid JSON with a \texttt{candidates} array. Each record
    contains \texttt{candidate\_index}, \texttt{field},
    \texttt{accuracy\_score} (0, 1, or 2),
    \texttt{is\_target\_field}, \texttt{matched\_gold\_field\_path}, and
    a brief \texttt{reason}.
\end{tcolorbox}
\begin{minipage}{\linewidth}
\captionsetup{hypcap=false}
\captionof{figure}{Judge prompt for Memory QA precision.}
\label{fig:judge-memory-precision-scheduled}
\end{minipage}

\end{document}